%% file: main.tex
\pdfoutput=1
\documentclass{article}
\usepackage[T1]{fontenc}
\usepackage{iclr2027_conference,times}
\usepackage{amsmath,amssymb,amsthm,booktabs,graphicx,array,tabularx}
\usepackage{tikz}
\usetikzlibrary{arrows.meta,positioning,calc}
\definecolor{paperblue}{HTML}{0173B2}
\definecolor{paperorange}{HTML}{DE8F05}
\definecolor{paperteal}{HTML}{029E73}
\usepackage{xurl}
\usepackage{hyperref}
\usepackage{url}
\newcommand{\E}{\mathbb{E}}
\newcommand{\R}{\mathbb{R}}
\newcommand{\KL}{\operatorname{KL}}
\newcommand{\C}[1]{\texttt{#1}}

\newtheorem{proposition}{Proposition}

\title{Bridging Singular Learning Theory\\and Mechanistic Interpretability\\at the Billion-Parameter Scale}
\author{Ethan Ji, Wei Wang \& Yizhou Sun \\
Department of Computer Science\\
University of California, Los Angeles\\
\texttt{\{eji,weiwang,yzsun\}@cs.ucla.edu}
}
\iclrfinalcopy
\begin{document}
\maketitle
\lhead{Preprint}
\begin{abstract}
Mechanistic interpretability (MI) explains a computation in a trained network by identifying the components and variables that implement it and testing them with interventions. Singular learning theory (SLT) describes the same network through the geometry of the loss around its parameters, summarized by the local learning coefficient, whose estimates track stages of training. In pretrained language models the two have not been compared: SLT's estimates at this scale concern whole models or components whose roles were not independently established, and the estimator, which averages the loss over parameter settings drawn near the trained ones, has had no reference for how far those draws may move the parameters. We show that both are functions of one object, the map from a component's parameters to the model's predictions under inputs and interventions, whose first derivatives determine the Fisher information, the learning coefficient, and intervention effects, and whose higher derivatives add a quartic correction and pairwise terms. We test this correspondence in Pythia-1B on Dyck bracket matching at all 154 training checkpoints, against a five-stage circuit identified by interventions. We measure the local geometry of all 144 components inside the region where predictions are preserved, with the Fisher spectrum at every checkpoint, which gives the quadratic part of the learning-coefficient estimate at every temperature. Changes of the Fisher trace track changes of causal effect (median partial rank correlation $0.51$ between adjacent checkpoints, replicated on independent inputs), largely as a trend shared across the model, whereas the gain Fisher of a component's output and the large-temperature count are specific to the component. The count follows causal change across the population but not the growth of a formed circuit component. The quartic part is what a sampler that leaves the region registers beyond curvature. Pairwise interaction curvature tracks joint-ablation interactions.
\end{abstract}
\input{sections/introduction}
\input{sections/related}
\input{sections/background}

\input{sections/motivation}
\input{sections/findings}

\input{sections/discussion}

\input{sections/conclusion}
\input{sections/statements}
\bibliography{references}
\bibliographystyle{iclr2027_conference}
\clearpage
\appendix
\input{sections/theory_appendix}
\input{sections/nonquadratic_theory}
\input{sections/interaction_methods}
\input{sections/grid_methods}
\input{sections/temperature_sweep}
\input{sections/methods_appendix}

\input{sections/locality_appendix}
\section{Calibration of local geometry against matched causal effects}
\label{app:calibration}
\input{sections/matched_gibbs_findings}
\input{sections/sensitivity_findings}
\input{sections/gibbs_methods}

\input{sections/matched_gibbs_appendix}
\input{sections/semantic_projection_appendix}
\input{sections/sensitivity_appendix}

\input{sections/development_detail}
\input{sections/algorithm_appendix}

\input{sections/readout_development}
\end{document}

%% file: sections/introduction.tex
\section{Introduction}
\label{sec:intro}

A trained language model carries out many computations, such as tracking which brackets are still open, and two research directions describe how. Mechanistic interpretability (MI) draws a wiring diagram: it identifies the \emph{components} that carry a computation (attention heads and MLPs) and the task quantities they compute and pass on, such as the current nesting depth, which we call \emph{variables}, and it confirms each assignment by intervening on the network, such as removing a component's output \citep{olsson2022induction,wang2023ioi,conmy2023acdc}. Singular learning theory (SLT) surveys the terrain around the trained parameters: many nearby parameter settings compute the same function, and the local learning coefficient (LLC) $\lambda$ measures the effective dimension of the trained solution: the volume of parameter settings whose loss lies within $\epsilon$ of the minimum shrinks like $\epsilon^{\lambda}$, so a model whose $d$ parameters are all pinned down by the data has $\lambda=d/2$, and every direction the data leave free lowers it \citep{watanabe2009,lau2024llc}. If MI's diagram is the wiring of a machine, the LLC measures how far its parts can move without changing what it does; it marks stages of learning in small transformers \citep{hoogland2024stagewise,wang2024rllc}.

The two descriptions concern the same computation, so a component that carries more of it should both matter more under intervention and be more tightly constrained: more of the output depends on its parameters, so removing its output changes the prediction more and displacing its parameters raises the loss faster. Establishing this correspondence matters because each description supplies what the other lacks: MI's accounts are assembled by hand, one task and checkpoint at a time, and depend on the choice of intervention, whereas SLT's measurements rest on a theory of generalization and can be repeated automatically over every component and checkpoint, but cannot by themselves say which computation a change in geometry reflects. A correspondence would let geometry measured throughout training signal when and where a mechanism forms, which is what monitoring the development of capabilities in large models requires, but three obstacles stand in the way. A pretrained language model offers no independently identified mechanism to check against. The standard estimator draws parameter settings with stochastic-gradient Langevin dynamics (SGLD), and nothing at this scale says how far the draws may move: under standard localization they move each weight by several times its magnitude and change the model's predictions, while a localization strong enough to hold them where predictions are preserved leaves only the prior (Appendix~\ref{app:locality}). And near a trained network any such estimate is a function of the Fisher information, which says how steeply the loss rises in each direction, plus corrections, so whether SLT adds anything beyond curvature can only be decided by computing both.

We first make explicit what both descriptions depend on (Section~\ref{sec:motivation}): the map from a component's parameters to the model's next-token predictions on fixed inputs. Writing $\theta$ for the trained parameters and $\theta+\delta\theta$ for a modified model, both descriptions measure change by the Kullback--Leibler (KL) divergence $\KL\bigl(p(y\mid x;\theta)\,\|\,p(y\mid x;\theta+\delta\theta)\bigr)$ from the trained model's next-token distribution to the modified one, averaged over inputs $x$. The first derivatives of the predictions with respect to the parameters determine the Fisher matrix, which describes how this divergence rises quadratically, as a parabola, around the trained parameters, and with it the effects of small interventions and, when their rank is constant, the learning coefficient. Beyond the parabola, averaging over perturbations of either sign cancels the odd-order terms, so the first correction is quartic in the size of the perturbation, and we summarize it by a quartic coefficient; and when two components are perturbed together, the divergence contains \emph{mixed terms}, those that depend on both perturbations at once, which express how changing one component alters the network's response to the other (Section~\ref{sec:background} states these expansions precisely). We then test the correspondence in Pythia-1B at all 154 released training checkpoints \citep{biderman2023pythia} on bounded Dyck languages of nested brackets, where the model must predict the next closing bracket. The task is simple by design: testing whether geometry and mechanism agree (in the sense of Section~\ref{sec:background}) requires a mechanism known independently of the geometry, and Dyck languages, the canonical model of the nested structure of syntax and code \citep{chomsky1963algebraic,hewitt2020rnns}, come with a known algorithm, exact values for every intermediate variable, and well-studied transformer solutions \citep{yao2021bounded,strobl2024survey}, yet Pythia-1B acquires the skill only partially and gradually during pretraining (Section~\ref{sec:background}). It falls short because it is a base model, trained to continue text rather than to complete brackets: at most positions where a closing bracket is scored, an opening bracket is also a legal continuation, $77\%$ of its errors are such openers, and among the four closers it ranks the correct one first at $86\%$ of positions (Appendix~\ref{app:methods}). SLT estimates $\lambda$ by treating the parameters near a checkpoint as random: it draws whole parameter settings from a distribution concentrated around the trained values (a localized, tempered posterior), like dropping marbles near the bottom of a valley and recording how high each comes to rest, and averages the loss over the draws; the average grows with the number of directions in which the valley is steep. Our draws change one component at a time and stay inside an explicit region where the predictions are preserved, and evaluating the Fisher approximation on the same draws splits each estimate into its Fisher and quartic parts.

\begin{figure}[t]
\centering
\resizebox{\linewidth}{!}{\input{figures/study_pipeline}}
\caption{\textbf{Workflow.} The circuit is identified with interventions alone, the local geometry of every component is measured inside the region where predictions are preserved, and the two are compared term by term across the 154 checkpoints.}
\label{fig:study-pipeline}
\end{figure}
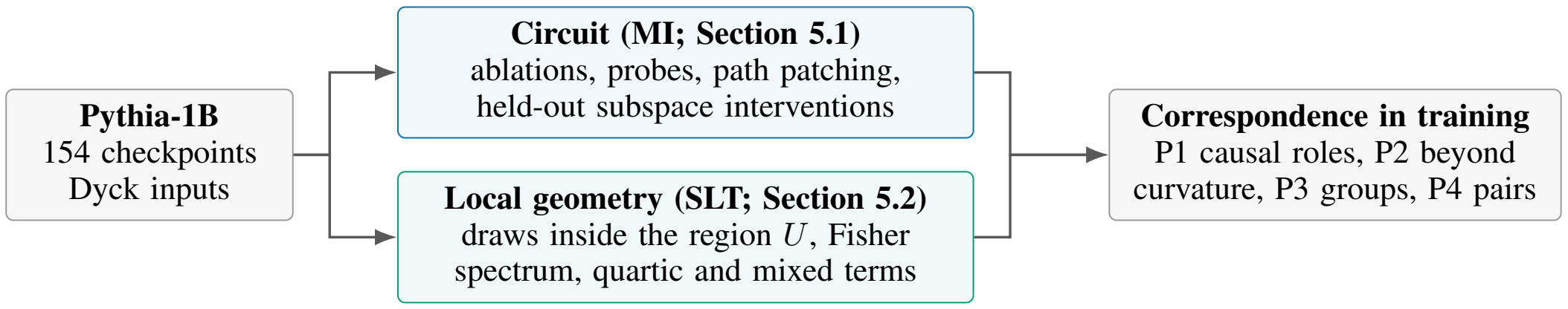

Our contributions (Figure~\ref{fig:study-pipeline}) are the following. \textbf{Problem.} The first test of whether SLT's geometry and MI's causal account agree in a pretrained language model, against a five-stage bracket circuit in Pythia-1B validated by held-out interventions (Section~\ref{sec:circuit}). \textbf{Solution.} A formal correspondence between the terms of the loss expansion~\eqref{eq:intro-expansion} and intervention effects (Section~\ref{sec:motivation}), and a measurement whose draws stay where predictions are preserved, with the Fisher spectrum of all 144 components at all 154 checkpoints and a replication on independent inputs (Section~\ref{sec:local-development}). \textbf{Results.} Fisher-trace changes track causal-effect changes (median partial rank correlation $0.51$), largely as a trend shared across the model, while the gain Fisher and the large-temperature count carry what is specific to a component; the quartic part is what a nonlocal sampler registers; and the mixed terms track joint-ablation interactions (Sections~\ref{sec:correspondence}--\ref{sec:interaction-development}).

%% file: figures/study_pipeline.tex
\begin{tikzpicture}[x=1cm,y=1cm,>=Latex,font=\footnotesize,
 box/.style={draw=black!40,rounded corners=2pt,align=center,inner sep=4pt},
 arrow/.style={->,line width=.7pt,black!65}]
\path[use as bounding box] (-.1,-1.55) rectangle (15.05,1.55);
\node[box,fill=black!3,text width=2.5cm] (input) at (1.35,0)
 {\textbf{Pythia-1B}\\154 checkpoints\\Dyck inputs};
\node[box,draw=paperblue,fill=paperblue!5,text width=5.3cm] (mi) at (6.55,0.8)
 {\textbf{Circuit (MI; Section~\ref{sec:circuit})}\\ablations, probes, path patching,\\held-out subspace interventions};
\node[box,draw=paperteal,fill=paperteal!6,text width=5.3cm] (slt) at (6.55,-0.8)
 {\textbf{Local geometry (SLT; Section~\ref{sec:local-development})}\\draws inside the region $U$, Fisher\\spectrum, quartic and mixed terms};
\node[box,fill=black!3,text width=4.1cm] (compare) at (12.85,0)
 {\textbf{Correspondence in training}\\P1 causal roles, P2 beyond\\curvature, P3 groups, P4 pairs};
\draw[arrow] (input.east) -- ++(.35,0) |- (mi.west);
\draw[arrow] (input.east) -- ++(.35,0) |- (slt.west);
\draw[arrow] (mi.east) -- ++(.35,0) |- (compare.west);
\draw[arrow] (slt.east) -- ++(.35,0) |- (compare.west);
\end{tikzpicture}

%% file: sections/related.tex
\section{Related Works}
\label{sec:related}

\paragraph{Singular learning theory in trained networks.}
Singular learning theory describes a trained network by the shape of its loss around the trained parameters. Its central quantity is the local learning coefficient (LLC) $\lambda$, the effective number of parameter directions that the data pin down: the volume of parameter settings whose loss lies within $\epsilon$ of the trained model's shrinks like $\epsilon^{\lambda}$ (Section~\ref{sec:background}). \citet{lau2024llc} made it estimable by drawing parameter settings with SGLD from a localized, tempered posterior \citep{welling2011sgld,devinterp}. Its trajectory marks learning stages in small transformers \citep{hoogland2024stagewise,carroll2025transient}, its weight-refined version separates the roles of attention heads in models with known mechanisms \citep{wang2024rllc}, and it, with related susceptibility methods \citep{baker2025susceptibilities,gordon2026spectroscopy,wang2026patterning}, has been applied to Pythia models up to 6.9B parameters \citep{urdshals2025compressibility} and to a 9B reward model \citep{wang2026rewardpatterning}. At this scale, however, no independently identified mechanism has been available, the estimator has had no reference for how far its draws may move the parameters, and the estimate has not been separated from the Fisher information, itself a standard measure of parameter importance \citep{kirkpatrick2017ewc,molchanov2019importance,mackay1992practical}. We supply all three: a circuit validated by interventions, draws confined to where predictions are preserved, and the exact Fisher approximation evaluated on the same draws as the primary control (Equation~\eqref{eq:soft-rank}).

\paragraph{Circuits and causal interpretation.}
Mechanistic interpretability explains a behavior by a circuit: components, the variables they compute, and the interventions that confirm each assignment \citep{wang2023ioi,conmy2023acdc,geiger2025causal}. What an intervention measures depends on the replacement value, the grouping of components, and the output metric \citep{heimersheim2024patching}, and a role shared by several components can be missed by removing any one, as with backup heads \citep{wang2023ioi}. We establish the circuit with held-out interventions on variable subspaces rather than probes alone, compare geometry with joint as well as single ablations, and obtain from the geometry a measure of a role that needs no replacement value.

%% file: sections/background.tex
\section{Problem Formulation and Preliminaries}
\label{sec:background}

We ask: \emph{for a computation performed by a pretrained language model, do the components that MI finds to carry it, and the points in training at which their causal roles form, coincide with where and when SLT finds the parameters constrained?} (Q1) At a checkpoint, does a component's geometric coefficient rank components as intervention effects do, and (Q2) across checkpoints, do the coefficient and the effect change at the same times? We say that geometry and mechanism \emph{agree} to the extent that both answers are yes, as measured by the rank correlations of Section~\ref{sec:correspondence}. We now define the terms these questions need.

\paragraph{The Dyck bracket matching task.}
The bounded Dyck language $\mathrm{Dyck}(k,D)$ contains the correctly nested strings of $k$ bracket types whose depth never exceeds $D$, such as \texttt{( [ ] ( ) )} for $k=2$; wherever the next token is a closing bracket, the correct one is the partner of the most recent unmatched opener. It suits our question for four reasons. It is the canonical hierarchical language: every context-free language is a homomorphic image of a Dyck language intersected with a regular language \citep{chomsky1963algebraic}, and the depth bound mirrors the limited nesting of natural-language syntax \citep{hewitt2020rnns} and puts the language within reach of fixed-size self-attention networks \citep{yao2021bounded}, which unbounded Dyck is not \citep{hahn2020limits}. It is solved by a known algorithm, a stack, whose variables (the depth, the type on top of the stack, the distance to the most recent unmatched opener) have exact values at every position, giving probes and interventions exact targets. It is a standard test of what recurrent networks and transformers learn and express \citep{suzgun2019lstm,bhattamishra2020ability,strobl2024survey}, with well-characterized solutions that attend to the top of the stack \citep{yao2021bounded,ebrahimi2020dyck} and can be written and learned as interpretable programs \citep{weiss2021thinking,friedman2023learning}; yet trained networks can also solve it without stack-like attention \citep{wen2023myopic} and carry decodable variables they do not use \citep{sharma2026dissociating}, so its circuit must be established by interventions. Finally, Pythia-1B, trained on the Pile \citep{gao2020pile}, which includes source code, reaches a final closing-bracket accuracy of $0.595$ for $k=4$ and improves across the 154 checkpoints.

\paragraph{Components and the residual stream.}
A decoder-only transformer keeps one vector per token position, which every layer updates: the heads and the MLP of a layer read the vector, compute an output, and add it back, so the vector entering a layer is the sum of the token embedding and all earlier outputs: the residual stream \citep{elhage2021mathematical}, the shared workspace through which components communicate. A \emph{component} is one head or one MLP, identified with its parameters (Table~\ref{tab:protocol}). Pythia-1B has 16 layers of eight heads and one MLP, 144 components; \C{L}$\ell$\C{H}$h$ is head $h$ of layer $\ell$ and \C{L}$\ell$\C{.mlp} its MLP, counting from zero.

\begin{figure}[t]
\centering
\begin{minipage}[c]{.5\linewidth}
\centering
\resizebox{\linewidth}{!}{\input{figures/bridge_intuition}}
\end{minipage}\hfill
\begin{minipage}[c]{.47\linewidth}
\caption{\textbf{Two descriptions of the same computation.} MI intervenes on the variables that a circuit computes (A); SLT studies the parameter settings that preserve its behavior (B). For $K(u,v)=(u+v)^2$, cancelling changes leave the output unchanged, and $\{K<\epsilon\}$ is a strip of area $\propto\epsilon^{1/2}$: $\lambda=\tfrac12$ with two parameters.}
\label{fig:bridge-intuition}
\end{minipage}
\end{figure}
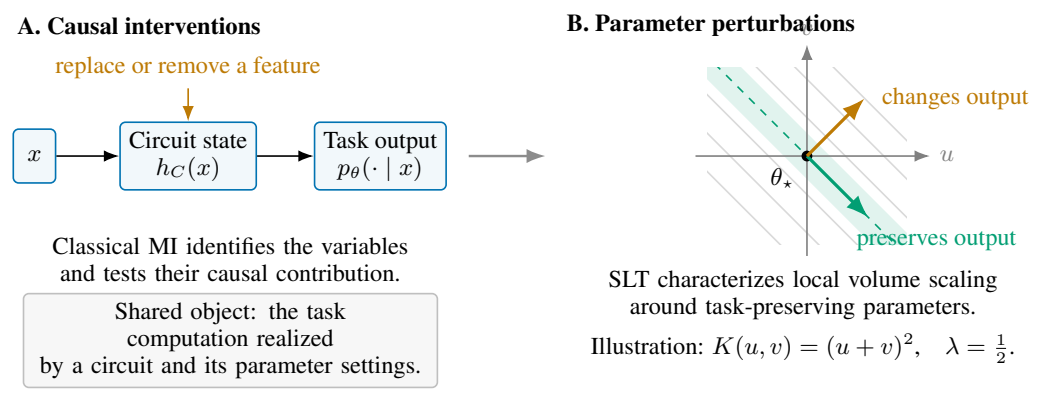

\paragraph{Variables, circuits, probes, and interventions.}
A \emph{variable} $s$ is a quantity defined by the task's algorithm, such as the depth or the type of the most recent unmatched opener, and the algorithm's \emph{operations} compute variables from others (``push the opener's type onto the stack''). The algorithm assigns each variable a value $s_t$ after the first $t$ tokens; whether the network represents or uses it is tested by the probes and interventions below. A \emph{circuit hypothesis} names a set of components, the variables each reads from and writes to the residual stream, and the operations by which it does so \citep{wang2023ioi,conmy2023acdc,geiger2025causal}. A \emph{probe} tests whether a variable is represented: a linear map $\omega$ fitted to predict $s_t$ from the residual stream $h_{\ell,t}$ at layer $\ell$ is scored by its held-out $R^2$ \citep{alain2016probes,belinkov2022probing}. An \emph{intervention} tests whether the variable is used: it replaces a component's output by a modified value (zero, a mean over inputs, or a projection onto a subspace) and measures the change in the prediction; replacement by zero, \emph{zero-ablation}, measures how much the prediction depends on the component, and a large effect of removing a variable's subspace, against small effects for random subspaces of equal dimension, shows that the component acts through that variable (Figure~\ref{fig:bridge-intuition}A).

\paragraph{The loss geometry of a component.}
Write $x$ for a token prefix and $y$ for the next token, so that the model defines $p_\theta(y\mid x)$. Fix a checkpoint $\theta_\star\in\R^d$ and a component $c$, whose parameters occupy $d_c$ of the $d$ coordinates. A displacement $u\in\R^{d_c}$ of those coordinates alone, a $d_c$-dimensional vector, gives the model $\theta(u)=\theta_\star+E_cu$, where the $d\times d_c$ matrix $E_c$
 places each coordinate of $u$ at the component's position and scales it by the root-mean-square (RMS) of its tensor, so that one unit of $u$ moves a weight by the typical ``average'' magnitude of its tensor. On fixed task inputs, define
\begin{equation}
 K_c(u)=\E_{x}\,\KL\!\bigl(p_{\theta_\star}(\cdot\mid x)\,\|\,p_{\theta(u)}(\cdot\mid x)\bigr),
 \label{eq:component-kl}
\end{equation}
averaged over the positions at which the task is scored. We use the KL to the checkpoint's predictions rather than the cross-entropy of the observed tokens because it makes the checkpoint an exact minimum, as SLT's local analysis assumes, and intervention effects use the same divergence. Its Hessian at $u=0$, the matrix of second derivatives that says how steeply the loss rises in each direction, is the component's Fisher information matrix $F_c$. SLT characterizes the geometry of $K_c$ by how fast the prior volume of the displacements that keep the loss below $\epsilon$ shrinks as $\epsilon\to0$ \citep{watanabe2009}: if the loss were quadratic with a Hessian of rank $\varrho$, that volume would shrink like $\epsilon^{\varrho/2}$, and in general it shrinks like a power of $\epsilon$ whose exponent is the LLC $\lambda_c$.
For a neighborhood $U$ of zero and a smooth positive prior density $\varphi$,
\begin{equation}
 \int_{U\cap\{K_c(u)<\epsilon\}}\varphi(u)\,du
 \asymp\epsilon^{\lambda_c}\bigl(\log(1/\epsilon)\bigr)^{m_c-1}
 \qquad(\epsilon\to0),
 \label{eq:volume}
\end{equation}
where $f\asymp g$ means that $f/g$ stays between two positive constants, $\lambda_c$ is the LLC of the component (the weight-refined LLC of \citealp{wang2024rllc}) and $m_c$ its multiplicity. A direction along which the loss rises only as $u^4$ contributes $1/4$ to $\lambda_c$ rather than $1/2$, and an unused direction $0$ (Figure~\ref{fig:bridge-intuition}B), so $\lambda_c$ counts the constrained directions weighted by how the loss rises along them.

\paragraph{Beyond the Fisher matrix.}
For a displacement $u=rz$ of scale $r$ and Gaussian direction $z\sim\mathcal N(0,I)$, the loss and its average over directions expand as
\begin{equation}
 K_c(rz)=\tfrac12r^2z^\top F_cz+r^3K_3(z)+r^4K_4(z)+O(r^5),\qquad
 \E_zK_c(rz)=r^2A_c+r^4B_c+O(r^6),
 \label{eq:intro-expansion}
\end{equation}
where the cubic and quartic Taylor terms $K_3$ and $K_4$ also involve second derivatives of the predictions. The first term is a parabola in $r$, whose average is set by the Fisher trace $A_c=\tfrac12\operatorname{tr}F_c$; since $z$ and $-z$ are equally likely, the odd terms average out, and the first departure from the parabola is the quartic term, with the quartic coefficient $B_c=\E_zK_4(z)$. For two components $a$ and $b$ displaced together, $u=(u_a,u_b)$, the loss is not the sum of the two separate losses: the difference $K_{ab}(u_a,u_b)-K_{ab}(u_a,0)-K_{ab}(0,u_b)=u_a^\top F_{ab}u_b+\cdots$ collects the \emph{mixed terms}, those containing both displacements. The leading one is set by the off-diagonal block $F_{ab}$ of the joint Fisher matrix; the higher ones depend also on the joint second derivative of the predictions, which is nonzero when changing one component alters the network's response to the other.

\paragraph{Estimating the coefficient.}
\citet{lau2024llc} estimate the LLC by drawing parameter settings from a distribution that favors low loss and stays near $\theta_\star$, a posterior tempered to inverse temperature $n\beta$ and localized by a Gaussian term of strength $\gamma$, and averaging the loss increase over the draws, scaled by $n\beta$. For one component and the loss~\eqref{eq:component-kl}, with the neighborhood made explicit, the estimate is
\begin{equation}
 \pi_c(u)\propto \mathbf1_U(u)\,e^{-\tau K_c(u)}\,\varphi_r(u),
 \qquad
 q_c(\tau,r)=\tau\,\E_{\pi_c}K_c(u),
 \label{eq:estimator}
\end{equation}
with inverse temperature $\tau$ in place of $n\beta$ and a Gaussian prior $\varphi_r=\mathcal N(0,r^2I)$ whose standard deviation $r$, the radius, sets the localization ($\gamma=1/r^2$). If the volume law~\eqref{eq:volume} holds where the draws lie, $q_c\to\lambda_c$ as $\tau\to\infty$. For a quadratic loss, with $\mu_i$ the eigenvalues of $F_c$,
\begin{equation}
 q_c(\tau,r)=\frac12\sum_{i=1}^{d_c}\frac{\tau r^2\mu_i}{1+\tau r^2\mu_i},
 \label{eq:soft-rank}
\end{equation}
half the effective number of well-determined parameters of \citet{mackay1992practical}, the \emph{soft rank} of $\tau r^2F_c$: each direction with curvature above $1/(\tau r^2)$ contributes about one half, so at large $\tau$ the estimate counts constrained directions, whereas $q_c\approx\tfrac12\tau r^2\operatorname{tr}F_c$ when all $\tau r^2\mu_i$ are small. Unlike prior SGLD-based estimates \citep{lau2024llc,wang2024rllc}, we fix $U$ explicitly and compare each estimate with that of the quadratic approximation $Q_c(u)=\tfrac12u^\top F_cu$ on the same draws, isolating what lies beyond curvature.

%% file: figures/bridge_intuition.tex
\begin{tikzpicture}[x=1cm,y=1cm,>=Latex,font=\small,line width=.6pt,
  circuit/.style={draw=paperblue,fill=paperblue!5,rounded corners=2pt,minimum height=.70cm,align=center}]
\path[use as bounding box] (-.15,-1.65) rectangle (13.5,3.25);
\node[font=\small\bfseries,anchor=west] at (0,2.95) {A. Causal interventions};
\node[circuit,minimum width=.55cm] (x) at (.35,1.25) {$x$};
\node[circuit,minimum width=1.75cm] (h) at (2.35,1.25) {Circuit state\\$h_C(x)$};
\node[circuit,minimum width=1.70cm] (p) at (4.85,1.25) {Task output\\$p_\theta(\cdot\mid x)$};
\draw[->] (x) -- (h);
\draw[->] (h) -- (p);
\node[font=\footnotesize,text=paperorange!85!black] (change) at (2.35,2.4) {replace or remove a feature};
\draw[->,paperorange!85!black] (change.south) -- (h.north);
\node[align=center,text width=5.4cm] at (2.9,-.1)
  {Classical MI identifies the variables\\and tests their causal contribution.};
\node[draw=black!25,rounded corners=2pt,fill=black!3,align=center,
  text width=5.15cm,font=\footnotesize] at (2.9,-1.15)
  {Shared object: the task computation realized\\by a circuit and its parameter settings.};
\draw[->,black!45,line width=.9pt] (6.0,1.25) -- (7.0,1.25);
\node[font=\small\bfseries,anchor=west] at (7.15,2.95) {B. Parameter perturbations};
\begin{scope}[shift={(10.4,1.25)}]
  \begin{scope}
    \clip (-1.3,-1.15) rectangle (1.3,1.15);
    \fill[paperteal!12] (-1.6,1.30) -- (-1.30,1.60) -- (1.60,-1.30) -- (1.30,-1.60) -- cycle;
    \foreach \a in {-1.5,-1.0,-.5,.5,1.0,1.5}
      \draw[black!15] (-2,2+\a) -- (2,-2+\a);
    \draw[paperteal,dashed] (-1.5,1.5) -- (1.5,-1.5);
  \end{scope}
  \draw[->,black!50] (-1.45,0) -- (1.6,0) node[right] {$u$};
  \draw[->,black!50] (0,-1.3) -- (0,1.45) node[above] {$v$};
  \fill[black] (0,0) circle (2pt);
  \node[anchor=north east,font=\footnotesize] at (-.04,-.03) {$\theta_\star$};
  \draw[->,paperteal,line width=1.2pt] (0,0) -- (.80,-.80);
  \draw[->,paperorange!85!black,line width=1.2pt] (0,0) -- (.75,.75);
  \node[anchor=west,font=\footnotesize,text=paperorange!85!black] at (.85,.75) {changes output};
  \node[anchor=west,font=\footnotesize,text=paperteal] at (.52,-1.1) {preserves output};
\end{scope}
\node[align=center,text width=6cm] at (10.35,-.55)
  {SLT characterizes local volume scaling\\around task-preserving parameters.};
\node[font=\footnotesize] at (10.35,-1.25)
  {Illustration: $K(u,v)=(u+v)^2$,\quad $\lambda=\tfrac12$.};
\end{tikzpicture}

%% file: sections/motivation.tex
\section{Methodology: one object, four predictions}
\label{sec:motivation}

\paragraph{The shared quantity: a circuit's responses.}
We view the transformer as a directed acyclic graph whose nodes are the token embedding, every head and MLP, and the unembedding, with an edge from $w$ to $v$ when $v$ lies in a later layer (the unembedding counts as the last), because $v$ reads the residual stream that $w$ writes to. Node $v$ outputs $o_v=f_v(\sum_{w\in\mathrm{pa}(v)}o_w;\theta_v)$, with $\mathrm{pa}(v)$ its parents and $\theta_v$ its parameters, and a circuit is a subgraph whose nodes carry the variables and operations of a circuit hypothesis \citep{geiger2025causal}. Fix inputs $x\sim D$ with scored positions. The circuit's \emph{responses} are its predictions $p_\theta(\cdot\mid x)$ on these inputs as a function of the parameters; both descriptions depend on this response map: MI characterizes the computation that produces the responses, and SLT the parameters that preserve them, through the loss~\eqref{eq:component-kl}, which vanishes exactly where every response is preserved. The two tests act on the same network: patching the output change $\Delta o_c(x;u)$ caused by a displacement $u$ into the unperturbed network reproduces the perturbed prediction exactly,
\begin{equation}
 p_{\theta(u)}(\cdot\mid x)
 =p_{\theta_\star}^{\mathrm{patch}[\Delta o_c(u)]}(\cdot\mid x),
 \label{eq:parameter-activation}
\end{equation}
so a parameter perturbation is an activation intervention, and projecting $\Delta o_c$ onto variable subspaces shows which variable carries a geometric response (Appendix~\ref{app:semantic-projection}).

\paragraph{Why the measurements should agree.}
Let $G_c(u)$ collect the changes in the model's log-odds (log-ratios of output probabilities) at every input, scored position, and intervention. Since all probabilities are positive, $K_c$ lies between two multiples of $\|G_c\|^2$ near $u=0$, so the local geometry of the loss is that of this response map.
\begin{proposition}[Shared local constraints]
\label{prop:bridge}
If the Jacobian $DG_c$ has constant rank $\varrho_c$ near zero, then $\lambda_c=\varrho_c/2$. Moreover, $K_c(tu)=\tfrac12t^2u^\top F_cu+o(t^2)$ with $F_c=DG_c(0)^\top\Sigma\,DG_c(0)$, where $\Sigma$ is the Fisher information of the categorical output distribution (the metric under which two nearby output distributions are as far apart as they are distinguishable), so that $\ker F_c=\ker DG_c(0)$.
\end{proposition}
The LLC counts the independent constraints that the responses place on the component, and $F_c$ measures how strongly each parameter direction moves them (Appendix~\ref{app:bridge}). An intervention moves the same responses. Write $\Sigma_c$ for the Fisher metric that the downstream network induces on the component's output activation $o_c$: then $F_c$ is its pullback through $\partial o_c/\partial u$, so the Fisher trace $A_c=\tfrac12\operatorname{tr}F_c=\tfrac12\sum_i\|\partial o_c/\partial u_i\|_{\Sigma_c}^2$, and the KL effect of scaling the output by $1+g$ is $\tfrac12g^2\|o_c\|_{\Sigma_c}^2$ to leading order (Appendix~\ref{app:intervention-scale}). Both are norms under the same metric, which grows as later components read that output, so the two grow together. \textbf{P1:} \emph{changes in a component's Fisher information track changes in its causal effect.} As a geometric coefficient is nonnegative and a task-score effect is signed, we measure effects by the KL of the ablated model from the clean one.

\paragraph{Beyond curvature: the quartic coefficient.}
Write $q_K$ and $q_Q$ for the estimate~\eqref{eq:estimator} computed with $K_c$ and with its quadratic part $Q_c$ under the same prior. Averaging the expansion~\eqref{eq:intro-expansion}, with $Q_c(z)=\tfrac12z^\top F_cz$, under the prior gives (Appendix~\ref{app:nonquadratic-geometry})
\begin{equation}
 q_Q=\tau r^2A_c+O(\tau^2r^4),\qquad
 q_K-q_Q=\tau r^4B_c+O(r^6).
 \label{eq:main-quartic}
\end{equation}
The quartic coefficient $B_c=\tfrac18\sum_{i,j}\partial_i^2\partial_j^2K_c(0)$ is the average quartic term: positive when the loss rises faster than its quadratic approximation and negative when it saturates, with $B_c/A_c$ the relative excess per unit $r^2$. Expanding $K_c=\E_x\log\E_{p_{\theta_\star}}e^{G_c}$ shows that $K_4$ depends on the second derivatives of the response map and on the skewness and kurtosis of the output distribution, which all components share and which change as the predictions sharpen, so that components with identical Fisher matrices can differ in $B_c$ (Appendix~\ref{app:nonquadratic-geometry}). \textbf{P2:} \emph{$B_c$ carries what an estimate registers beyond curvature, a trend shared across components rather than a component's own effect.}

\paragraph{Groups and pairs.}
When several components write the same variable, each ablation can be small while the joint ablation is large, and for a quadratic loss the joint LLC estimate of a group is at most the sum of its members' estimates, with equality only for uncoupled directions (Equation~\eqref{eq:joint-geometry}). \textbf{P3:} \emph{for a redundant group, summed single-component geometry tracks summed single ablations rather than the joint ablation, and the joint estimate lies below the sum.} For components $a,b$ with outputs scaled by $1+g_a$ and $1+g_b$, and $G(g_a,g_b)$ collecting the log-odds changes, the mixed terms beyond the Fisher block (Section~\ref{sec:background}) are measured by the interaction curvature $N_{ab}$, the squared $\Sigma$-norm of the part of the joint response $\partial_{g_a}\partial_{g_b}G(0)$ that neither component's own response can produce (Equation~\eqref{eq:normal-interaction-curvature}; \citealp{efron1975curvature}), and by the mixed quartic coefficient $B_{ab}=\tfrac14\partial_{g_a}^2\partial_{g_b}^2(K-K_{\mathrm{lin}})(0)$, with $K_{\mathrm{lin}}$ the loss of a reference model whose logits are linear in $(g_a,g_b)$ (Equation~\eqref{eq:interaction-free-energy-correction}); both vanish for logits linear in $g_a$ and $g_b$. \textbf{P4:} \emph{$N_{ab}$ and $B_{ab}$ track the nonadditive part of joint ablations.}

%% file: sections/findings.tex
\section{Findings}
\label{sec:findings}

Following Figure~\ref{fig:study-pipeline}, inputs are $\mathrm{Dyck}(2,10)$ or $\mathrm{Dyck}(4,16)$ strings, written as space-separated brackets after the prefix \texttt{Brackets:}. At every position whose next token is a closer (a closing bracket), we score, for the unmodified (clean) model and for intervened ones, closer accuracy (whether the most probable token in the vocabulary is the correct closer) and the contrast $S$, the logit difference between the correct closer and one fixed, randomly drawn incorrect closer \citep{wang2023ioi}. Table~\ref{tab:protocol} lists the protocol of the geometric measurements, which share their inputs with the ablations.

\begin{table}[t]
\centering\scriptsize
\setlength{\tabcolsep}{4pt}
\renewcommand{\arraystretch}{0.9}
\begin{tabularx}{\linewidth}{@{}>{\raggedright\arraybackslash}p{.13\linewidth}X@{}}
\toprule
Model & Pythia-1B at its 154 checkpoints (steps 0, 1, 2, 4, \dots, 512, then every 1,000), in float64; steps 0, 512, 2,000, 8,000, 32,000, and 143,000 served to choose settings, and statistics use the other 148. \\
Inputs & 128 strings of 128 tokens with $k=4$ (opening probability $0.45$ when both actions are legal; seed 0); averages over the 5,475 closer positions. Replication: an independent batch (seed 20261001) for the 33-component panel. \\
Components & All 144 heads and MLPs; the 26 circuit components of Section~\ref{sec:circuit} and a panel of 33 (circuit plus seven controls drawn uniformly; Appendix~\ref{app:grid-methods}) that carries the radius--temperature grid and the replication. \\
Parameters & A head's slices of $W_Q,W_K,W_V,W_O,b_Q,b_K,b_V$; an MLP's $W_{\rm in},b_{\rm in},W_{\rm out},b_{\rm out}$; each tensor in units of its RMS at the final checkpoint, so that a displacement means the same change of weights at every checkpoint. \\
Region $U$ & At $u$, $u/2$, and $u/4$: KL $\le0.02$ nats, cross-entropy change $\le0.1$ nats, accuracy change $\le0.02$ (closer and all positions); no tensor displaced by more than $0.5$ RMS. \\
Draws & 128 directions $z_i\sim\mathcal N(0,I)$ per component, shared across checkpoints; the draws are the parameter settings $\theta(\pm rz_i)$, $\theta(\pm rz_i/2)$, $\theta(\pm rz_i/4)$ for $r\in\{0.00625,0.0125,0.025,0.05,0.1,0.15,0.2,0.4\}$ (panel) and the first 64 directions at $r=0.1$ (all components). \\
$A_c$, $B_c$ & $A_c=\operatorname{mean}_i\tfrac12z_i^\top F_cz_i$ by exact Hessian--vector products; $B_c=\operatorname{mean}_i[\bar K_i(r)-4\bar K_i(r/2)]/(\tfrac34r^4)$ with $\bar K_i(r)=\tfrac12[K_c(rz_i)+K_c(-rz_i)]$, which cancels the quadratic term exactly. \\
Spectrum & Four unit-norm Gaussian probes $\times$ 32 Lanczos steps with full reorthogonalization on exact Hessian--vector products of $K_c$; the Ritz pairs $(\theta_{vk},w_{vk})$ give $q_Q(\tau,r)=\tfrac12\tfrac{d_c}{4}\sum_{v,k}w_{vk}\,\tau r^2\theta_{vk}/(1+\tau r^2\theta_{vk})$. \\
Estimates & $q_K$ and $q_Q$ at 16 values of $\tau$ from $10^2$ to $10^6$ by importance weights $\mathbf 1_Ue^{-\tau K_c}$ or $\mathbf 1_Ue^{-\tau Q_c}$ on the 256 draws, where the diagnostics of \citet{vehtari2024psis} pass (Appendix~\ref{app:grid-methods}). Quartic correction at $\tau r^2\le10^4$ from six draws of $\mathcal N(0,(\tau F_c+I/r^2)^{-1})$ per setting by conjugate gradients (panel, 28 checkpoints). \\
Ablation & Output weights set to zero (a head's $W_O$ slice; an MLP's $W_{\rm out}$, $b_{\rm out}$): KL from the clean model (primary) and drop in $S$; output scaled by $1+g$, $g=-0.05$ (local ablation); gain Fisher $F_{gg}=\partial_g^2K_c(0)$ by central differences. \\
\bottomrule
\end{tabularx}
\caption{\textbf{Protocol of the local-geometry measurements} (22,176 component--checkpoint pairs; Appendices~\ref{app:grid-methods} and~\ref{app:spectrum}).}
\label{tab:protocol}
\end{table}

\subsection{A circuit for bracket completion and its development}
\label{sec:circuit}
We identify the circuit with standard MI tools, without geometry. Zero-ablating each of the 144 components, and greedily restoring components into an ablated model until $80\%$ of the clean contrast is recovered (on $k=2$ strings), gives the candidates; attention patterns, probes, direct logit attribution (each component's direct contribution to the output logits), and path patching (replacing a component's output only on the path to one downstream component) assign roles to 26 circuit components in five stages: \C{L0.mlp} and \C{L0H0}--\C{H7} encode the token, \C{L1}--\C{L5.mlp}, \C{L1H1}, \C{L2H1}, and \C{L2H5} update the stack, \C{L7H2}, \C{L7H4}, \C{L7H5}, \C{L9H4}, and \C{L14H1} read the required bracket type, \C{L11.mlp} maps it to a closer, and \C{L13}--\C{L15.mlp} shape the final prediction (Figure~\ref{fig:algorithm-mapping}, Appendix~\ref{app:algorithm-detail}). To test the assignment we fit four variable subspaces on $k=4$ strings, TOKEN (token-dependent outputs of \C{L0.mlp}; 11 dimensions), STACK (an additive code of each stack slot's occupancy and type; 24), POSITION (32), and ANSWER (centered closer unembeddings; 3): projecting the outputs of the 24 writing components onto the subspaces preserves $0.763\pm0.011$ of the clean contrast on held-out strings, and removing STACK from the stack writers lowers held-out closer accuracy from $61.4\%$ to $14.8\%$ ($58.4$--$62.6\%$ for random subspaces of equal rank) and removing ANSWER from the 15 components with direct closer evidence to $39.3\%$ (Appendix~\ref{app:algorithm-detail}).

\input{sections/development_findings}
\input{sections/local_development_findings}
\input{sections/interaction_findings}

%% file: sections/development_findings.tex
\label{sec:development}
\paragraph{Development.}
Across checkpoints, prefix depth is linearly decodable already at initialization (layer-2 probe $R^2=0.864$ for $k=2$), whereas the distance to the most recent unmatched opener is learned ($R^2$ from $0.034$ to $0.684$; Appendix~\ref{app:development-detail}), \label{sec:late}and the eight layer-0 heads act as a redundant group whose joint ablation far exceeds the sum of their single ablations (Appendix~\ref{app:bank-detail}).

%% file: sections/local_development_findings.tex
\subsection{Local geometry of circuit components across training}
\label{sec:local-development}

\paragraph{The measurement stays where predictions are preserved.}
At $r=0.025$ all 256 draws lie inside $U$ for $99.96\%$ of the 5,082 panel pairs, and at $r=0.05$, $0.1$, and $0.2$ for $98.3\%$, $88.5\%$, and $52\%$; SGLD chains instead move each weight by several times its RMS, under standard settings too (Appendix~\ref{app:locality}), so $U$ must be imposed on the draws. We reweight the prior draws by $e^{-\tau K_c}$ (importance weighting), so $K_c$ and $Q_c$ are compared on the same draws.

\paragraph{Inside $U$ the loss is quadratic plus quartic.}
The mean loss agrees with its Fisher approximation to within $0.04\%$ at $r=0.025$ (median), and $B_c$ estimated at $r=0.05$ and $r=0.2$ agrees with its value at $r=0.1$ (median ratios $1.000$ and $0.998$; Appendix~\ref{app:grid-methods}), so $B_c$ is a property of the checkpoint, not the radius. Importance weighting passes its diagnostics where $\tau$ times the spread of $K_c$ under the prior is below about one; there the posterior stays close to the prior, and $q_Q=\tau r^2A_c$ and $q_K-q_Q=\tau r^4B_c$ hold to within $0.4\%$ and $1.1\%$ (medians, $r\le0.1$; Appendices~\ref{app:grid-methods} and~\ref{app:temperature-sweep}).

\begin{figure}[t]
\centering
\includegraphics[width=\linewidth]{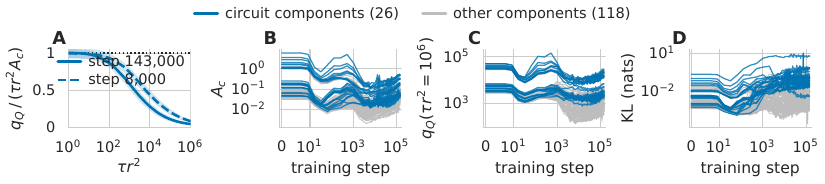}
\caption{\textbf{Local geometry across training.} A: median over components of $q_Q/(\tau r^2A_c)$ at the final checkpoint (solid) and at step 8,000 (dashed), with interquartile bands. B--D: $A_c$, $q_Q$ at $\tau r^2=10^6$, and zero-ablation KL of all 144 components, circuit in blue.}
\label{fig:geometry-development}
\end{figure}

\paragraph{The quadratic part at every temperature.}
Beyond this region the quadratic part remains the spectral sum of Equation~\eqref{eq:soft-rank} and can be evaluated at every $\tau$ and $r$ without drawing. We compute the Fisher spectrum of all 144 components at all 154 checkpoints by stochastic Lanczos quadrature (Appendix~\ref{app:spectrum}). In Figure~\ref{fig:geometry-development}A, $q_Q/(\tau r^2A_c)$ is one for $\tau r^2\lesssim10$, departs from one near $\tau r^2=10^2$, and falls to medians of $0.58$, $0.28$, and $0.04$ at $\tau r^2=10^3$, $10^4$, and $10^6$. The estimate leaves linearity and grows sublinearly, as $(\tau r^2)^{0.6}$ between $10^4$ and $10^6$: it counts the directions whose curvature exceeds $1/(\tau r^2)$, and because the spectrum has no gap the count keeps growing with $\tau$ (Appendix~\ref{app:spectrum}). The quartic correction to this count, evaluated on draws from the Gaussian posterior $\mathcal N(0,(\tau F_c+I/r^2)^{-1})$ (Appendix~\ref{app:quartic-posterior}), equals $\tau r^4B_c$ throughout and is 0.032\% of $q_Q$ at $\tau r^2=10^4$ ($90\%$ of the cells below 0.33\%); inside $U$ the large-temperature estimate is this count.

\paragraph{The Fisher trace and the quartic coefficient develop differently.}
The Fisher trace of every component falls during training, by a median factor of $9.4$, and rises again late (Figure~\ref{fig:geometry-development}B), whereas $B_c/A_c$ peaks between steps 128 and 512 and then decays (Figure~\ref{fig:geometry-stages}). These trajectories are largely shared (median pairwise rank correlation $0.91$ for $A_c$ and $0.93$ for $B_c$, against $0.47$ for the ablation effects), so a component's geometry is a modulation of a model-wide trend.

\subsection{Correspondence between geometry and causal roles}
\label{sec:correspondence}

\begin{figure}[t]
\centering
\includegraphics[width=\linewidth]{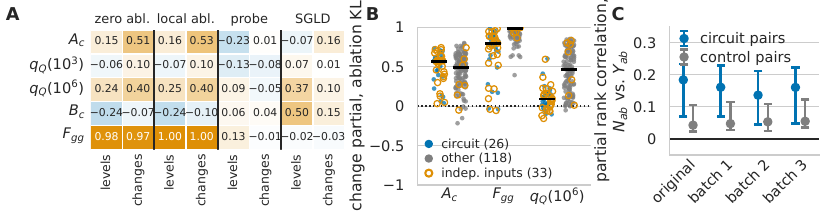}
\caption{\textbf{Correspondence between local geometry and mechanism.} A: median partial rank correlation over the 144 components between each geometric quantity and each target, for levels and adjacent-checkpoint changes. B: per-component change associations with the ablation KL (circuit blue, others gray, medians as ticks; independent batch in orange). C: median partial correlation between $N_{ab}$ and $Y_{ab}$ for 43 circuit and 43 control pairs on four batches, with $95\%$ block-bootstrap intervals.}
\label{fig:correspondence}
\end{figure}

\paragraph{Across training, the Fisher part tracks causal roles (P1).}
At the final checkpoint (Q1) the Metropolis-chain LLC estimate of all 144 components ranks them as zero-ablation does (Spearman $0.64$; Appendix~\ref{app:calibration}). Across training (Q2), for each of the 144 components we correlate the $A_c$ and ablation-KL trajectories directly (Spearman), after regressing both on clean loss, weight scale, and training step (partial rank correlation), and between adjacent-checkpoint changes given the same covariates (block bootstrap, blocks of 16 checkpoints). The medians are $0.59$ (Spearman) and $0.15$ (partial) for levels and $0.51$ for changes, positive for $97\%$ of the components and with a bootstrap interval above zero for $88\%$ (Figure~\ref{fig:correspondence}A--B); the change association is $0.56$ for the 26 circuit components and $0.49$ for the other 118. Much of this co-movement is shared across the model: pairing each component's $A_c$ with the population-median ablation trajectory gives $0.50$, and with a random other component's trajectory $0.26$; the component-specific part, the excess over such pairings, is positive for $82\%$ of the components. The gain Fisher $F_{gg}=\|o_c\|_{\Sigma_c}^2$, the squared norm of the component's output under the downstream Fisher metric of Section~\ref{sec:motivation}, tracks the zero-ablation KL (partial $0.98$ in levels, $0.97$ in changes) and is specific to its component (changes $0.12$ against the population-median trajectory), and, once among the covariates, leaves $A_c$ no association with the ablation ($0.02$ for levels, $0.07$ for changes). Splitting each change of $A_c$ exactly among the downstream, own, and upstream weights (Shapley values over hybrid checkpoints; Appendix~\ref{app:decomposition}) assigns them $0.44$, $0.21$, and $0.35$ of it, and the late rise of $A_c$ comes from the input (median share $1.08$; own weights $-0.23$).

\paragraph{The large-temperature estimate counts rather than sums.}
$q_Q$ at $\tau r^2=10^6$ is specific to its component (changes $0.40$ against its own ablation, $0.13$ against the population median) and tracks ablation changes across the population ($0.47$ for the 118 non-circuit components) but weakly for the circuit components ($0.09$; levels $-0.01$), within heads and within MLPs alike (Appendix~\ref{app:spectrum}). A circuit component's causal growth is growth in magnitude along directions that are already constrained, which the trace registers and the count does not (Figure~\ref{fig:geometry-development}B--D).

\paragraph{The quartic part is what a nonlocal sampler registers (P2).}
With $A_c$ among the covariates, $B_c$ tracks the component's nonlocal SGLD estimate (partial $0.50$; Appendix~\ref{app:methods}), whereas $A_c$ and $F_{gg}$ do not ($-0.07$, $-0.02$). $B_c$ is negatively associated with the component's effect in levels ($-0.24$; changes $-0.07$) and weakly with the opener-distance probe ($0.06$), whose formation the nonlocal estimate tracks ($0.46$; Appendix~\ref{app:locality}) and no local quantity does. The nonlocal estimator registers the growth of the loss beyond its quadratic part, a shared trend that does not mark a component's role.

\paragraph{The layer-0 heads as a group (P3).}
The sum of the eight heads' Fisher traces tracks the sum of their individual ablation effects (adjacent changes $0.71$ on the KL) but not their joint ablation (changes $0.03$; Table~\ref{tab:bank-associations}), and the group's joint LLC estimate lies $5.8\%$ below the sum of its members' (Appendix~\ref{app:calibration}), as Equation~\eqref{eq:joint-geometry} predicts. An independent input batch reproduces the panel's trajectories (rank correlation $0.997$ for $A_c$) and associations ($A_c$ changes $0.55$ on both; Figure~\ref{fig:correspondence}B).

%% file: sections/interaction_findings.tex
\subsection{Pairwise geometry tracks circuit interactions}
\label{sec:interaction-development}
To test P4 we measure 43 circuit pairs (28 among the layer-0 heads; Appendix~\ref{app:interaction-methods}) and 43 matched controls. The target is the nonadditive part of the joint ablation beyond the linearized model, $Y_{ab}=|f(-1,-1)-f(-1,0)-f(0,-1)|$ with $f=K-K_{\mathrm{lin}}$, which has the network's Fisher matrix in $(g_a,g_b)$, so the association lies beyond curvature. Over circuit pairs, the median partial rank correlation of the trajectories, given that Fisher matrix, clean loss, and step, is $0.16$, $0.14$, and $0.16$ on three independent batches, against $0.05$ for control pairs (Figure~\ref{fig:correspondence}C; $B_{ab}$: $0.16$--$0.19$, Appendix~\ref{app:interaction-methods}).

%% file: sections/discussion.tex
\section{Discussion}
\label{sec:discussion}

\paragraph{What the correspondence offers.}
The correspondence gives SLT's developmental measurements a mechanistic reading, provided that they stay where predictions are preserved, and says which quantity answers which question: the Fisher trace follows the model-wide course of causal effect, whereas the gain Fisher and the large-temperature count follow a component's own role and give MI a measure of it that needs no replacement value. Limitations: one model, one synthetic task, one input batch, and fixed coordinates (Appendix~\ref{app:grid-methods}); on natural bracket contexts, where the circuit was not identified, the final-checkpoint correspondence falls from $0.64$ to $0.16$ (Appendix~\ref{app:calibration}).

%% file: sections/conclusion.tex
\section{Conclusion}
MI and SLT describe one object, a circuit's responses and the parameters that preserve them; in Pythia-1B their descriptions correspond term by term: Fisher trace to the trend of causal effect, large-temperature count to a component's own role, quartic part to what a nonlocal sampler registers, pair terms to circuit interactions.

%% file: sections/statements.tex
\subsection*{AI use statement}
Generative AI tools assisted with software implementation, experiment orchestration, literature search, illustration, and manuscript preparation. The authors are responsible for the content, experimental claims, and accompanying software.

\subsection*{Reproducibility statement}
Table~\ref{tab:protocol} lists the protocol of the main measurements. The appendices specify the circuit interventions, sampling objectives, parameter conventions, numerical checks, and derivative estimators. Proofs appear in Appendices~\ref{app:bridge} and~\ref{app:nonquadratic-geometry}; the developmental geometry protocol is in Appendix~\ref{app:grid-methods}, the direct posterior sampling sweep in Appendix~\ref{app:temperature-sweep}, the developmental interaction protocol in Appendix~\ref{app:interaction-methods}, and further experimental details in Appendix~\ref{app:methods}.

\subsection*{Acknowledgments}
This work was partially supported by NSF 2303037, 2312501, 2531008, 2609582, and 2617441; DARPA HR001126CE054; NIH OT2OD038003, U54OD036472, U54DK097771, and U54HG012517; the SRC JUMP 2.0 Center; CDSC Industry Partners; Amazon Research Awards; Amazon; Google; Google Gifts; NEC; Optum; and Snapchat.

%% file: sections/theory_appendix.tex
\section{The local bridge}
\label{app:bridge}
\subsection{Precise local setting}
Fix finitely many input--intervention pairs indexed by $a$, with weights
$w_a>0$ summing to one, and a finite output alphabet of size $M\geq2$.
This includes a fixed evaluation set; a population extension requires
additional uniform regularity and integrability assumptions. Write
$p_a(u)=p_{\theta^\star+E_cu}^{I_a}(\cdot\mid x_a)$ and assume every
coordinate is positive and analytic in a neighborhood of $u=0$.
Define log-odds coordinates
\begin{equation}
 r_{a,j}(u)=\log\frac{p_{a,j}(u)}{p_{a,M}(u)},
 \qquad j=1,\ldots,M-1,
\end{equation}
and stack $G_c(u)=(r_a(u)-r_a(0))_a$. The weights can equivalently
be absorbed into the residual blocks. The nonnegative analytic loss is
\begin{equation}
 K_c(u)=\sum_a w_a D_{\mathrm{KL}}(p_a(0)\|p_a(u)).
\end{equation}
We localize the integral to a sufficiently small bounded neighborhood
with a smooth prior bounded above and below by positive constants.
For $K_c\equiv0$, its partition integral is constant and we assign
exponent $\lambda_c=0$.

\subsection{Proof of Proposition~\ref{prop:bridge}}
For one categorical distribution, set
$A(r)=\log(1+\sum_{j=1}^{M-1}e^{r_j})$. Its KL divergence from the
reference is
\begin{equation}
 A(r)-A(r_0)-\nabla A(r_0)^\top(r-r_0).
\end{equation}
The Hessian at $r_0$ is
$W=\operatorname{diag}(\bar p)-\bar p\bar p^\top$, where
$\bar p=(p_1,\ldots,p_{M-1})$. Positivity of all $M$ probabilities
makes $W$ positive definite. Continuity, a sufficiently small
neighborhood, and finitely many pairs give constants $0<b<B<\infty$
such that
\begin{equation}
 b\|G_c(u)\|^2\leq K_c(u)\leq B\|G_c(u)\|^2.
 \label{eq:kl-norm-equivalence}
\end{equation}
If $G_c$ has constant rank $\varrho_c$ near zero, the analytic constant-rank
theorem gives local source and target coordinates in which it takes
the form $(s_1,\ldots,s_{r_c},0,\ldots,0)$. Coordinate changes have
bounded, nonsingular Jacobians locally, so
$K_c\asymp\sum_{j=1}^{r_c}s_j^2$. Its partition integral scales as
$t^{-r_c/2}$: the constrained coordinates give Gaussian integrals and
the remaining coordinates contribute bounded factors. Thus
$\lambda_c=r_c/2$, with multiplicity one for $r_c>0$.

For $J_a=Dr_a(0)$, Taylor expansion gives
\begin{equation}
 F_c=\sum_a w_aJ_a^\top W_aJ_a,
 \qquad
 K_c(tv)=\tfrac12t^2v^\top F_cv+o(t^2).
\end{equation}
Every $W_a$ is positive definite, hence
$v^\top F_cv=0$ if and only if $J_av=0$ for all $a$, equivalently
$DG_c(0)v=0$. Thus the relevant differential metric is the categorical Fisher
at the reference outputs.\hfill$\square$

\paragraph{Connection to activation interventions.}
For the finite transport in Equation~\eqref{eq:parameter-activation}, replace every operation affected by the selected parameters with its output in the perturbed computation. All remaining structural functions are fixed. Induction over the directed computation graph then gives identical outputs: each replaced node receives its prescribed perturbed value, and every other node receives the same parent values and applies the same function. The patch must include every relevant occurrence of the component. In the transformer experiments, it covers the complete head or MLP output at every sequence position.

Suppose the component parameters affect the rest of the computation
only through its output $h_c(x;u)$, and hold the other component
parameters fixed. For a direction $v$, let
$\delta h_c(x)=D_uh_c(x;0)v$. Applying the activation intervention
$h_c(x)\leftarrow h_c(x;0)+t\delta h_c(x)$ has the same first-order
output derivative as changing the component parameters to $tv$.
This follows from the chain rule for the fixed downstream computation.
The perturbation must include all relevant occurrences of a shared
component, and a single $v$ must generate the perturbations across
inputs. An arbitrary patch, zero-ablation, or mean replacement need
not meet these conditions. Vanishing first-order response also does
not exclude a higher-order response along a straight parameter line.

\paragraph{Interventional versus observational equivalence.}
The theorem applies to $\mathcal I=\{\mathrm{id}\}$ as well as to an
intervention-augmented loss. The former has a weaker zero set: two
implementations can have identical unperturbed outputs and different
responses to internal interventions. The posterior and gain measurements evaluate the identity member at perturbed parameters. The aligned component-gain interventions directly probe the same response map, and the separate MI intervention battery establishes its computational interpretation.

\subsection{Feature constraints and distribution changes}
Suppose a mechanistic hypothesis supplies analytic residuals
$H_j:U\to\mathbb R^{d_j}$ with $H_j(0)=0$ and
$K_c\asymp\sum_j\|H_j\|^2$. If their stacked Jacobian has full row
rank $R=\sum_jd_j$ at zero, it remains full row rank locally. The same
normal-form calculation gives $\lambda_c=R/2$. This is the precise
sense in which independently observable task constraints add.
If feature blocks overlap, additivity need not hold. If a feature is
decodable but unused downstream, its residual need not be bounded by
the output loss, so the needed local equivalence fails.More generally, if $bK_1\leq K_2\leq BK_1$ near the reference, their
local integrability thresholds, and hence real log canonical thresholds (RLCTs), agree. One sufficient
condition for distribution invariance is bounded density ratio:
if $0<b\leq dQ/dP\leq B<\infty$ on the input--intervention pairs,
then nonnegativity of pointwise KL gives
$bK_P\leq K_Q\leq BK_P$. Equality of RLCTs under this condition
does not imply equality of finite-temperature localized estimates.
It also does not apply merely because two datasets concern the same
formal language.

\subsection{Counterexamples and finite measurements}
\paragraph{Constraint dimension is not effect size.}
Let a Gaussian output have unit variance and mean $s\theta$, with
$\theta^\star=1$ and $s>0$. Its reference KL is
$K(\theta)=\tfrac12s^2(\theta-1)^2$, giving $\lambda=1/2$ for every
$s$. Zeroing the parameter has KL effect $s^2/2$, which can be
arbitrarily small or large. The geometric coefficient quantifies a nonnegative constraint; the
sign of a component's contribution is supplied by the chosen causal
readout.

\paragraph{Redundancy and grouping.}
For $K(u,v)=(u+v)^2$, the joint coefficient is $1/2$ although the
two one-dimensional restricted coefficients each equal $1/2$.
Both parameters constrain the same output direction. A nonlinear
example is $f(a,b)=a+b-ab$ at $(a,b)=(1,1)$: setting either parameter
to zero leaves $f=1$, whereas setting both to zero changes $f$ to
zero. In local coordinates $u=a-1,v=b-1$, its squared output error
is $u^2v^2$. Each individual restriction is identically zero, but the
joint partition integral scales as $t^{-1/2}\log t$. Thus joint
geometry and finite joint effects can reveal structure invisible to
both marginal quantities. The measured bank can be compared with these examples through
its joint and marginal geometric and causal responses.

\paragraph{Why mean replacement can conceal a context interaction.}
\label{app:mean-visibility}
Consider a component with output $h_{a,b}(t,x)=a\,m(t)+b\,g(x)$ and a Gaussian task readout $N(h_{a,b},1)$, centered at $(a_\star,b_\star)$. Here $m(t)$ is a position-dependent write and $g(x)$ is a context feature. Suppose $g(x)=0$ on a synthetic distribution $D_s$, with $\E_{D_s}m(t)^2>0$. Its reference loss is
\begin{equation}
 K_{D_s}(a,b)=\tfrac12\E_{D_s}[m(t)^2](a-a_\star)^2,
\end{equation}
so this component has local learning coefficient $1/2$ and a positive LLC estimate. Position-conditioned mean replacement at the reference leaves $h_{a_\star,b_\star}$ unchanged on $D_s$, giving zero causal effect. On a richer distribution $D_n$ with $\E[g(x)\mid t]=0$, $\E[g(x)^2]>0$, and $b_\star\ne0$, the same replacement removes $b_\star g(x)$ and has expected reference KL
\begin{equation}
 \E_{D_n}K_{\rm mean}=\tfrac12 b_\star^2\E_{D_n}[g(x)^2]>0.
\end{equation}
Geometric screening on $D_s$ identifies the component through its constrained average write; interventions on $D_n$ reveal its context-dependent contribution. This example explains how the two measurements can guide discovery at different stages of a circuit investigation.

\paragraph{A regular model can have a singular component slice.}
The residual $H(x,y)=y+x^2$ has rank one everywhere near the origin,
and $K(x,y)=H(x,y)^2$ has joint coefficient $1/2$. Freezing $y=0$
restricts the loss to $x^4$, whose integral scales as $t^{-1/4}$,
so the restricted coefficient is $1/4$ despite zero restricted
Jacobian rank at the origin. A pointwise rank calculation is therefore
insufficient for component-restricted LLCs. The constant-rank
assumption in Proposition~\ref{prop:bridge} is essential.

\paragraph{Finite-time sampling response.}
The all-step average of the SGLD sweep (Appendix~\ref{app:methods}) need not be stationary to have a
geometric interpretation in an ideal quadratic model. Let
$K(u)=\tfrac12u^\top Hu$, $H\succeq0$, start at $u_0=0$, and use
the exact loss gradient in the update
$u_{s+1}=[I-\eta(\tau H+\gamma I)/2]u_s+\sqrt\eta\,\xi_s$,
with independent standard Gaussian $\xi_s$. For an eigenvalue $\mu_i$
of $H$, set $a_i=1-\eta(\tau\mu_i+\gamma)/2$ and assume
$|a_i|<1$. The variance of its coordinate before update $s$ is
$\eta\sum_{j=0}^{s-1}a_i^{2j}$, hence for $S$ pre-update losses
\begin{equation}
 \mathbb E\!\left[\frac{\tau}{S}\sum_{s=0}^{S-1}K(u_s)\right]
 =\frac{\tau\eta}{2}\sum_{i:\,\mu_i>0}
 \frac{\mu_i}{1-a_i^2}
 \left[1-\frac{1-a_i^{2S}}{S(1-a_i^2)}\right].
 \label{eq:finite-time-sgld}
\end{equation}
For $S\geq2$, each positive-curvature direction contributes positively
and null directions contribute zero. When $H=F_c$ is the matched-KL
quadratic Hessian, this connects finite perturbation response to the
task-sensitive subspace in Proposition~\ref{prop:bridge}. The weights depend on curvature, step
size, and run length. This finite-response calculation
assumes an exact quadratic loss, a minimum at the initial point, and
no additional minibatch-gradient noise. It provides an explicit link between finite-time sampling and
task-sensitive directions under these assumptions.

\paragraph{Finite localization retains curvature scale.}
In the exactly quadratic case $L(u)-L(0)=\tfrac12u^\top Hu$, with
$H\succeq0$, let
$q(u)\propto\exp[-aL(u)-\tfrac12\gamma\|u\|^2]$, where
$a=n\beta>0$ and $\gamma>0$. If $\mu_i$ are the eigenvalues of $H$,
the Gaussian calculation gives
\begin{equation}
 a\{\mathbb E_qL-L(0)\}
 =\frac12\sum_i\frac{a\mu_i}{a\mu_i+\gamma}.
 \label{eq:finite-localization}
\end{equation}
It approaches $\operatorname{rank}(H)/2$ as $a\to\infty$ at fixed
$\gamma$, but at finite $a$ it depends on curvature scale and
localization. Shared asymptotic RLCTs can therefore coexist with different
finite localized responses, even in this regular example.

\paragraph{Joint component geometry with a factorized prior.}
The quadratic model also predicts how component measurements combine.
Partition the local coordinates into disjoint components and set
$A=\tau r^2F\succeq0$ in their common prior coordinates. The joint response is
$q_{\rm joint}=\tfrac12\operatorname{tr}[A(I+A)^{-1}]$; freezing the
other components gives
$q_j=\tfrac12\operatorname{tr}[A_{jj}(I+A_{jj})^{-1}]$.
Then
\begin{equation}
 q_{\rm joint}\leq\sum_jq_j.
 \label{eq:joint-geometry}
\end{equation}
Indeed, the Schur-complement identity gives
$(I+A)^{-1}_{jj}\succeq(I+A_{jj})^{-1}$.
Taking traces and using $A(I+A)^{-1}=I-(I+A)^{-1}$ proves the inequality.
Equality holds when the off-diagonal blocks vanish; coupling gives a
strict inequality at finite positive temperature and scale. This
subadditivity measures coupling of local task-sensitive directions.
Joint causal removal can instead have a superadditive effect on the
output, as the explicit redundancy example above shows. The two
measurements thus characterize different aspects of a shared component
group. Our bank experiment uses the product of the eight individual
head priors so that the marginal and joint measurements share this chart.

\paragraph{An off-minimum center can give a negative statistic.}
For $L(u)=\tfrac12(u-b)^2$ centered at $u=0$, the same localized
posterior has variance $(a+\gamma)^{-1}$ and mean $ab/(a+\gamma)$.
Consequently
\begin{equation}
 a\{\mathbb E_qL-L(0)\}
 =\frac{a}{2(a+\gamma)}
 -\frac{a^2b^2(a+2\gamma)}{2(a+\gamma)^2},
\end{equation}
which is negative for sufficiently large $|b|$. The loss is
nonnegative; it is the subtraction of the loss at a nonminimal center
that produces the sign. The signed statistic reflects displacement from a nonminimal
center; the RLCT remains nonnegative.

\paragraph{The experimental objectives differ.}
The developmental SGLD loss is
\begin{equation}
 L(\theta)=\frac{1}{B(T-1)}\sum_{b=1}^B\sum_{t=1}^{T-1}
 m_{bt}\,[-\log p_\theta(x_{b,t+1}\mid x_{b,\leq t})],
 \label{eq:actual-masked-loss}
\end{equation}
where $m_{bt}$ indicates that the target is a closer and $p_\theta$
is normalized over the full vocabulary. Non-closer positions
contribute zero, but remain in the denominator. Hence closer
frequency changes the overall loss scale, which matters to a finite
localized statistic such as \eqref{eq:finite-localization}.
On fixed weighted contexts and for a conditional target distribution $q$,
cross-entropy excess
$L_q(\theta)-L_q(\theta^\star)$ equals the reference KL if
$q=p_{\theta^\star}$. The developmental objective uses deterministic closer labels,
whereas the matched reference loss uses the checkpoint's full output
distribution as its target.

Readout choice also selects different functional relations. Adding a common constant to
all closer logits leaves every closer--closer logit difference
unchanged, while changing their probability relative to non-bracket
tokens and thus changing \eqref{eq:actual-masked-loss}. Conversely,
two models can give the same correct-token probability while
allocating the remaining probability differently among distractors.
The matched experiment fixes the output discrepancy and compares
parameter and activation perturbations under that common readout.

%% file: sections/nonquadratic_theory.tex
\section{Nonquadratic geometry of circuit responses}
\label{app:nonquadratic-geometry}

This section relates the measured geometric quantities to a common circuit response map. All statements fix the input distribution, output readout, component restriction, and parameter coordinates. For a finite empirical distribution, the expectations below are finite weighted sums over scored positions and output symbols.

\subsection{The response map and its quadratic approximation}

Let $p_0(y\mid x)$ be the checkpoint distribution and let $z(u;x)$ be the logits after a component-restricted parameter change $u$. Define the centered logit response
\begin{equation}
 G(u;x,y)=z(u;x,y)-z(0;x,y)
 -\E_{Y\sim p_0(\cdot\mid x)}[z(u;x,Y)-z(0;x,Y)].
 \label{eq:centered-response-map}
\end{equation}
Equip these responses with the inner product
$\langle f,g\rangle_0=\E_{x\sim D,Y\sim p_0}[f(x,Y)g(x,Y)]$.
Subtracting the reference-weighted mean removes the arbitrary additive constant in each logit vector. The same expression is obtained by centering $\log p_u-\log p_0$, because the softmax normalizer is constant across output symbols.

The predictive discrepancy is
\begin{equation}
 K(u)=\E_D\KL(p_0\|p_u)
 =\E_D\log\E_{p_0}\exp G(u).
 \label{eq:response-cumulant-loss}
\end{equation}
Consequently $K(0)=0$, $DK(0)=0$, and
\begin{equation}
 D^2K(0)[v,w]=\langle DG(0)v,DG(0)w\rangle_0.
 \label{eq:response-fisher-pullback}
\end{equation}
The full restricted Fisher matrix is therefore the Gram matrix of the response tangents. This identity includes its off-diagonal entries.

For a fixed pair of component gains $(a,b)$, write $J_a=\partial_aG(0)$, $J_b=\partial_bG(0)$, and $H_{ab}=\partial_a\partial_bG(0)$. Let $P_T$ be the orthogonal projection onto $T=\operatorname{span}\{J_a,J_b\}$. The interaction curvature is
\begin{equation}
 N_{ab}=\|(I-P_T)H_{ab}\|_0^2.
 \label{eq:normal-interaction-curvature}
\end{equation}
It measures the part of the joint response that cannot be expressed as a linear combination of the individual responses. Its numerical evaluation uses the $2\times2$ Fisher Gram matrix and the inner products of $H_{ab}$ with the two tangents.

\begin{proposition}[Interactions beyond the Fisher matrix]
\label{prop:normal-beyond-fisher}
For a model with logits $z_{\mathrm{lin}}(a,b)=z_0+a\,\partial_a z(0)+b\,\partial_b z(0)$, the centered response is exactly $aJ_a+bJ_b$, and $N_{ab}=0$. Models can have identical reference predictions and identical Fisher matrices but different values of $N_{ab}$.
\end{proposition}
\begin{proof}
Centering eliminates the softmax normalizer, so a linear logit map has zero mixed derivative. Conversely, add $ab\,h(x,y)$ to $z_{\mathrm{lin}}$, where $h$ is centered and orthogonal to $J_a,J_b$. This preserves the reference predictions and both first derivatives, hence the entire Fisher Gram matrix. The normal curvature, the interaction curvature $N_{ab}$ of Section~\ref{sec:motivation}, becomes $\|h\|_0^2$, which can take any nonnegative value.
\end{proof}

This construction separates nonlinear computation from the nonlinearity of the categorical readout. The linearized-logit control retains the exact softmax and the complete response tangents. Its predictions can exhibit nonlinear KL effects even though its centered mixed response vanishes.

\paragraph{Functional gain coordinates.}
Output gains specify interventions in units of a component's own write. An internal reparameterization that preserves each component's input--output function and commutes with scalar output multiplication preserves the entire gain-response map. Induction along the computation graph then gives the same $G(a,b)$, hence the same $F$, $N_{ab}$, and mixed-volume coefficient (the mixed quartic coefficient $B_{ab}$ of Section~\ref{sec:motivation}). This includes inverse changes of basis in a head's value and output maps. The gain coordinates thus fix the comparison at the level of the component operation.

\paragraph{An exactly solvable circuit interaction.}
Consider eight output symbols indexed by $(x,y,z)\in\{-1,1\}^3$, with logits
$\ell_{xyz}(a,b)=ax+by+\kappa ab z$ and uniform reference probabilities at $(a,b)=(0,0)$.
The first two terms transmit individual component responses; the third represents their multiplicative interaction. Direct normalization gives
\begin{equation}
 K_\kappa(a,b)=\log\cosh a+\log\cosh b+\log\cosh(\kappa ab).
 \label{eq:solvable-interaction}
\end{equation}
Every member of this family has the same restricted Fisher matrix $F=I_2$.
Its mixed response $\kappa z$ is orthogonal to the individual responses $x,y$, so $N_{ab}=\kappa^2$.
The linearized-logit loss is $K_{\mathrm{lin}}=\log\cosh a+\log\cosh b$.
Consequently $B_{ab}=\kappa^2/2$, while the absolute nonlinear joint zero-ablation interaction is
\begin{equation}
 \left|f(-1,-1)-f(-1,0)-f(0,-1)\right|
 =\log\cosh\kappa
 =\log\cosh\sqrt{N_{ab}},
 \qquad f=K_\kappa-K_{\mathrm{lin}}.
\end{equation}
Thus local curvature, the mixed loss-volume correction, and the finite causal interaction vary together despite an identical full Fisher matrix. All losses are regular at the origin, with learning coefficient one; the varying geometric information resides in their finite-volume corrections. For every finite $\kappa$, the gain radius can be chosen small enough that the entire measurement square has arbitrarily small predictive KL.

\subsection{Finite corrections to local loss volume}

For an untruncated Gaussian prior of covariance $r^2I$, the full quadratic response has the closed form
\begin{equation}
 q_Q(\tau,r)=\frac12\sum_i\frac{\tau r^2\mu_i}{1+\tau r^2\mu_i},
 \label{eq:finite-bridge}
\end{equation}
where $\mu_i$ are the restricted Fisher eigenvalues. Comparisons with an additional neighborhood $U$ apply the same restriction to the actual and quadratic targets.

Let $U$ be a fixed neighborhood containing an open ball around zero. Assume that $K$ is analytic on a neighborhood of its closure and that the local integrals are finite. With a Gaussian prior of covariance $r^2I$, define
\begin{equation}
 Z_K(\tau,r)=\E_{u\sim N(0,r^2I)}[\mathbf1_U(u)e^{-\tau K(u)}],
 \quad \mathcal F_K=-\log Z_K,
 \quad q_K=\tau\,\partial_\tau\mathcal F_K.
 \label{eq:nonquadratic-partition}
\end{equation}
Normalizing the prior inside $U$ changes $\mathcal F_K$ by a term independent of $\tau$, and leaves $q_K$ unchanged. The energy fluctuation is
\begin{equation}
 C_K=\tau^2\operatorname{Var}_{\pi_K}[K],
 \qquad \partial_{\log\tau}q_K=q_K-C_K.
 \label{eq:energy-fluctuation-identity}
\end{equation}
For a local volume law with exponent $\lambda$ and multiplicity $m$, the corresponding large-$\tau$ expansion gives
$q_K=\lambda-(m-1)/\log\tau+o(1/\log\tau)$.
Finite responses additionally contain information about curvature and higher response derivatives.

Write
\begin{equation}
 K(rz)=r^2K_2(z)+r^3K_3(z)+r^4K_4(z)+O(r^5),
 \qquad K_2(z)=\tfrac12z^\top Fz.
\end{equation}
For a standard Gaussian $z$, set
$A=\tfrac12\operatorname{tr}F$ and
$B=\E K_4(z)=\tfrac18\sum_{i,j}\partial_i^2\partial_j^2K(0)$.

\begin{proposition}[The first nonquadratic volume correction]
\label{prop:quartic-gibbs-correction}
At fixed $\tau$, as $r\to0$,
\begin{align}
 \log Z_K
 &=-\tau Ar^2+r^4\left[-\tau B+\frac{\tau^2}{4}\operatorname{tr}(F^2)\right]+O(r^6),\\
 q_K-q_Q&=\tau Br^4+O(r^6),
 \qquad Q(u)=\tfrac12u^\top Fu.
 \label{eq:quartic-gibbs-correction}
\end{align}
The two partition functions use the same neighborhood and prior.
\end{proposition}
\begin{proof}
The Gaussian expectation of every odd homogeneous polynomial vanishes. Thus
$\E K(rz)=Ar^2+Br^4+O(r^6)$ and
$\operatorname{Var}K(rz)=\tfrac12r^4\operatorname{tr}(F^2)+O(r^6)$.
The cumulant expansion of $\log\E e^{-\tau K}$ gives the first formula. Differentiating with respect to $\tau$ and subtracting the quadratic expression gives the second. Because $U$ contains a fixed ball, its boundary contributes an exponentially small correction in this small-$r$ expansion.
\end{proof}

An antithetic radius contrast estimates $B$ without subtracting an estimated Fisher trace:
\begin{equation}
 B_r=\frac{\E K(rz)-4\E K(rz/2)}{(3/4)r^4}=B+O(r^2).
\end{equation}
Using identical directions at both radii cancels every quadratic form sample by sample. For example, $K_c(u)=u^2/2+cu^4$ has the same Fisher at the origin for every $c$, while its leading finite-response correction is $3c\tau r^4$.

The comparison at fixed $s=\tau r^2$ has a different expansion. Let $\mu_s$ be the Gaussian measure with density proportional to $e^{-sK_2(z)}$ relative to $N(0,I)$. Then
\begin{equation}
 \mathcal F_K-\mathcal F_Q
 =r^2\left[s\E_{\mu_s}K_4-\frac{s^2}{2}\E_{\mu_s}K_3^2\right]+O(r^4),
\end{equation}
and
\begin{equation}
 q_K-q_Q
 =r^2s\,\partial_s\left[s\E_{\mu_s}K_4-\frac{s^2}{2}\E_{\mu_s}K_3^2\right]+O(r^4).
 \label{eq:fixed-product-correction}
\end{equation}
Both cubic and quartic response terms contribute in this regime. Halving $r$ while multiplying $\tau$ by four therefore leaves the Gaussian reference unchanged and reduces the leading nonquadratic correction by a factor of four, provided that boundary effects are negligible.

\subsection{Circuit coupling in the partition function}

For a two-component loss $L(a,b)$, define the symmetric rectangular contrast
\begin{equation}
 \mathcal D_rL=\frac14\sum_{s,t\in\{-1,1\}}L(sr,tr)
 -\frac12\sum_sL(sr,0)-\frac12\sum_tL(0,tr)+L(0,0).
\end{equation}
Every polynomial of total degree at most three, including the off-diagonal Fisher term, is canceled. Taylor expansion gives
\begin{equation}
 \frac{\mathcal D_rK-\mathcal D_rK_{\mathrm{lin}}}{r^4}
 =\frac14\partial_a^2\partial_b^2(K-K_{\mathrm{lin}})(0)+O(r^2).
 \label{eq:mixed-volume-coefficient}
\end{equation}
Here $K_{\mathrm{lin}}$ is the KL loss of the exact-softmax linearized-logit control. Denote the limiting coefficient by $B_{ab}$.

Let $\mathcal F_{ab}$ be the free energy with both gains varied, and let $\mathcal F_a,\mathcal F_b$ be the corresponding one-gain free energies, using matched Gaussian scales. Define their nonadditive part
$\mathcal J_K=\mathcal F_{ab}-\mathcal F_a-\mathcal F_b$.
Since $K$ and $K_{\mathrm{lin}}$ have the same Hessian, Proposition~\ref{prop:quartic-gibbs-correction} gives
\begin{equation}
 \mathcal J_K-\mathcal J_{K_{\mathrm{lin}}}
 =\tau r^4 B_{ab}+O(r^6)
 \label{eq:interaction-free-energy-correction}
\end{equation}
at fixed $\tau$. Thus the rectangular contrast measures a mixed local-volume correction beyond both the Fisher approximation and the nonlinear categorical readout of a linear response.

The normal curvature in Equation~\eqref{eq:normal-interaction-curvature} resolves an identifiable part of this higher-order geometry. Expanding Equation~\eqref{eq:response-cumulant-loss}, the term $abH_{ab}$ contributes
$\tfrac12a^2b^2\|H_{ab}\|_0^2$ to the quadratic cumulant. Higher response derivatives and categorical third cumulants contribute additional terms to the complete mixed-volume coefficient. Measuring both quantities separates a positive nonlinear-response contribution from their combined effect on loss volume.

\subsection{Correspondence with causal interaction tests}

A parameter change to a component can be transported exactly into the clean network by patching its complete altered activation tensor. For several components, use their activations from the joint perturbed forward pass; induction along the computation graph then reproduces the same downstream output. Parameter geometry and matched activation interventions consequently evaluate the same response map.

For a scalar circuit readout $S(a,b)$, the finite joint response to zero-gain interventions satisfies the exact identity
\begin{equation}
 S(-1,-1)-S(-1,0)-S(0,-1)+S(0,0)
 =\int_{-1}^{0}\!\int_{-1}^{0}\partial_a\partial_b S(a,b)\,da\,db.
 \label{eq:causal-interaction-integral}
\end{equation}
For a pairwise logit contrast, $\partial_a\partial_bS$ is a linear readout of the mixed centered response $H_{ab}$. If this derivative differs from its value at zero by at most $\epsilon$ throughout the integration square, the finite causal interaction differs from the local mixed derivative by at most $\epsilon$. This supplies a sufficient condition for agreement. Changes in the direction or sign of the interaction away from the reference can instead produce cancellation in the finite intervention. A norm of the local response also aggregates output directions that a particular scalar contrast may weight differently.

The KL interaction used in the experiments is governed by a corresponding higher-order condition. Write $H_\perp=(I-P_T)H_{ab}$ and decompose
\begin{equation}
 f(a,b)=K(a,b)-K_{\mathrm{lin}}(a,b)
       =\tfrac12a^2b^2\|H_\perp\|_0^2+E(a,b).
\end{equation}
Here $E$ collects tangent mixed responses, the remaining higher response derivatives, and categorical cumulants beyond the quadratic term. The displayed contribution follows by expanding
$\tfrac12\|aJ_a+bJ_b+abH_\perp\|_0^2$ and using orthogonality.
If $|E|\leq\epsilon$ at the three nonzero intervention corners, then
\begin{equation}
 \left|\,\left|f(-1,-1)-f(-1,0)-f(0,-1)\right|-\tfrac12N_{ab}\right|
 \leq 3\epsilon.
\end{equation}
This gives a sufficient condition for the specific geometric and causal quantities to agree. Interactions can depart from this relation when the remainder dominates or changes sign across training. Equation~\eqref{eq:solvable-interaction} gives an exact monotone correspondence even when the quadratic-cumulant approximation to the finite intervention is inaccurate.

Input and intervention choices specify the response being tested. If two input distributions have density ratio bounded above and below by positive constants, their KL losses are locally equivalent and have the same learning coefficient. Finite responses can still differ through their weighting of the same constraints. A change of support can add or remove constraints. Likewise, position-mean replacement preserves a mean component write, whereas a parameter intervention can change that write. These distinctions explain why agreement is tested with matched inputs, output discrepancies, and component interventions.

%% file: sections/interaction_methods.tex
\section{Developmental interaction measurements}
\label{app:interaction-methods}

\subsection{Fixed circuit pairs, inputs, and checkpoints}
The interaction experiment contains 43 circuit-based pairs and 43 control pairs. The circuit set contains every unordered pair of the eight first-layer heads, together with
\begin{align*}
 &(\C{L0H0},\C{L7H4}),\ (\C{L0H2},\C{L7H4}),\\
 &(\C{L0.mlp},\C{L2.mlp}),\ (\C{L1.mlp},\C{L2.mlp}),\\
 &(\C{L2.mlp},\C{L3.mlp}),\ (\C{L3.mlp},\C{L4.mlp}),\ (\C{L4.mlp},\C{L5.mlp}),\\
 &(\C{L2.mlp},\C{L7H4}),\ (\C{L3.mlp},\C{L7H4}),\\
 &(\C{L4.mlp},\C{L7H4}),\ (\C{L5.mlp},\C{L7H4}),\\
 &(\C{L7H4},\C{L11.mlp}),\ (\C{L7H5},\C{L11.mlp}),\\
 &(\C{L11.mlp},\C{L14H1}),\ (\C{L7H4},\C{L14H1}).
\end{align*}
Control pairs match component types and layer separation, using seed 20260923 and excluding duplicate pairs. Selection follows the mechanistic account; the confirmation outcomes are evaluated for every selected pair.

All 154 released checkpoints are measured. The calibration checkpoints are 0, 512, 2,000, 8,000, 32,000, and 143,000. The other 148 checkpoints define the confirmation set. The primary task is $k=4$, depth bound 16, sequence length 128, and opening probability $0.45$. The original seed-0 batch contains 128 strings and 5,475 closer targets. Independent seeds 20261001, 20261002, and 20261003 each generate 128 strings, with 5,553, 5,482, and 5,421 closer targets. Geometry and interventions use identical examples and masks within each batch. The complete replicated grid contains $4\times154\times86=52{,}976$ pair--checkpoint--batch measurements.

\subsection{Parameterization and numerical measurements}
For head $h$, gain $a$ replaces its output-projection block by $(1+a)W_O^h$, preserving the shared attention output bias. For an MLP, the gain multiplies both $W_{\mathrm{out}}$ and $b_{\mathrm{out}}$. Gain $-1$ therefore implements the corresponding zero-output intervention. Joint gains are applied in the same forward pass, including their downstream effect on other components. The gains are specified in component-write coordinates at each checkpoint.

We load the fixed converted float32 checkpoint values, then promote evaluation arithmetic to float64. The reference KL uses the full vocabulary at closer positions. Separate functional checks use all valid next-token positions. Centered response tangents are computed by symmetric differences at gains $0.0025$ and $0.00125$, with Richardson extrapolation. The resulting $2\times2$ Gram matrix supplies the full restricted Fisher control. The normal projection uses a relative pseudoinverse cutoff $10^{-8}$; the smallest observed eigenvalue ratio is $0.00329$, so neither tangent direction is discarded in any cell.

At each radius $r\in\{0.025,0.05\}$, we evaluate the four signed corners $(\pm r,\pm r)$ and four axis points $(\pm r,0),(0,\pm r)$. Centering the mixed log-probability difference removes the softmax normalizer exactly. Its normal squared norm estimates $N_{ab}$ with second-order radius error. For $f=K-K_{\mathrm{lin}}$, the mixed-volume estimate is
\begin{equation}
 B_{ab,r}
 =\frac{\frac14\sum_{s,t=\pm1}f(sr,tr)
             -\frac12\sum_{s=\pm1}f(sr,0)
             -\frac12\sum_{t=\pm1}f(0,tr)}{r^4}.
\end{equation}
Here $f(0,0)=0$. The linearized-logit control retains the full individual response tangents and exact categorical normalization. The primary normal-curvature statistic and secondary volume contrast are specified before confirmation analysis.

Every local state is screened for mean KL at most $0.02$ nats, absolute cross-entropy change at most $0.1$ nats, and absolute full-vocabulary accuracy change at most $0.02$. Both closer positions and all valid positions must pass. All 52,976 pair--checkpoint--batch cells pass at the primary radius. Maximum all-position KL across the four batches is $0.01427$, $0.01415$, $0.01523$, and $0.01389$ nats. Halving the radius gives normal-curvature rank correlations $0.99987$, $0.99966$, $0.99951$, and $0.99975$. For the absolute mixed-volume coefficient the corresponding correlations are $0.99266$, $0.99088$, $0.99422$, and $0.99212$.

Repeated-baseline checks require a maximum log-probability error below $10^{-10}$; two later-layer cells that exceeded it with cached prefixes were evaluated from the frozen embedding output instead, with the same parameters, inputs, gains, and arithmetic.

\begin{figure}[t]
\centering
\includegraphics[width=\linewidth]{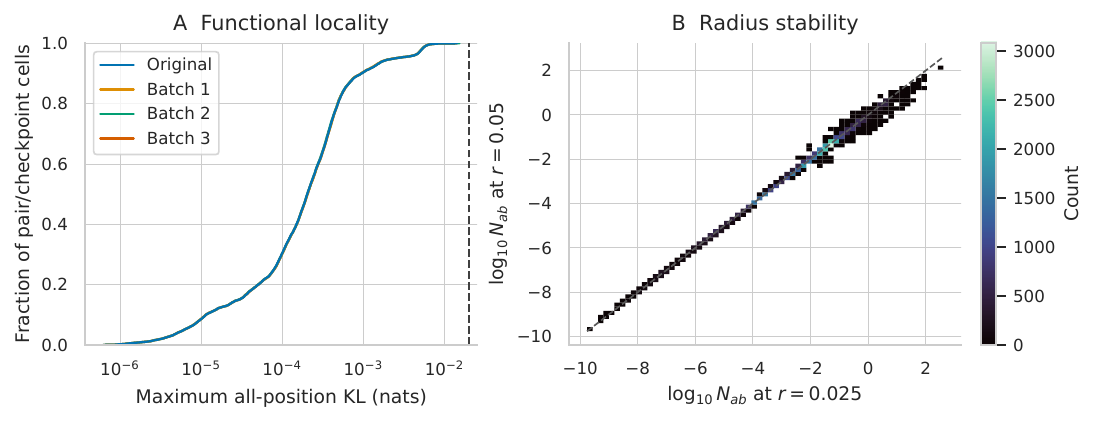}
\caption{\textbf{Locality and radius stability of the interaction measurement.} A: empirical distributions (the four batches coincide at this scale) of the largest all-position KL across each pair's eight local configurations, including all 154 checkpoints. The dashed line marks the fixed $0.02$-nat threshold. B: normal-curvature estimates at the two radii over the complete four-batch grid.}
\label{fig:interaction-locality}
\end{figure}

For all 43 prespecified circuit pairs, the dense-path check evaluates 16 equally spaced nonzero magnitudes along each of the eight signed local directions, from the reference to radius $0.05$. All $43\times154\times4\times128=3{,}390{,}464$ states are present and lie inside $U$ at closer and at all positions. Maximum all-position KL, loss change, and accuracy change are $0.01353$ nats, $0.06042$ nats, and $0.01819$; the corresponding closer-position maxima are $0.01013$, $0.04471$, and $0.01587$. Midpoints and endpoints reproduce the stencil measurements to within $10^{-9}$ in KL and cross-entropy. Checkpoint, pair, direction, and gain identifiers verify full coverage of this fixed circuit set. These evaluations provide a finite-grid check of the sampled paths (Figure~\ref{fig:interaction-locality}).

\subsection{Temporal comparisons and dependence}
For each pair, we compute a temporal Spearman correlation across the 148 confirmation checkpoints. Partial-rank correlations regress both ranked measurements on an intercept, the smaller and larger Fisher diagonal entries, the absolute cross entry, clean closer loss, and training step. An additional control retains the signed cross entry; all adjusted primary results reported in the main text include it. The reported population summary is the median over pairs, with circuit and comparison strata retained separately. These controls quantify incremental rank association under the specified regression.

Adjacent-change comparisons difference the ordered checkpoint series before ranking. An interval is retained only when both neighboring released checkpoints belong to the confirmation set. Resampling never creates a difference across artificial block boundaries. The controls are differenced in the same way. Checkpoint IDs, rather than file order, align all measurements.

\begin{figure}[t]
\centering
\includegraphics[width=\linewidth]{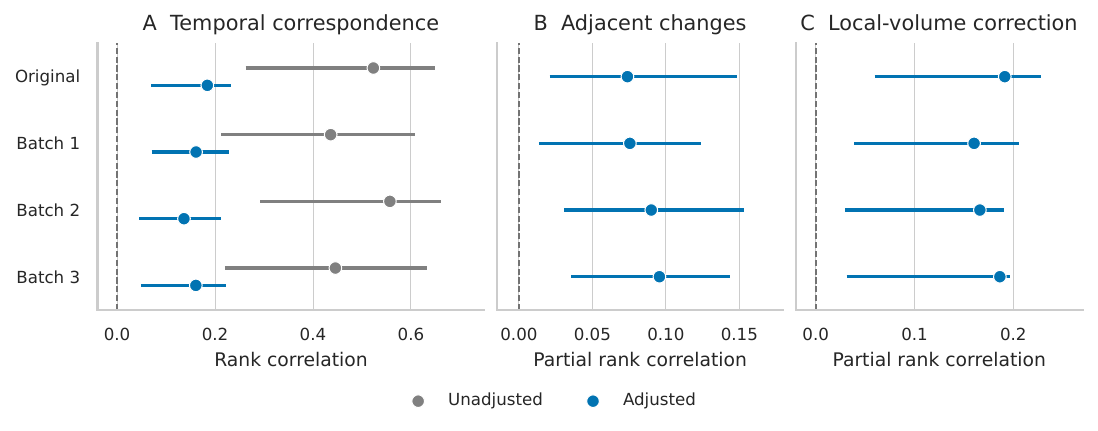}
\caption{\textbf{Pairwise geometry tracks circuit interactions beyond Fisher geometry.} Medians over all 43 circuit pairs. A: temporal associations between local normal curvature and the causal interaction. B: associations between adjacent checkpoint changes. C: associations for the absolute mixed local-volume correction. Adjusted estimates control the restricted Fisher terms, clean closer loss, and training step (their differences in B). Bars are $95\%$ shared temporal-block intervals with block length 16. Three independent input batches replicate the original-batch result.}
\label{fig:interaction-confirmation}
\end{figure}

We draw 1,000 shared temporal-block replicates at each block length 8, 16, and 32. Every pair receives the same sampled checkpoint blocks, preserving dependence through shared inputs and components. The $95\%$ percentile intervals in Figure~\ref{fig:interaction-confirmation} use length 16; the complete results retain all three lengths. A separate deletion analysis removes every pair containing one component, then recomputes the median. These intervals describe stability within the observed training trajectory. The independent input batches replicate the measurement on fresh data from the same task distribution (Table~\ref{tab:interaction-replication}).

\begin{table}[t]
\centering\small
\begin{tabular}{lrrrr}
\toprule
& Original & Batch 1 & Batch 2 & Batch 3\\
\midrule
All pairs: temporal $\rho$ & .534 & .491 & .563 & .482\\
All pairs: adjusted $\rho$ & .086 & .098 & .097 & .112\\
Circuit pairs: temporal $\rho$ & .523 & .436 & .557 & .446\\
Circuit pairs: adjusted $\rho$ & .184 & .161 & .137 & .161\\
Control pairs: adjusted $\rho$ & .043 & .048 & .053 & .055\\
Circuit pairs: adjusted changes & .074 & .076 & .090 & .096\\
Circuit pairs: adjusted $|B_{ab,r}|$ & .191 & .160 & .166 & .186\\
\bottomrule
\end{tabular}
\caption{Complete temporal summaries at $r=0.05$. Each entry is the median within-pair association with the matched nonlinear causal interaction. Adjusted values include the signed cross-Fisher control, clean loss, and time. All pairs have complete trajectories over the confirmation checkpoints.}
\label{tab:interaction-replication}
\end{table}

The smallest lower endpoint among all three block lengths and all four batches is $0.030$ for the circuit-pair primary adjusted association, $0.009$ for adjusted adjacent changes, and $0.015$ for the adjusted absolute mixed-volume coefficient. At half radius, the independent-batch median adjusted volume associations are $0.181$, $0.154$, and $0.180$.

A float32 comparison across tasks measures the same 86 pairs on both $k=2$ and $k=4$; on $k=2$, 83 pairs lie inside $U$ at closer positions at every confirmation checkpoint. The four-batch float64 experiment is the primary independent-input confirmation and adds the all-position locality checks.

\begin{figure}[t]
\centering
\includegraphics[width=\linewidth]{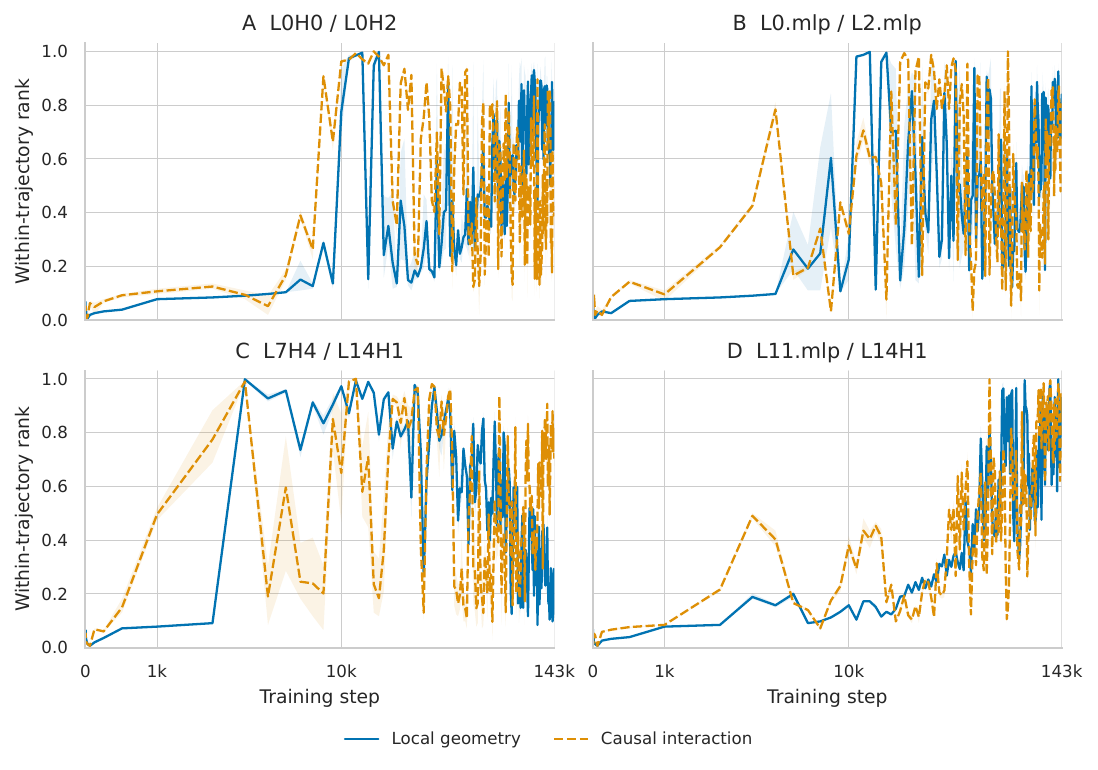}
\caption{\textbf{Circuit-role examples across all 154 checkpoints.} A: first-layer head bank. B: an early MLP pathway. C: retrieval and readout heads. D: an MLP/readout-head pair. Curves show within-trajectory ranks, averaged over four input batches; shading spans the batch values. The statistics use unsmoothed individual-batch measurements. The examples cover distinct circuit roles; Figures~\ref{fig:interaction-heatmap-circuit} and~\ref{fig:interaction-heatmap-controls} show every measured pair.}
\label{fig:interaction-trajectories}
\end{figure}

The prespecified signed closer-contrast comparison evaluates the local mixed derivative of the correct-versus-incorrect closer contrast against its joint zero-ablation interaction. Median unadjusted temporal correlations across circuit pairs are $0.453$ on the original batch and $0.409$, $0.360$, and $0.427$ on the independent batches. These values retain the signed task readout; the primary KL comparison aggregates the predictive response (Figure~\ref{fig:interaction-trajectories}).

\begin{figure}[p]
\centering
\includegraphics[width=\linewidth,height=.86\textheight,keepaspectratio]{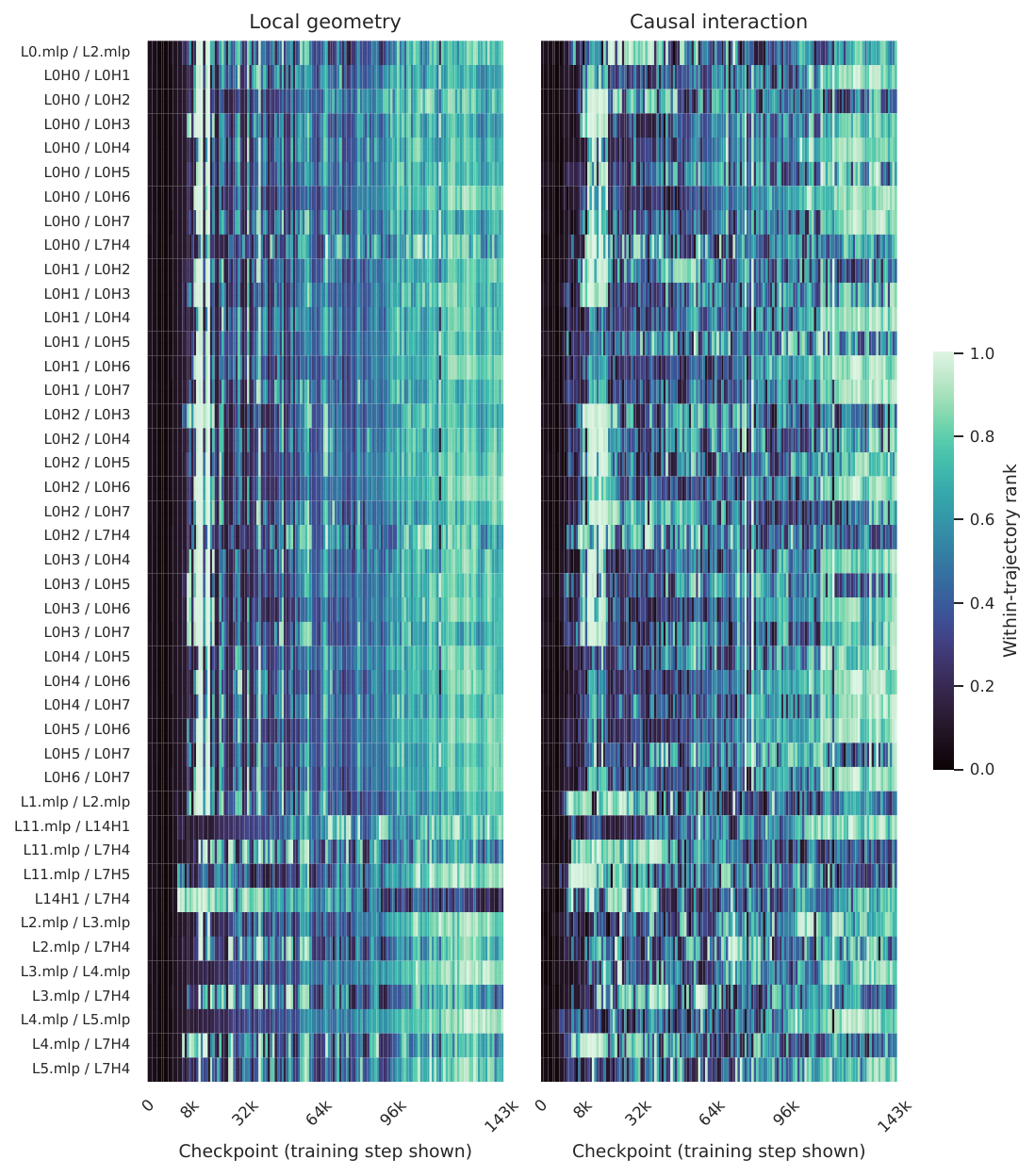}
\caption{\textbf{Complete circuit-pair trajectories.} Each measurement is ranked within its 154-checkpoint trajectory, then averaged across four input batches. Rows include every one of the 43 circuit pairs. Columns are equally spaced released checkpoints, with training-step labels.}
\label{fig:interaction-heatmap-circuit}
\end{figure}
\begin{figure}[p]
\centering
\includegraphics[width=\linewidth,height=.86\textheight,keepaspectratio]{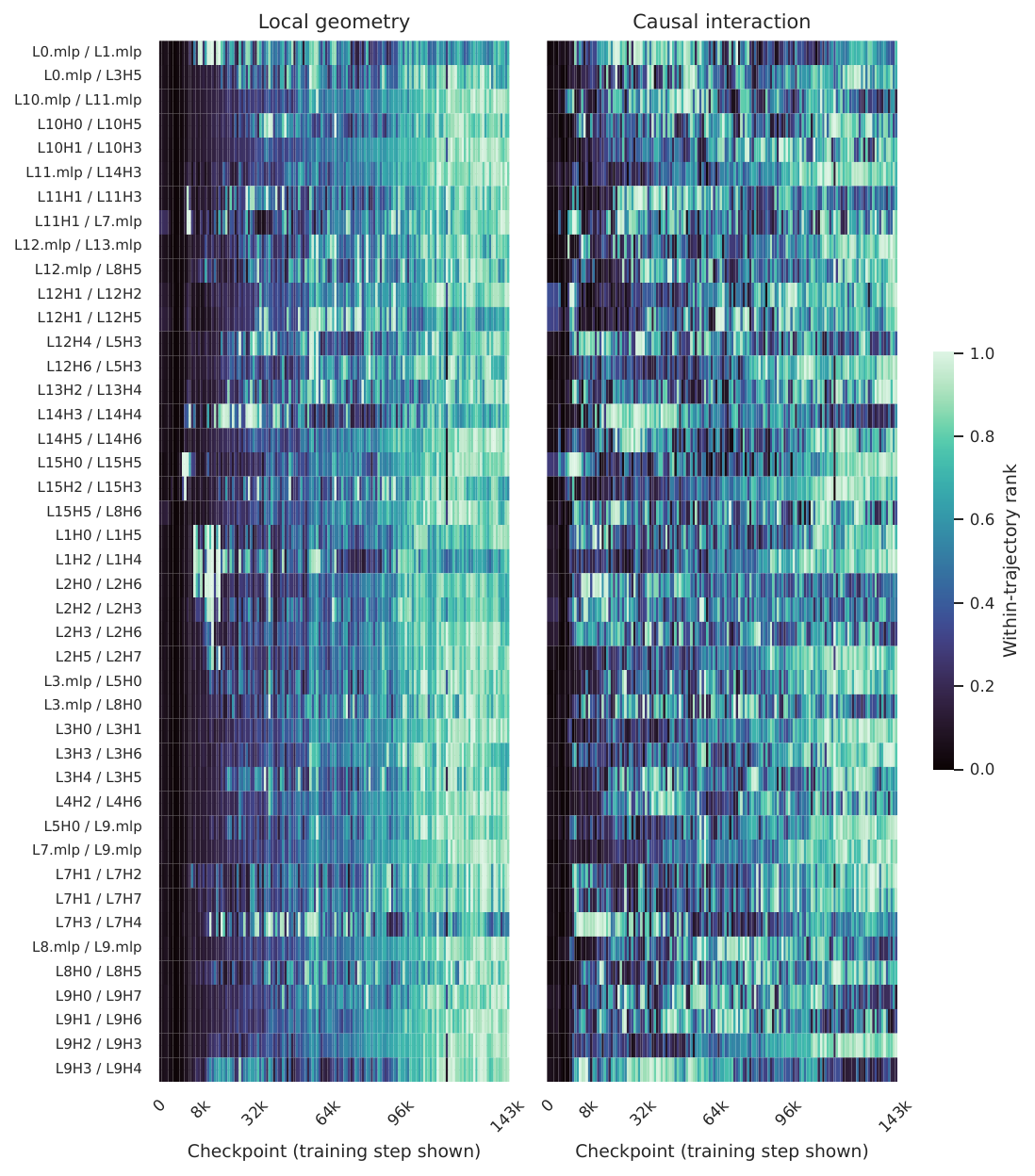}
\caption{\textbf{Complete control-pair trajectories.} The same display for all 43 pairs matched by component type and layer separation.}
\label{fig:interaction-heatmap-controls}
\end{figure}

\paragraph{Robustness of the temporal association.}
For the 43 circuit pairs, the Fisher-adjusted partial temporal association between $N_{ab}$ and $Y_{ab}$ has shared temporal-block bootstrap intervals that remain positive at block lengths 8, 16, and 32 on all three independent batches, and removing every pair incident to any one component leaves a positive median association (lowest value $0.10$).

\paragraph{Locality of the dense paths.}
All 3,390,464 dense path states of the 43 circuit pairs lie inside the region $U$ at every checkpoint on every batch (maximum all-position KL $0.0136$ nats), as does every primary-radius stencil; halving the gain radius changes the normal curvature by a median $0.25$--$0.27\%$ across the three batches.

%% file: sections/grid_methods.tex
\section{Local geometry across training: protocol}
\label{app:grid-methods}

This appendix specifies the local-geometry measurement reported in Section~\ref{sec:local-development} and summarized in Table~\ref{tab:protocol}. Every quantity is computed from the KL loss~\eqref{eq:component-kl} of one component at one checkpoint; the estimator, its Fisher control, and the region $U$ were fixed before any developmental comparison.

\paragraph{Components, checkpoints, and inputs.}
The radius--temperature grid covers a panel of 33 components: the 26 circuit components and seven control components sampled uniformly from the others (seed 20260926), namely the circuit components \C{L0.mlp}, \C{L0H0}--\C{H7}, \C{L1}--\C{L5.mlp}, \C{L1H1}, \C{L2H1}, \C{L2H5}, \C{L7H2}, \C{L7H4}, \C{L7H5}, \C{L9H4}, \C{L14H1}, \C{L11.mlp}, and \C{L13}--\C{L15.mlp} and the controls \C{L2H3}, \C{L6.mlp}, \C{L11H0}, \C{L12.mlp}, \C{L13H7}, \C{L14H0}, and \C{L14H7}; the spectrum measurement of Appendix~\ref{app:spectrum} covers all 144 components. All 154 released checkpoints are measured. The inputs are the fixed $k=4$ developmental batch: 128 strings of length 128 with 11,334 valid next-token positions, 5,475 of which are closer positions. The reference distribution at every scored position is the checkpoint's own full-vocabulary next-token distribution.

\paragraph{Parameters and coordinates.}
A head restriction varies its slices of $W_Q,W_K,W_V,W_O,b_Q,b_K,b_V$ ($d_c=2{,}097{,}920$); an MLP restriction varies $W_{\mathrm{in}},b_{\mathrm{in}},W_{\mathrm{out}},b_{\mathrm{out}}$ ($d_c=33{,}564{,}672$). Coordinates rescale each selected tensor by a fixed scale (Equation~\eqref{eq:gibbs-rms-chart}). The primary coordinates use the final-checkpoint RMS scales at every checkpoint, so that a coordinate displacement means the same change of weights throughout training. A second set of coordinates rescales by the current checkpoint's RMS values; it is measured for the seven components of the temperature sweep (Appendix~\ref{app:temperature-sweep}) as a coordinate control.

\paragraph{Directions and radii.}
For each component we draw 128 independent standard Gaussian directions $z_i$ in these coordinates, shared across all checkpoints and both coordinate choices through a component-indexed seed. Each direction is evaluated with both signs at the twelve amplitudes $\{r,\;r/2,\;r/4\}$ generated by the eight radii $r\in\{0.00625,0.0125,0.025,0.05,0.1,0.15,0.2,0.4\}$; identical amplitudes are evaluated once. The endpoint closer KL $K(\pm rz_i)$, its all-position counterpart, cross-entropy, and full-vocabulary accuracy are recorded at every amplitude. The full restricted Fisher quadratic $Q(rz_i)=\tfrac12r^2z_i^\top F_cz_i$ is evaluated once per direction by an exact Hessian-vector product of the closer reference KL, computed by automatic differentiation through the same forward pass; it is rescaled exactly to every radius. Model arithmetic is float64 after loading the identical float32 checkpoint values. Independent CPU tests verify the Hessian-vector product against full-forward directional derivatives, the exact $r^2$ scaling of $Q$, the small-radius agreement of $K$ and $Q$, and parameter-zero versus activation-zero parity for heads and MLPs.

\paragraph{The region $U$.}
A sample $u=\pm rz_i$ lies inside $U$ if, at $u$, $u/2$, and $u/4$, both at closer positions and at all valid positions, $\mathrm{KL}\leq0.02$ nats, $|\Delta\mathrm{CE}|\leq0.1$ nats, and $|\Delta\mathrm{accuracy}|\leq0.02$, and every tensor's unit-RMS displacement is at most $0.5$. A stricter $0.2$-displacement mask is retained. The same indicator enters the actual and the quadratic targets, so both LLC estimates are defined on the same region.

\paragraph{Radial statistics.}
Writing $\bar K_i(r)=\tfrac12[K(rz_i)+K(-rz_i)]$ for the average over both signs (antithetic sampling), which cancels every odd-order term, we report at each radius the mean profile $\E_i\bar K_i(r)$, the Fisher profile $\E_iQ(rz_i)$, their relative excess, and the quartic coefficient
\begin{equation}
 \widehat B_i(r)=\frac{\bar K_i(r)-4\bar K_i(r/2)}{(3/4)\,r^4},
 \qquad \widehat B(r)=\E_i\widehat B_i(r),
\end{equation}
which cancels the quadratic term direction by direction and estimates $B=\E K_4(z)$ of Proposition~\ref{prop:quartic-gibbs-correction} with an $O(r^2)$ bias. Agreement of $\widehat B(r)$ across radii certifies the quadratic-plus-quartic regime. The quadratic form $\E_i\tfrac12z_i^\top F_cz_i$ estimates $A=\tfrac12\operatorname{tr}F_c$, the Fisher trace in RMS units. Direction-level standard errors and 1,000 direction bootstrap replicates quantify Monte Carlo uncertainty.

\paragraph{LLC estimates.}
For every radius and each of the sixteen inverse temperatures
\begin{align*}
\tau\in\{&100,\,250,\,500,\,10^3,\,2500,\,4000,\,5000,\,10^4,\\
&1.6\times10^4,\,2.5\times10^4,\,4\times10^4,\,6.4\times10^4,\,10^5,\,2.5\times10^5,\,4\times10^5,\,10^6\},
\end{align*}
the estimates $q_K(\tau,r)$ and $q_Q(\tau,r)$ of Equation~\eqref{eq:nonquadratic-partition} are computed by self-normalized importance weighting of the 256 prior samples with weights $\mathbf 1_U(u)e^{-\tau K(u)}$ and $\mathbf 1_U(u)e^{-\tau Q(u)}$. The sign pair is the unit of resampling. A setting passes the diagnostics when the effective sample size over the 128 pairs is at least 100 for both targets and both Pareto tail indices $\hat k$ are below $0.5$ \citep{vehtari2024psis}; other settings are excluded from the developmental summaries. Because a Gaussian prior in millions of dimensions concentrates $K$ around its mean, the passing region is $\tau\,\mathrm{sd}[K(rz)]\lesssim1$; in this region $q_K-q_Q=\tau r^4B+O(r^6)$. Larger inverse temperatures are covered by the direct posterior sampling of Appendix~\ref{app:temperature-sweep}.

\paragraph{Matched ablation.}
At every component and checkpoint, the same run measures zero-ablation on the same inputs and positions: the component's output projection weights ($W_O$ slice, or $W_{\mathrm{out}}$ and $b_{\mathrm{out}}$) are set to zero, holding the shared attention output bias fixed, and the drop in the closer contrast, the KL from the clean model, cross-entropy, and accuracy are recorded. Distractors are one seeded incorrect closer per scored position, identical across components and checkpoints. This intervention is not part of the geometry sample set.

\paragraph{Execution.}
Each of the 6,160 component--checkpoint--coordinate combinations is evaluated in float64 with batches of eight sequences; the computation before the component's layer is cached, so the cost of a direction pair scales as $1+35(16-\ell)/16$ with the layer $\ell$. Every run records its clean metrics and coordinate scales, and a cell is assembled only when they agree across its runs.

\begin{figure}[t]
\centering
\includegraphics[width=\linewidth]{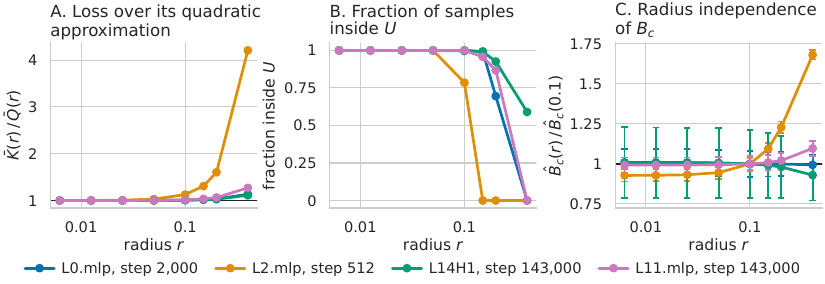}
\caption{\textbf{Inside $U$ the loss is quadratic plus quartic.} A: ratio of the mean KL along random directions (both signs) to its Fisher approximation, against radius, for four component--checkpoint pairs. B: fraction of samples inside $U$. C: the quartic coefficient estimated at each radius, relative to its value at $r=0.1$, with two direction-level standard errors; its constancy supports Equation~\eqref{eq:main-quartic}.}
\label{fig:local-regime}
\end{figure}

\begin{figure}[t]
\centering
\includegraphics[width=\linewidth]{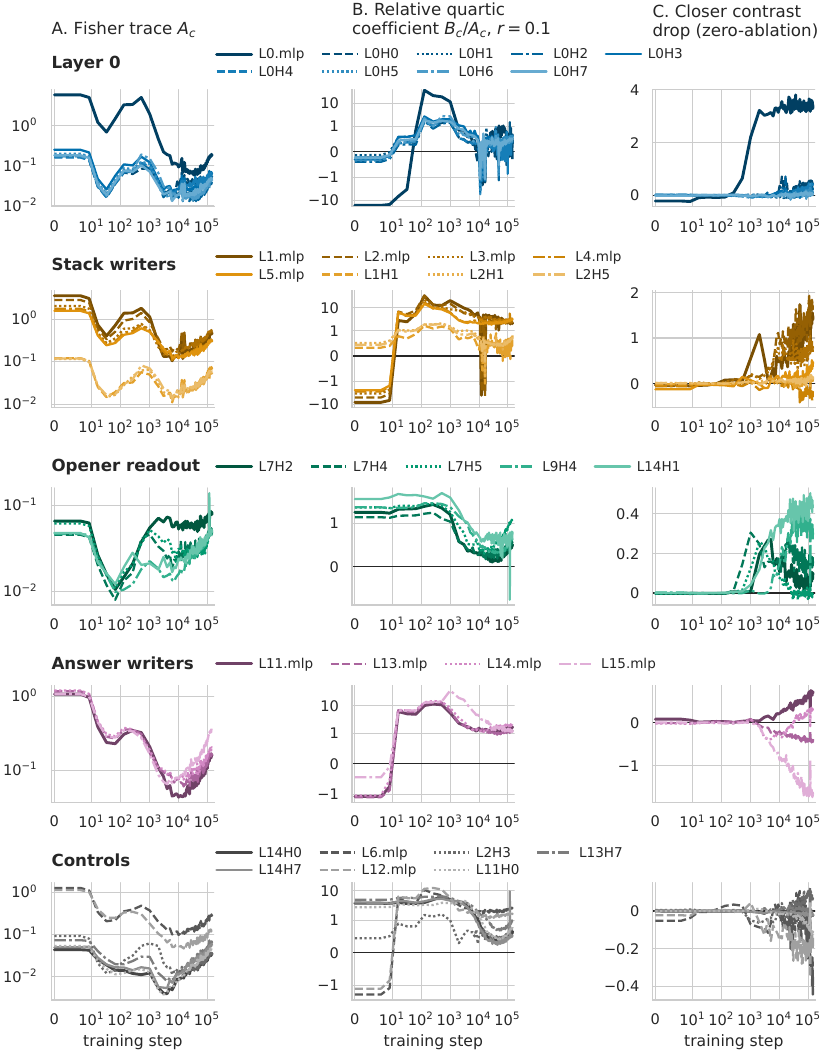}
\caption{\textbf{Local geometry across training, by circuit stage (33-component panel).} Rows are the circuit stages (stages 1--3 of Figure~\ref{fig:algorithm-mapping} and the answer-writing MLPs of stages 4--5) and the seven controls. Columns: Fisher trace $A_c$ (log scale), relative quartic coefficient $B_c/A_c$ at $r=0.1$ (symmetric logarithmic scale), and the drop in the closer contrast under zero-ablation measured on the same inputs. All 154 checkpoints; step 0 is drawn at the left of the logarithmic axis.}
\label{fig:geometry-stages}
\end{figure}

\begin{figure}[t]
\centering
\includegraphics[width=\linewidth]{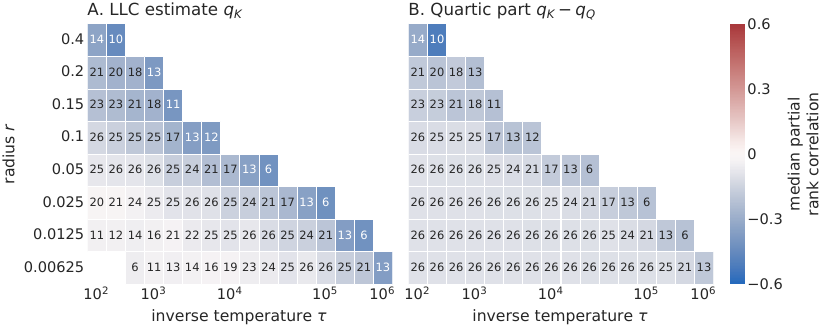}
\caption{\textbf{$A_c$-adjusted associations of the importance-sampled estimates across radius and temperature.} Median, over circuit components, of the partial rank correlation (adjusted for $A_c$, clean loss, weight scale, and training step) between the ablation-KL trajectory and the trajectory of $q_K$ (A) or of its quartic part $q_K-q_Q$ (B). Cells are blank where fewer than five components pass the importance-sampling diagnostics at $90\%$ of checkpoints or where the correlation is undefined; numbers give the components included, a set that shrinks as $\tau$ and $r$ grow (at $r=0.4$ only attention heads qualify). Within the passing region the estimates are linear in $\tau$ (Section~\ref{sec:local-development}), so at fixed $r$ the unadjusted rank association of $q_K$ equals that of $A_c$ over the same components at every $\tau$ ($0.44$ over the circuit components shown here, $0.55$ over all 33); what remains after adjusting for $A_c$ is the quartic term $\tau r^4B_c$ together with the second-order Fisher term $-\tfrac12\tau^2r^4\operatorname{tr}F_c^2$, and the variation of A across the grid, down to $-0.4$ in cells with six components, reflects the growing weight of that term and the shrinking passing set, not a change of regime. The quartic part (B) stays close to the association of $B_c$ throughout.}
\label{fig:association-surface}
\end{figure}

\paragraph{Grid tables.}
Table~\ref{tab:grid-locality} summarizes locality and the quadratic-plus-quartic regime over the complete grid, and Table~\ref{tab:grid-ab} lists the Fisher trace and quartic coefficient of every component at the six calibration checkpoints. Tables~\ref{tab:grid-locality-current} and~\ref{tab:grid-ab-current} repeat both summaries in current-checkpoint RMS units for the seven components of the temperature sweep (1,078 component--checkpoint pairs). In these units the trajectories have median within-component rank correlations of $0.60$ ($A_c$) and $0.49$ ($B_c$) with those in final-checkpoint units: dividing out the growth of the weights changes the trajectories in part, and we report the fixed units because a displacement in them means the same change of weights at every checkpoint.

\input{sections/grid_tables}
\input{sections/grid_tables_current}

\paragraph{LLC estimates across radius and temperature.}
The importance-weighted estimates $q_K(\tau,r)$ and $q_Q(\tau,r)$ pass the diagnostics (effective sample size at least 100 of 128 pairs and Pareto $\hat k<0.5$ for both targets) for $55\%$ of the (pair, $r$, $\tau$) settings, the region $\tau\,\mathrm{sd}[K(rz)]\lesssim1$ beyond which the prior no longer covers the posterior and Appendix~\ref{app:temperature-sweep} supplies direct sampling. Within it $q_Q=\tau r^2A_c$ to within $0.4\%$ and $q_K-q_Q=\tau r^4B_c$ to within $1.1\%$ (medians at $r\leq0.1$), as Equation~\eqref{eq:main-quartic} predicts, so the estimates across temperature and radius are generated by $A_c$ and $B_c$: temperature rescales them and radius sets their mixture, from pure curvature at $r\leq0.025$ to a quartic share of $2.4\%$ at $r=0.2$ and $9.5\%$ at $r=0.4$. Figure~\ref{fig:association-surface} shows the $A_c$-adjusted associations of the estimates with the ablation trajectories across the passing region.

\section{The Fisher spectrum at every checkpoint}
\label{app:spectrum}

\paragraph{Measurement.}
For each of the 144 components at each of the 154 checkpoints, in the fixed final-checkpoint RMS coordinates of Appendix~\ref{app:grid-methods}, one run measures on the developmental batch: (i) the causal checks of the grid, zero-ablation (contrast drop and KL), the local gain ablation ($g=-0.05$), and the gain Fisher $F_{gg}=\partial_g^2K_c(0)$ by central differences at $\epsilon=0.0025$ and $\epsilon/2$; (ii) the grid's first 64 directions at $r=0.1$, $r/2$, and $r/4$ with both signs, the exact quadratic form $\tfrac12z_i^\top F_cz_i$ and $\|F_cz_i\|^2$ by Hessian--vector products, which give $A_c$, $\operatorname{tr}F_c^2$, and $B_c$ exactly as in the grid (for the 33 panel components the values coincide with the grid's records bit for bit); (iii) the Fisher spectrum by stochastic Lanczos quadrature \citep{ubaru2017slq}: four unit-norm Gaussian probes, 32 Lanczos steps each with full reorthogonalization, on exact Hessian--vector products of the closer reference KL in float64. Each probe yields Ritz pairs $(\theta_{vk},w_{vk})$, and any spectral sum is estimated as $\sum_if(\mu_i)\approx\tfrac{d_c}{4}\sum_{v,k}w_{vk}f(\theta_{vk})$; the quadratic estimate at any $(\tau,r)$ is this sum with $f(\mu)=\tfrac12\tau r^2\mu/(1+\tau r^2\mu)$ (Equation~\eqref{eq:soft-rank}). On a two-layer transformer with a computable spectrum the per-probe quadrature reproduces the exact spectral sums to $10^{-15}$, the gain Fisher equals twice the quadratic form of the gain direction to $10^{-7}$, and the Hutchinson trace \citep[the mean of $z^\top F_cz$ over Gaussian $z$;][]{hutchinson1990trace} agrees with the exact trace within $1\%$; on Pythia-1B the quadrature trace agrees with the 64-direction $A_c$ within its standard error at every cell. Every cell verifies that its replay of the clean forward pass matches the reference to $10^{-7}$, and all 22,176 cells are complete.

\paragraph{Independent inputs.}
The 33-component panel is measured a second time on an independently generated batch of 128 strings (seed 20261001; 11,441 valid positions, 5,553 closer positions), with new directions and probes, for a replication in which nothing is shared with the original batch but the model. Its 5,078 complete cells (of 5,082) give the values quoted in Section~\ref{sec:correspondence}.

\paragraph{Growth of the count, precision, and component specificity.}
Table~\ref{tab:saturation} lists the median over components of $q_Q/(\tau r^2A_c)$ at the two checkpoints of Figure~\ref{fig:geometry-development}A. The count grows by a median factor of $3.9$ and $4.0$ per decade of $\tau r^2$ ending at $10^5$ and $10^6$ (exponent $0.6$) and by $6.7$ and $8.4$ ending at $10^7$ and $10^8$, as the bulk of the spectrum (smallest Ritz values $\approx10^{-9}$) comes into range: there is no plateau. The count is precise: across the four probes, $q_Q(10^6)$ at the final checkpoint has a relative standard deviation of $0.014$ (median over components; $0.029$ at the 90th percentile). $B_c$ averages all 64 directions; $97.9\%$ of the 144-component cells have every one of their 64 directions inside $U$ at $r=0.1$, and the mean fraction inside is $0.998$ (the panel's 128-direction grid gives $88.5\%$ of pairs fully inside, Table~\ref{tab:grid-locality}). Pairing each component's geometry with the population-median zero-ablation trajectory instead of its own gives adjacent-change partials of $0.50$ for $A_c$ (own $0.51$), $0.12$ for $F_{gg}$ (own $0.97$), and $0.13$ for $q_Q(10^6)$ (own $0.40$); with a randomly chosen other component's trajectory, $0.26$, $0.03$, and $0.08$; the own association exceeds the median of the cross pairings for $82\%$, $99\%$, and $83\%$ of the components. The circuit/other contrast of $q_Q(10^6)$ holds within heads (adjacent-change partial $0.10$ against $0.48$) and within MLPs ($0.08$ against $0.26$). Table~\ref{tab:spectrum-associations} lists the associations of every geometric quantity with every target, over all components and by group, computed as in Section~\ref{sec:correspondence} (partial rank correlations given clean loss, weight scale, and training step, with $A_c$ added for quantities other than $A_c$; adjacent-change partials given the same covariates; moving-block bootstrap intervals with blocks of 16 checkpoints).

\begin{table}[t]
\centering\small
\begin{tabular}{@{}lccccc@{}}
\toprule
$\tau r^2$ & $10^2$ & $10^3$ & $10^4$ & $10^5$ & $10^6$ \\
\midrule
step 8,000, 144 components & 0.934 & 0.744 & 0.437 & 0.206 & 0.082 \\
step 143,000, 144 components & 0.866 & 0.584 & 0.279 & 0.108 & 0.043 \\
step 8,000, panel on independent inputs & 0.878 & 0.682 & 0.402 & 0.202 & 0.082 \\
step 143,000, panel on independent inputs & 0.816 & 0.576 & 0.303 & 0.135 & 0.050 \\
\bottomrule
\end{tabular}
\caption{\textbf{Growth of the quadratic estimate.} Median over components of $q_Q(\tau,r)/(\tau r^2A_c)$, the ratio of the spectral estimate to its small-temperature limit.}
\label{tab:saturation}
\end{table}

\begin{table}[t]
\centering\scriptsize
\setlength{\tabcolsep}{3pt}
\begin{tabular}{@{}llcccccccc@{}}
\toprule
& & \multicolumn{2}{c}{zero-ablation KL} & \multicolumn{2}{c}{local ablation KL} & \multicolumn{2}{c}{opener-distance probe} & \multicolumn{2}{c}{SGLD estimate} \\
\cmidrule(lr){3-4}\cmidrule(lr){5-6}\cmidrule(lr){7-8}\cmidrule(lr){9-10}
geometry & group & levels & changes & levels & changes & levels & changes & levels & changes \\
\midrule
$A_c$ & all (144) & $0.15$ & $0.51$ & $0.16$ & $0.53$ & $-0.23$ & $0.01$ & $-0.07$ & $0.16$ \\
 & circuit (26) & $0.13$ & $0.56$ & $0.14$ & $0.60$ & $-0.19$ & $-0.07$ & $-0.01$ & $0.14$ \\
 & other (118) & $0.16$ & $0.49$ & $0.16$ & $0.50$ & $-0.23$ & $0.02$ & $-0.08$ & $0.16$ \\
$q_Q$, $\tau r^2=10^3$ & all & $-0.06$ & $0.10$ & $-0.07$ & $0.10$ & $-0.13$ & $-0.08$ & $0.07$ & $0.01$ \\
 & circuit & $-0.15$ & $0.08$ & $-0.23$ & $0.03$ & $-0.09$ & $-0.08$ & $0.02$ & $0.05$ \\
 & other & $-0.04$ & $0.12$ & $-0.05$ & $0.11$ & $-0.14$ & $-0.08$ & $0.08$ & $0.00$ \\
$q_Q$, $\tau r^2=10^6$ & all & $0.24$ & $0.40$ & $0.25$ & $0.40$ & $0.09$ & $-0.05$ & $0.37$ & $0.10$ \\
 & circuit & $-0.01$ & $0.09$ & $-0.05$ & $0.08$ & $0.18$ & $-0.03$ & $0.32$ & $0.04$ \\
 & other & $0.29$ & $0.47$ & $0.30$ & $0.47$ & $0.08$ & $-0.06$ & $0.39$ & $0.10$ \\
$B_c$ & all & $-0.24$ & $-0.07$ & $-0.24$ & $-0.10$ & $0.06$ & $0.04$ & $0.50$ & $0.15$ \\
 & circuit & $-0.10$ & $0.02$ & $-0.15$ & $-0.02$ & $0.14$ & $-0.00$ & $0.36$ & $0.08$ \\
 & other & $-0.27$ & $-0.09$ & $-0.26$ & $-0.12$ & $0.04$ & $0.04$ & $0.53$ & $0.16$ \\
$B_c/A_c$ & all & $-0.27$ & $-0.11$ & $-0.28$ & $-0.13$ & $0.20$ & $0.04$ & $0.47$ & $0.16$ \\
 & circuit & $-0.21$ & $-0.03$ & $-0.23$ & $-0.08$ & $0.13$ & $0.01$ & $0.33$ & $0.09$ \\
 & other & $-0.29$ & $-0.14$ & $-0.30$ & $-0.17$ & $0.24$ & $0.04$ & $0.51$ & $0.17$ \\
$F_{gg}$ & all & $0.98$ & $0.97$ & $1.00$ & $1.00$ & $0.13$ & $-0.01$ & $-0.02$ & $-0.03$ \\
 & circuit & $0.87$ & $0.80$ & $1.00$ & $1.00$ & $0.16$ & $0.05$ & $-0.00$ & $-0.04$ \\
 & other & $0.99$ & $0.98$ & $1.00$ & $1.00$ & $0.13$ & $-0.02$ & $-0.02$ & $-0.03$ \\
$A_c$ given $F_{gg}$ & all & $0.02$ & $0.07$ & & & & & & \\
 & circuit & $-0.00$ & $0.11$ & & & & & & \\
 & other & $0.02$ & $0.05$ & & & & & & \\
\bottomrule
\end{tabular}
\caption{\textbf{Associations across training from the spectrum measurement.} Median over components of the partial rank correlation for levels and for adjacent-checkpoint changes; ``$A_c$ given $F_{gg}$'' adds the gain Fisher to the covariates. Over the 144 components the change association of $A_c$ with the zero-ablation KL is positive for $97\%$ and its block-bootstrap interval lies above zero for $88\%$; that of $F_{gg}$ for $99\%$ and $98\%$; that of $q_Q$ at $\tau r^2=10^6$ for $95\%$ and $67\%$.}
\label{tab:spectrum-associations}
\end{table}

\section{The quartic correction at large temperature}
\label{app:quartic-posterior}

\paragraph{Estimator.}
Write $K_c=Q_c+V_c$ with $V_c$ the part of the loss beyond its quadratic approximation, and $\pi_Q=\mathcal N(0,(\tau F_c+I/r^2)^{-1})$ for the posterior of the quadratic loss on the prior $\varphi_r$. Expanding the estimate~\eqref{eq:estimator} to first order in $\tau V_c$ gives
\begin{equation}
 q_K-q_Q=\tau\,\E_{\pi_Q}[V_c]-\tau^2\operatorname{Cov}_{\pi_Q}(Q_c,V_c)+O(\tau^3V_c^2),
 \label{eq:quartic-correction}
\end{equation}
which reduces to $\tau r^4B_c$ when $\tau r^2\mu_i\ll1$ for every eigenvalue. The first term is the quartic part of the loss averaged over the directions the posterior still explores; the covariance term involves only the directions that the posterior has pinned, whose variance is $1/(\tau\mu_i)$ rather than $r^2$, and is of order $N_{\mathrm{stiff}}/\tau$, where $N_{\mathrm{stiff}}$ is the number of such directions. We draw from $\pi_Q$ exactly: with $J$ the Jacobian of the scored log-probabilities and $\Sigma=\operatorname{diag}(p)-pp^\top$ the categorical Fisher metric at each scored position, $w=\sqrt\tau\,J^\top\Sigma^{1/2}\xi_1+\xi_2/r$ with standard normal $\xi_1,\xi_2$ has covariance $\tau F_c+I/r^2$, so that $u=(\tau F_c+I/r^2)^{-1}w$, obtained by conjugate gradients on exact Hessian--vector products (relative residual $10^{-4}$), is a draw from $\pi_Q$. Each draw costs one backward pass, the conjugate-gradient solve, one further Hessian--vector product for $Q_c(u)=\tfrac12u^\top F_cu$, and two forward passes for $K_c(u)$ and $K_c(-u)$, whose average removes every odd-order term. On a two-layer transformer with a computable spectrum the covariance of $J^\top\Sigma^{1/2}\xi$ reproduces $F_c$, the solve agrees with a direct solve to $10^{-5}$, $\tau\,\E[Q_c]$ agrees with Equation~\eqref{eq:soft-rank} within $2\%$, and $K_c(u)-Q_c(u)$ agrees draw by draw with the quartic term isolated by the radius contrast of Appendix~\ref{app:grid-methods} to $10^{-4}$; on Pythia-1B, solving to a residual of $10^{-7}$ instead changes $Q_c$ and $V_c$ by less than $0.5\%$.

\paragraph{Campaign.}
For the 33-component panel at the six calibration checkpoints and every sixth confirmation checkpoint (28 checkpoints), six draws at each of six settings, $r=0.025$ with $\tau r^2\in\{10,10^2,10^3\}$ and $r=0.0125$ with $\tau r^2\in\{10^2,10^3,10^4\}$, in the fixed final-checkpoint coordinates and on the developmental batch; every draw is checked against the region $U$.

\paragraph{Results.}
Table~\ref{tab:quartic} summarizes the 924 complete cells. At every setting the Monte Carlo quadratic part reproduces the quadrature $q_Q$ at the same $\tau r^2$ (median ratio 1.001--1.022; within $10\%$ for 99\% of the cells at $\tau r^2=10^4$, where the posterior is narrow, and within $20\%$ for 81\% at $\tau r^2=10$, where six draws of a prior-dominated quadratic form are noisy), which cross-validates the Lanczos spectra with an independent estimator. The posterior-weighted quartic coefficient $\E[V_c]/r^4$ equals $B_c$ at every setting (medians 0.91--1.05 of $B_c$), so the quartic part of the loss lies in the directions the posterior still explores, the correction is $\tau r^4B_c$ throughout, and it grows linearly in $\tau$ while $q_Q$ grows sublinearly; its share of the estimate is 0.041\% at $\tau r^2=10$ and 0.032\% at $10^4$ (medians; $90$th percentiles below 0.33\%). The expansion is first order in $\tau V_c$: $\max\tau|V_c|<0.3$ holds for every cell at $\tau r^2\le10^2$, for 97\% and 80\% at $10^3$ ($r=0.0125$ and $0.025$), and for 70\% at $10^4$, the exceptions being MLPs, whose $B_c$ is largest; for those cells the first-order correction is 0.12\% of the estimate in median and 2.39\% at most, and the sample covariance term stays below 0.42 of $\delta q$ in median, so the large-temperature estimate remains the soft rank to within a percent. Every draw lies inside $U$. Projected with the measured spectra, the leading term $\tau r^4B_c$ would be 0.19\% of $q_Q$ at $\tau r^2=10^6$ for $r=0.0125$ (median over cells).

\begin{table}[t]
\centering\small
\begin{tabular}{@{}cccccc@{}}
\toprule
$r$ & $\tau r^2$ & $\delta q/\hat q$ (median [IQR]) & $\tau\E[Q_c]/q_Q$ & $\E[V_c]/(r^4B_c)$ & $\max\tau|V_c|$ (p90) \\
\midrule
0.0125 & $10^{2}$ & 0.013\% [0.008\%, 0.037\%] & 1.017 [0.965, 1.094] & 1.05 [0.96, 1.16] & 0.017 \\
0.0125 & $10^{3}$ & 0.019\% [0.012\%, 0.050\%] & 1.001 [0.973, 1.028] & 1.02 [0.94, 1.15] & 0.162 \\
0.0125 & $10^{4}$ & 0.032\% [0.018\%, 0.079\%] & 1.002 [0.986, 1.021] & 0.91 [0.74, 1.01] & 1.367 \\
0.025 & $10^{1}$ & 0.041\% [0.026\%, 0.13\%] & 1.022 [0.956, 1.118] & 1.01 [0.91, 1.11] & 0.006 \\
0.025 & $10^{2}$ & 0.049\% [0.031\%, 0.15\%] & 1.020 [0.967, 1.086] & 1.01 [0.92, 1.10] & 0.064 \\
0.025 & $10^{3}$ & 0.076\% [0.048\%, 0.20\%] & 1.003 [0.972, 1.035] & 1.01 [0.94, 1.12] & 0.614 \\
\bottomrule
\end{tabular}
\caption{\textbf{The quartic correction at large temperature.} Medians over the panel cells of the share of the correction in the corrected estimate $\hat q=\tau\E[Q_c]+\delta q$, the Monte Carlo quadratic part relative to the quadrature $q_Q$ at the same $\tau r^2$, the posterior-weighted quartic coefficient relative to $B_c$, and the first-order validity $\max\tau|V_c|$.}
\label{tab:quartic}
\end{table}

\section{Where the change of the Fisher trace comes from}
\label{app:decomposition}

\paragraph{Hybrid checkpoints.}
$A_c=\tfrac12\operatorname{tr}(J^\top\Sigma_cJ)$ depends on the component's own weights (through its output map $J=\partial o_c/\partial u$), on the upstream weights that set its input, and on the downstream weights that set the metric $\Sigma_c$. For two adjacent checkpoints $t<t'$ of the 28 checkpoints of Appendix~\ref{app:quartic-posterior} we partition the parameters into three groups, the component's own tensors (J), the embedding and the blocks before its layer (I), and everything else, the blocks after its layer, the final normalization, the unembedding, and the remaining tensors of its own block (M), and evaluate $A_c$ on the eight models that take each group from $t$ or from $t'$ (the loader folds layer norms and centers writing weights within a block, so splicing whole blocks is exact). The reference distribution is each hybrid's own predictions, the 64 directions are those of the spectrum campaign, and the evaluation of the all-$t$ model replays the spectrum profile bit for bit, as the all-$t'$ hybrid replays the next checkpoint. The change $\Delta A_c=A_c(t')-A_c(t)$ is attributed exactly to the three groups by Shapley values over the eight evaluations, which sum to $\Delta A_c$ cell by cell; the single-hybrid parts alone do not add up (their residual is $38\%$ of the attributed magnitude), because a hybrid of checkpoints 6,000 steps apart is a mismatched network whose clean loss moves by up to $0.2$ nats. All 891 cells (33 components, 27 pairs) are complete.

\paragraph{Results.}
Over the 615 cells in which $|\Delta A_c|$ exceeds $5\%$ of $A_c$, the metric group carries $0.44$ of the attributed change, the input group $0.35$, and the component's own tensors $0.21$; the metric and input attributions have the sign of $\Delta A_c$ in $68\%$ and $67\%$ of the cells (median signed shares $0.31$ and $0.44$), the own-tensor attribution in $49\%$ (median share $0.00$). Heads and MLPs differ: for heads the metric group carries a median share of $0.50$ of a change and the input group $0.24$; for MLPs the input group carries $0.92$ and the metric group $0.06$. The late rise of $A_c$ (pairs from step 8,000 on, summed per component) is carried by the input group (median $1.08$ of the rise, interquartile range $0.33$--$1.99$), with the metric group at $0.05$ ($-0.86$--$0.47$) and the component's own tensors at $-0.23$ ($-0.49$--$0.00$): on their own, the component's parameters would have lowered its Fisher trace. In adjacent-change associations with the component's zero-ablation KL over these 28 checkpoints (partials given clean loss, weight scale, and, for the parts, $A_c$), the full $A_c$ series gives $0.44$ ($0.39$ against the population-median trajectory), the cumulative metric series $0.17$ ($0.12$), the own-tensor series $0.12$ ($-0.06$; $0.35$ in levels), and the input series $0.02$, so no single group carries the co-movement.

%% file: sections/grid_tables.tex
\begin{table}[t]
\centering\footnotesize
\setlength{\tabcolsep}{4pt}
\begin{tabular}{rrrrrl}
\toprule
$r$ & fully inside & median inside & max closer KL & $|\bar K/\bar Q-1|$ & $\widehat B(r)/\widehat B(0.1)$ (IQR) \\
\midrule
0.00625 & 1.000 & 1.000 & 0.0000 & 0.0000 & 1.001 (0.998, 1.006) \\
0.0125 & 1.000 & 1.000 & 0.0000 & 0.0001 & 1.001 (0.998, 1.005) \\
0.025 & 1.000 & 1.000 & 0.0001 & 0.0004 & 1.001 (0.998, 1.005) \\
0.05 & 0.983 & 1.000 & 0.0003 & 0.0016 & 1.000 (0.999, 1.004) \\
0.1 & 0.885 & 1.000 & 0.0011 & 0.0065 & 1.000 (1.000, 1.000) \\
0.15 & 0.692 & 1.000 & 0.0024 & 0.0143 & 0.999 (0.993, 1.002) \\
0.2 & 0.522 & 1.000 & 0.0044 & 0.0250 & 0.998 (0.984, 1.005) \\
0.4 & 0.179 & 0.855 & 0.0187 & 0.0977 & 0.994 (0.932, 1.021) \\
\bottomrule
\end{tabular}
\caption{\textbf{Locality and the quadratic-plus-quartic regime across the grid.} Over all 5,082 component--checkpoint pairs in fixed final-checkpoint RMS units: the fraction of cells in which every sample at the radius lies inside $U$, the median fraction of samples inside $U$, the median over cells of the maximum closer KL among the 256 samples, the median relative excess of the mean loss over its Fisher approximation, and the median (with interquartile range) of the quartic coefficient at the radius relative to its value at $r=0.1$.}
\label{tab:grid-locality}
\end{table}

\begin{table}[p]
\centering\footnotesize
\setlength{\tabcolsep}{3pt}
\begin{tabular}{lrrrrrrrrrrrr}
\toprule
& \multicolumn{2}{c}{0} & \multicolumn{2}{c}{512} & \multicolumn{2}{c}{2,000} & \multicolumn{2}{c}{8,000} & \multicolumn{2}{c}{32,000} & \multicolumn{2}{c}{143,000} \\
\cmidrule(lr){2-3} \cmidrule(lr){4-5} \cmidrule(lr){6-7} \cmidrule(lr){8-9} \cmidrule(lr){10-11} \cmidrule(lr){12-13}
Component & $A$ & $B$ & $A$ & $B$ & $A$ & $B$ & $A$ & $B$ & $A$ & $B$ & $A$ & $B$ \\
\midrule
\C{L0.mlp} & 5.894 & -100 & 5.088 & 96.6 & 0.465 & 0.390 & 0.114 & 0.007 & 0.064 & 0.037 & 0.191 & 0.079 \\
\C{L0H0} & 0.167 & -0.039 & 0.084 & 0.118 & 0.038 & 0.021 & 0.026 & -0.004 & 0.036 & 0.003 & 0.054 & 0.038 \\
\C{L0H1} & 0.187 & -0.026 & 0.106 & 0.166 & 0.034 & 0.031 & 0.021 & 0.010 & 0.021 & 0.009 & 0.056 & 0.029 \\
\C{L0H2} & 0.173 & -0.069 & 0.106 & 0.140 & 0.035 & 0.019 & 0.017 & 0.005 & 0.014 & 0.003 & 0.037 & 0.014 \\
\C{L0H3} & 0.251 & -0.077 & 0.169 & 0.335 & 0.045 & 0.048 & 0.025 & 0.016 & 0.026 & 0.013 & 0.066 & 0.042 \\
\C{L0H4} & 0.162 & -0.042 & 0.119 & 0.214 & 0.035 & 0.018 & 0.023 & 0.006 & 0.036 & 0.001 & 0.058 & 0.029 \\
\C{L0H5} & 0.201 & -0.069 & 0.127 & 0.210 & 0.039 & 0.027 & 0.017 & 0.009 & 0.018 & 0.007 & 0.048 & 0.028 \\
\C{L0H6} & 0.176 & -0.039 & 0.192 & 0.529 & 0.066 & 0.055 & 0.019 & 0.006 & 0.015 & 0.004 & 0.045 & 0.019 \\
\C{L0H7} & 0.184 & -0.045 & 0.104 & 0.177 & 0.031 & 0.029 & 0.017 & 0.009 & 0.019 & 0.007 & 0.048 & 0.024 \\
\C{L1.mlp} & 3.530 & -29.0 & 1.787 & 22.3 & 0.269 & 3.420 & 0.130 & 0.698 & 0.179 & 0.723 & 0.470 & 1.011 \\
\C{L2.mlp} & 2.801 & -14.1 & 1.395 & 18.0 & 0.344 & 2.802 & 0.204 & 0.675 & 0.243 & 0.635 & 0.506 & 1.215 \\
\C{L3.mlp} & 2.017 & -6.822 & 0.768 & 8.030 & 0.220 & 1.633 & 0.165 & 0.626 & 0.236 & 0.854 & 0.520 & 1.310 \\
\C{L4.mlp} & 1.736 & -4.756 & 0.722 & 7.220 & 0.256 & 0.754 & 0.140 & 0.360 & 0.157 & 0.326 & 0.317 & 0.992 \\
\C{L5.mlp} & 1.560 & -3.765 & 0.607 & 5.131 & 0.289 & 0.809 & 0.142 & 0.319 & 0.151 & 0.325 & 0.309 & 0.958 \\
\C{L1H1} & 0.121 & 0.040 & 0.057 & 0.085 & 0.026 & 0.017 & 0.016 & 0.008 & 0.022 & 0.008 & 0.071 & 0.037 \\
\C{L2H1} & 0.115 & 0.059 & 0.063 & 0.130 & 0.033 & 0.032 & 0.015 & 0.013 & 0.022 & 0.011 & 0.055 & 0.045 \\
\C{L2H5} & 0.119 & 0.053 & 0.082 & 0.159 & 0.038 & 0.032 & 0.016 & 0.013 & 0.025 & 0.011 & 0.061 & 0.042 \\
\C{L9H4} & 0.046 & 0.117 & 0.018 & 0.055 & 0.018 & 0.027 & 0.014 & 0.009 & 0.025 & 0.008 & 0.052 & 0.032 \\
\C{L14H1} & 0.048 & 0.195 & 0.020 & 0.117 & 0.018 & 0.038 & 0.019 & 0.015 & 0.030 & 0.017 & 0.050 & 0.037 \\
\C{L7H4} & 0.045 & 0.064 & 0.041 & 0.053 & 0.033 & 0.019 & 0.023 & 0.007 & 0.031 & 0.008 & 0.050 & 0.028 \\
\C{L7H5} & 0.061 & 0.154 & 0.036 & 0.098 & 0.044 & 0.040 & 0.022 & 0.013 & 0.027 & 0.010 & 0.047 & 0.056 \\
\C{L7H2} & 0.066 & 0.123 & 0.038 & 0.091 & 0.071 & 0.041 & 0.064 & 0.021 & 0.060 & 0.009 & 0.081 & 0.039 \\
\C{L11.mlp} & 1.072 & -1.293 & 0.323 & 3.460 & 0.108 & 0.245 & 0.048 & 0.056 & 0.057 & 0.068 & 0.161 & 0.253 \\
\C{L13.mlp} & 1.169 & -1.484 & 0.309 & 4.012 & 0.105 & 0.241 & 0.075 & 0.098 & 0.073 & 0.076 & 0.158 & 0.197 \\
\C{L14.mlp} & 1.200 & -1.324 & 0.283 & 3.938 & 0.106 & 0.373 & 0.078 & 0.099 & 0.094 & 0.122 & 0.203 & 0.387 \\
\C{L15.mlp} & 1.075 & -0.478 & 0.272 & 3.814 & 0.086 & 1.627 & 0.078 & 0.353 & 0.124 & 0.167 & 0.356 & 0.416 \\
\C{L14H0}$^\ast$ & 0.044 & 0.165 & 0.011 & 0.070 & 0.005 & 0.021 & 0.008 & 0.009 & 0.015 & 0.005 & 0.035 & 0.022 \\
\C{L6.mlp}$^\ast$ & 1.284 & -2.671 & 0.485 & 3.372 & 0.200 & 0.641 & 0.116 & 0.210 & 0.132 & 0.252 & 0.294 & 0.771 \\
\C{L2H3}$^\ast$ & 0.093 & 0.041 & 0.054 & 0.084 & 0.054 & 0.016 & 0.013 & 0.009 & 0.014 & 0.007 & 0.036 & 0.027 \\
\C{L13H7}$^\ast$ & 0.074 & 0.361 & 0.030 & 0.192 & 0.012 & 0.053 & 0.011 & 0.024 & 0.023 & 0.018 & 0.065 & 0.063 \\
\C{L14H7}$^\ast$ & 0.050 & 0.196 & 0.017 & 0.089 & 0.006 & 0.025 & 0.011 & 0.010 & 0.021 & 0.007 & 0.058 & 0.030 \\
\C{L12.mlp}$^\ast$ & 1.141 & -1.495 & 0.306 & 3.344 & 0.105 & 0.219 & 0.056 & 0.115 & 0.060 & 0.072 & 0.129 & 0.213 \\
\C{L11H0}$^\ast$ & 0.054 & 0.147 & 0.015 & 0.059 & 0.005 & 0.018 & 0.007 & 0.011 & 0.017 & 0.008 & 0.045 & 0.036 \\
\bottomrule
\end{tabular}
\caption{\textbf{Fisher trace and quartic coefficient at six checkpoints.} $A=\tfrac12\operatorname{tr}F_c$ is the Fisher trace in RMS units and $B$ the quartic coefficient at $r=0.1$, both in fixed final-checkpoint RMS units at closer positions. Control components are marked $^\ast$.}
\label{tab:grid-ab}
\end{table}

%% file: sections/grid_tables_current.tex
\begin{table}[t]
\centering\footnotesize
\setlength{\tabcolsep}{4pt}
\begin{tabular}{rrrrrl}
\toprule
$r$ & fully inside & median inside & max closer KL & $|\bar K/\bar Q-1|$ & $\widehat B(r)/\widehat B(0.1)$ (IQR) \\
\midrule
0.00625 & 1.000 & 1.000 & 0.0000 & 0.0000 & 1.001 (0.996, 1.012) \\
0.0125 & 1.000 & 1.000 & 0.0000 & 0.0001 & 1.001 (0.996, 1.011) \\
0.025 & 1.000 & 1.000 & 0.0001 & 0.0004 & 1.001 (0.996, 1.011) \\
0.05 & 1.000 & 1.000 & 0.0003 & 0.0016 & 1.001 (0.997, 1.009) \\
0.1 & 0.894 & 1.000 & 0.0014 & 0.0061 & 1.000 (1.000, 1.000) \\
0.15 & 0.661 & 1.000 & 0.0031 & 0.0134 & 0.998 (0.987, 1.004) \\
0.2 & 0.580 & 1.000 & 0.0057 & 0.0236 & 0.996 (0.970, 1.008) \\
0.4 & 0.186 & 0.838 & 0.0254 & 0.0910 & 0.985 (0.881, 1.030) \\
\bottomrule
\end{tabular}
\caption{\textbf{Locality and the quadratic-plus-quartic regime in current-checkpoint RMS units.} Over the 1,078 component--checkpoint pairs of the seven components of the temperature sweep in current-checkpoint RMS units: the fraction of cells in which every sample at the radius lies inside $U$, the median fraction of samples inside $U$, the median over cells of the maximum closer KL among the 256 samples, the median relative excess of the mean loss over its Fisher approximation, and the median (with interquartile range) of the quartic coefficient at the radius relative to its value at $r=0.1$.}
\label{tab:grid-locality-current}
\end{table}

\begin{table}[t]
\centering\footnotesize
\setlength{\tabcolsep}{3pt}
\begin{tabular}{lrrrrrrrrrrrr}
\toprule
& \multicolumn{2}{c}{0} & \multicolumn{2}{c}{512} & \multicolumn{2}{c}{2,000} & \multicolumn{2}{c}{8,000} & \multicolumn{2}{c}{32,000} & \multicolumn{2}{c}{143,000} \\
\cmidrule(lr){2-3} \cmidrule(lr){4-5} \cmidrule(lr){6-7} \cmidrule(lr){8-9} \cmidrule(lr){10-11} \cmidrule(lr){12-13}
Component & $A$ & $B$ & $A$ & $B$ & $A$ & $B$ & $A$ & $B$ & $A$ & $B$ & $A$ & $B$ \\
\midrule
\C{L0.mlp} & 0.357 & -0.339 & 0.359 & 0.942 & 0.145 & 0.660 & 0.088 & 0.081 & 0.094 & 0.075 & 0.191 & 0.079 \\
\C{L0H0} & 0.010 & 0.005 & 0.007 & 0.005 & 0.011 & 0.004 & 0.019 & 0.002 & 0.039 & 0.004 & 0.054 & 0.038 \\
\C{L0H2} & 0.010 & 0.005 & 0.008 & 0.005 & 0.008 & 0.003 & 0.010 & 0.004 & 0.016 & 0.004 & 0.037 & 0.014 \\
\C{L2.mlp} & 0.125 & 0.024 & 0.080 & 0.221 & 0.053 & 0.190 & 0.123 & 0.249 & 0.345 & 1.265 & 0.506 & 1.215 \\
\C{L14H1} & 0.001 & 0.001 & 0.001 & 0.000 & 0.003 & 0.001 & 0.009 & 0.003 & 0.034 & 0.022 & 0.050 & 0.037 \\
\C{L7H4} & 0.002 & 0.001 & 0.008 & 0.002 & 0.012 & 0.002 & 0.016 & 0.004 & 0.043 & 0.017 & 0.050 & 0.028 \\
\C{L11.mlp} & 0.028 & 0.015 & 0.010 & 0.012 & 0.019 & 0.038 & 0.046 & 0.065 & 0.106 & 0.244 & 0.161 & 0.253 \\
\bottomrule
\end{tabular}
\caption{\textbf{Fisher trace and quartic coefficient at six checkpoints in current-checkpoint RMS units.} $A=\tfrac12\operatorname{tr}F_c$ is the Fisher trace in RMS units and $B$ the quartic coefficient at $r=0.1$, both in current-checkpoint RMS units at closer positions, for the seven components of the temperature sweep.}
\label{tab:grid-ab-current}
\end{table}

%% file: sections/temperature_sweep.tex
\section{Direct posterior sampling across temperatures at six checkpoints}
\label{app:temperature-sweep}

\begin{figure}[t]
\centering
\includegraphics[width=\linewidth]{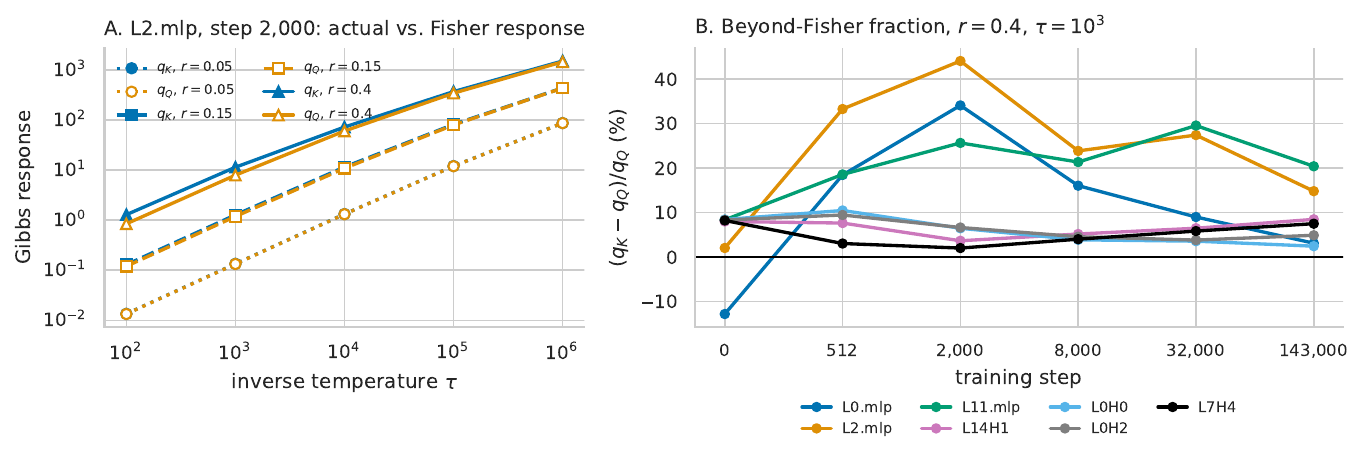}
\caption{\textbf{Direct posterior sampling across temperatures.} A: for \C{L2.mlp} at step 2,000, the LLC estimate $q_K$ and its Fisher counterpart $q_Q$ as functions of the inverse temperature at three radii. Both grow almost linearly in $\tau$ over the sampled range, with a mild sublinearity toward a soft count of constrained directions; the actual response exceeds the Fisher response most at moderate $\tau$ and large radius, where the posterior explores the scale at which quartic terms matter. B: the beyond-Fisher fraction $(q_K-q_Q)/q_Q$ at $r=0.4$ and $\tau=10^3$ across the six checkpoints for all seven components.}
\label{fig:temperature-sweep}
\end{figure}

The importance-weighted LLC estimates of Appendix~\ref{app:grid-methods} pass their diagnostics only where the posterior weights remain close to the prior. To cover larger inverse temperatures, where the posterior concentrates on the low-loss valley of a component, we sample the localized posteriors directly for seven components (\C{L0H0}, \C{L0H2}, \C{L0.mlp}, \C{L2.mlp}, \C{L7H4}, \C{L11.mlp}, \C{L14H1}) at six checkpoints (0, 512, 2,000, 8,000, 32,000, 143,000), three radii ($0.05$, $0.15$, $0.4$ in current-checkpoint tensor-RMS coordinates), and five inverse temperatures ($10^2$, $10^3$, $10^4$, $10^5$, $10^6$). Both the actual reference-KL target and the exact Fisher-quadratic target are sampled with the same prior: 1,260 targets, four chains each, 1,024 adaptive warmup iterations followed by 1,024 fixed-kernel draws per chain, with a Metropolis-adjusted prior-preserving Langevin proposal whose gradients are the exact full-batch gradients of the target (the final-checkpoint panel of Appendix~\ref{app:gibbs-methods} uses preconditioned Crank--Nicolson (pCN) proposals; this sweep uses gradient-informed proposals). Model arithmetic is float32 with float64 loss reductions. A target passes the energy diagnostic when the four chains have $\widehat R<1.01$ and effective sample sizes above 400 for the energy observable; 405 of the 630 matched actual/quadratic pairs pass for both targets, and 329 pass for every recorded observable. The targets that do not pass concentrate at $\tau\geq10^5$, where the stiffest directions force a small step and the remaining directions decorrelate slowly, and at $r=0.4$ for the MLPs, where the sampled states leave the region $U$; every target is reported (Figure~\ref{fig:temperature-sweep}).

\paragraph{The two regimes.}
At the smallest radius the two estimates agree to within $1.2\%$ at every temperature and checkpoint: the geometry is quadratic at that scale. At $r=0.15$ the excess is $1$--$10\%$ and at $r=0.4$ it reaches $30$--$60\%$ for the MLPs, but at $r=0.4$ the MLP chains leave the region $U$ (maximum closer KL up to $0.05$ nats and $|\Delta\mathrm{CE}|$ up to $0.38$), so only the heads pass there. Where the chains pass their diagnostics ($\tau\leq10^4$ at $r\leq0.15$), the response is linear in $\tau$ and the relative excess is independent of $\tau$ and proportional to $r^2$: the posterior is still the prior, and the sweep measures the same prior-averaged quartic term $\tau r^4B_c$ as the developmental grid, with the precision of 4,096 draws. At $\tau\geq10^5$ the sampled excess decreases toward zero, as expected when the posterior concentrates on states with $K\lesssim1/\tau$, but there the chains fail the $\widehat R$ and effective-sample-size screen, so the contraction regime is indicated rather than resolved. The information beyond curvature that these samplers can certify therefore resides in the finite-radius regime that the developmental grid measures at every checkpoint.

\input{sections/temperature_sweep_table}

\paragraph{Developmental signal.}
The relative excess $(q_K-q_Q)/q_Q$ is component-specific and changes across training. For \C{L0.mlp} at $r=0.4$ and $\tau=10^3$ it is $-13\%$ at initialization (the loss saturates below its quadratic approximation), rises to $+18\%$ at step 512 and $+34\%$ at step 2,000 during circuit formation, and decays to $+3\%$ at the final checkpoint. \C{L2.mlp} and \C{L11.mlp} rise to $+44\%$ and $+26\%$ at step 2,000 and remain elevated; the attention heads \C{L14H1}, \C{L0H0}, \C{L0H2}, and \C{L7H4} stay between $+2\%$ and $+11\%$. The same ordering holds at $\tau=10^2$ and at $r=0.15$. Table~\ref{tab:temperature-sweep} lists the complete values at both radii. The quartic coefficient of Section~\ref{sec:local-development} resolves the same quantity at every checkpoint with a direction-level Monte Carlo error rather than a chain diagnostic.

%% file: sections/temperature_sweep_table.tex
\begin{table}[t]
\centering\footnotesize
\setlength{\tabcolsep}{3pt}
\begin{tabular}{lrrrrrrrrrrrr}
\toprule
& \multicolumn{6}{c}{$r=0.15$, $\tau=10^3$} & \multicolumn{6}{c}{$r=0.4$, $\tau=10^3$} \\
\cmidrule(lr){2-7}\cmidrule(lr){8-13}
Component & 0 & 512 & 2,000 & 8,000 & 32,000 & 143,000 & 0 & 512 & 2,000 & 8,000 & 32,000 & 143,000 \\
\midrule
\C{L0.mlp} & -2.1 & +5.0 & +8.6 & +2.3 & +1.8 & +0.8 & -12.8$^\dagger$ & +18.5$^\dagger$ & +34.1 & +16.1$^\dagger$ & +9.0 & +3.1$^\dagger$ \\
\C{L2.mlp} & +0.4 & +5.9 & +7.6 & +3.9 & +6.9 & +4.1 & +2.0 & +33.3 & +44.1 & +23.9 & +27.4 & +14.8$^\dagger$ \\
\C{L11.mlp} & +1.2 & +2.6 & +4.4 & +3.4 & +5.2 & +3.3 & +8.4 & +18.6 & +25.7 & +21.4 & +29.6 & +20.4 \\
\C{L14H1} & +1.1 & +1.1 & +0.5 & +0.8 & +1.2 & +1.6 & +8.0 & +7.6 & +3.7 & +5.2 & +6.5 & +8.5 \\
\C{L0H0} & +1.2 & +1.5 & +1.0 & +0.3 & +0.7 & +0.6 & +8.4 & +10.5 & +6.5 & +3.9 & +3.6 & +2.4 \\
\C{L0H2} & +1.2 & +1.4 & +1.1 & +0.8 & +0.5 & +0.6 & +8.3 & +9.5 & +6.6 & +4.6 & +3.8 & +4.9 \\
\C{L7H4} & +1.1 & +0.6 & +0.3 & +0.3 & +0.8 & +1.1 & +8.2 & +3.0 & +2.0 & +4.0 & +5.8 & +7.5 \\
\bottomrule
\end{tabular}
\caption{\textbf{Beyond-Fisher fraction from direct posterior sampling.} Entries are $100\,(q_K-q_Q)/q_Q$ at inverse temperature $10^3$ for two radii and six checkpoints (columns are training steps). $^\dagger$ marks pairs in which at least one target fails the four-chain energy diagnostic. At $r=0.05$ every entry is within $1.2\%$ of zero.}
\label{tab:temperature-sweep}
\end{table}

%% file: sections/methods_appendix.tex
\section{Experimental protocols}
\label{app:methods}

\subsection{SGLD sweep of all components}
A complementary sweep applies the SGLD estimator of prior work \citep{lau2024llc} to every component at every checkpoint. Its chains move the weights well beyond $U$ at this scale (Appendix~\ref{app:locality}), so we treat its outputs as finite perturbation responses at a larger scale than the local measurements of Appendix~\ref{app:grid-methods} and compare the two in Section~\ref{sec:correspondence}. The sweep contains $2\times154\times(144+16+1)=49{,}588$ finite SGLD responses, covering both task variants, all single components, complete layers, and the full model. Each restriction uses two fp32 SGLD chains, comprising 100--150 initial updates and 60 subsequent draws. The response $R_C$ averages all updates. The objective is full-vocabulary cross-entropy at closer targets, with zero contribution at other positions and normalization over all next-token positions. Sampling settings are given below; Appendix~\ref{app:trace-reanalysis} analyzes the averaging window, and Appendix~\ref{app:locality} measures the resulting parameter displacements and compares them with localized posterior sampling.
\subsection{Bracket tasks and intervention scores}
Synthetic examples use space-prefixed bracket tokens after a \C{Brackets:} prefix. The bounded Dyck generator uses $(k,m)=(2,10)$ or $(4,16)$ and opening probability $0.45$ whenever either action is legal; the boundaries force the appropriate action. At the final checkpoint, the top prediction at the 5,475 scored closer positions of the $k=4$ developmental batch is the correct closer at $59.5\%$ of them, another closer at $6.0\%$, an opening bracket at $31.0\%$, and another token at $3.5\%$, so $77\%$ of the errors are openers; restricted to the four closers, the correct one ranks first at $86.0\%$. An opening bracket is a legal continuation at $89\%$ of the scored positions; at the others the string's end or the depth bound forces a closer. A scored closer position $t$ predicts $x_{t+1}$. The pairwise contrast is the correct next-token logit minus one seeded incorrect-closer logit, averaged over scored positions. Full-vocabulary accuracy separately measures the next-token argmax.

For natural-context MI on the natural bracket contexts, we recognize plain and space-prefixed brackets, reconstruct the stack, and score stack-consistent closers against same-form distractors. The developmental SGLD objective masks recognized closer token IDs. The reference-KL posterior and intervention comparisons use identical inputs and masks.

Zero-ablation removes a head's pre-output-projection aggregation or an MLP output. Position-conditioned mean replacement averages outputs across examples at each position. The longitudinal bank experiment additionally records a global mean over examples and positions; Figure~\ref{fig:bank} uses its zero-ablation condition. A positive contrast drop denotes a harmful intervention.

\subsection{Developmental SGLD sweep}
We use the unpreconditioned SGLD update of \citet{welling2011sgld} with a component mask and localization term.
The synthetic grid uses 128 strings, length 128, data seed 0, batch size 8, two fp32 chains, and the settings in Table~\ref{tab:hp}. Each chain saves 60 draws after its nominal burn-in. The reported statistic averages \emph{all} updates: 160 for heads and MLPs, 180 for layers, and 210 for the full model. The loss is
\begin{equation}
 \ell_t(\theta;x)=-\mathbf{1}\{x_{t+1}\in V_{\mathrm{close}}\}
                  \log p_\theta(x_{t+1}\mid x_{1:t}),\qquad
 L_B=\frac{1}{B(T-1)}\sum_{b,t}\ell_t(\theta;x_b).
\end{equation}
The all-position denominator makes closer frequency part of the objective's scale. The grid contains 49,588 finite estimates.

\begin{table}[h]
\centering
\begin{tabular}{lrrrr}
\toprule
Restriction & Step size & $n\beta$ & Localization $\gamma$ & Burn-in \\
\midrule
One head & $2.5\times10^{-5}$ & 1.77 & 0.012 & 100 \\
One MLP & $1.5\times10^{-5}$ & 1.77 & 0.025 & 100 \\
One layer & $1.0\times10^{-5}$ & 1.77 & 0.050 & 120 \\
Full model & $5.0\times10^{-6}$ & 1.77 & 0.100 & 150 \\
\bottomrule
\end{tabular}
\caption{Developmental SGLD settings, shared by both Dyck variants. Larger restrictions use manually scaled settings following a head-level calibration at checkpoint 2,000.}
\label{tab:hp}
\end{table}

For displacement $u$ from the checkpoint, the unpreconditioned update is
\begin{equation}
 u_{s+1}=u_s-\frac{\eta}{2}
 \bigl[n\beta\nabla L_{B_s}(u_s)+\gamma u_s\bigr]
 +\sqrt{\eta}\,\xi_s,\qquad \xi_s\sim\mathcal N(0,I).
\end{equation}
Parameter masks restrict both drift and noise. A head includes its slices of $W_Q,W_K,W_V,W_O$ and $b_Q,b_K,b_V$; the shared attention output bias remains fixed. MLP restrictions include input/output weights and biases. The recorded configurations use initial seed 100, matched minibatch order across chains, and cyclic replay of a shuffled first epoch. Appendix~\ref{app:trace-reanalysis} compares averaging conventions and characterizes the finite-horizon sampling response.

A natural-context SGLD sweep covers 12 checkpoints and all 144 components; its cross-distribution associations are reported separately from the reference-KL comparison.

\subsection{Gradient activity and temporal comparisons}
The developmental gradient baseline accumulates gradients of closer-normalized cross-entropy over chunks and then takes the squared block norm:
\begin{equation}
 G_c=\left\|\sum_b\nabla_{\theta_c}\bar L_b\right\|^2.
\end{equation}
Its magnitude reflects chunk normalization and cancellation between gradients. This quantity is labeled \emph{gradient activity} throughout.

Temporal Spearman correlations pair each component's $k=4$ series with the $k=4$ distance-probe trajectory. Rank-time adjustment regresses each ranked series on an intercept and checkpoint rank, then correlates the residuals. First differences measure increments between adjacent checkpoints on the nonuniform training grid.

The early-layer intervention analysis refits distance probes after zero-ablation, measuring recoverable decodability at 27 components and 12 checkpoints with position-level evaluation splits. The frozen-decoder experiment in Appendix~\ref{app:probe-repair} instead uses disjoint strings and a common decoder across conditions.

\subsection{Subspace and group interventions}
Subspace replacement uses $k=4$, depth bound 16, maximum bracket length 48 and fp32 at checkpoint 143,000. Basis reconstruction uses 256 seed-42 fitting strings; evaluations use 128 strings per seed. The named TOKEN, STACK, POSITION, and ANSWER spaces are defined in Appendix~\ref{app:algorithm-detail}. Ten MLPs and fourteen heads receive the designated replacements, with a token lookup at \C{L0.mlp} and position-conditioned means for the eight first-layer heads. Remaining components execute normally. Bases fitted on seed 42 are evaluated without refitting on seeds 43--45. Recovery divides the intervened contrast by the clean contrast on the same batch. The PCA comparator retains 128 directions per MLP and 64 per head. Named-space augmentation adds 64 orthogonal principal directions per component.

The cross-distribution bank experiment uses 128 examples, length 256, seed 42, and fp32, yielding 11,882 synthetic and 187 natural-context scored positions. Its final joint zero-ablation drop is $1.78$. The longitudinal experiment uses 256 synthetic examples, length 128, seed 0, and fp32 at the six checkpoints in Table~\ref{tab:bank-detail}; its final joint drop is $1.56$.

\input{sections/trace_appendix}

%% file: sections/trace_appendix.tex
\section{Averaging sensitivity in the developmental sweep}
\label{app:trace-reanalysis}
We compare estimates averaged over all SGLD updates with estimates using only the final 60 draws. Available traces cover 23,155 unique $k=2$ and 23,102 unique $k=4$ conditions out of 24,794 per variant, including layer and full-model restrictions. Each trace is matched to a measurement by its initial loss, ordered chain coefficients, and restriction configuration.

\paragraph{Effect of the averaging window.}
The $k=4$ atomic subset contains 20,654 paired cells: 142--146 checkpoints per component, with all 144 components represented at 140 checkpoints. On those complete checkpoints, the median correlation between the two component rankings is $0.994$. Retained-only estimates are larger by a median $0.539$; among all-update values above $0.1$, the median ratio is $1.89$. Thus averaging changes the absolute scale substantially while preserving the component ordering.

\begin{table}[h]
\centering
\begin{tabular}{lrrr}
\toprule
Averaging convention & Raw trajectory & Rank-time adjusted & First difference \\
\midrule
All optimization steps & 0.713 & 0.589 & 0.053 \\
Retained draws only & 0.728 & 0.615 & 0.062 \\
\bottomrule
\end{tabular}
\caption{Median componentwise association with the $k=4$ layer-2 distance-probe trajectory on paired observations. First differences use adjacent released checkpoints for which both traces are available. Trace coverage determines the subset of the full trajectory analyzed here.}
\label{tab:trace-paired}
\end{table}

The developmental association persists when averaging only the final 60 draws (Table~\ref{tab:trace-paired}). At ten checkpoints with complete shared coverage, applying this window to the synthetic measurements changes the median cross-distribution correlation from $0.897$ to $0.892$; natural-context measurements use all-update averages. The paired comparison covers the checkpoints with stored traces, which excludes the final $k=4$ checkpoint.

\paragraph{Finite-horizon behavior.}
Within the 60 retained draws, mean loss rises from the first to the last half in $90.5\%$ of the 20,654 atomic cells. The median increase is $0.128$ in all-position-normalized masked NLL. We therefore interpret the developmental sweep as a finite-horizon sampling response. The matched reference-KL experiment uses a separate posterior target and multichain diagnostic protocol.

%% file: sections/locality_appendix.tex
\section{Parameter displacement of SGLD chains}
\label{app:locality}

\subsection{Finite diffusion and parameter displacement}
The SGLD settings in Table~\ref{tab:hp} produce substantial diffusion within the recorded trajectories. With the loss-gradient term removed the localized update is $u_{s+1}=(1-\eta\gamma/2)u_s+\sqrt\eta\,\xi_s$ with $\xi_s\sim\mathcal N(0,I)$, whose per-coordinate variance after $s$ updates is $\eta\sum_{j<s}(1-\eta\gamma/2)^{2j}\approx\eta s$ when $\eta\gamma s\ll1$; for a head this gives an endpoint standard deviation close to $\sqrt{160(2.5\times10^{-5})}=0.0632$ per coordinate. The accumulated localization factor is $\eta\gamma S/2=2.4\times10^{-5}$. Across all 144 components and 154 checkpoints, the median ratio of the predicted endpoint standard deviation to the RMS of an individual weight matrix is $2.59$. Bias tensors are excluded from this relative-scale comparison.

We directly measure the dynamics at steps 0, 2,000, 32,000, and 143,000 for \C{L0H0}, \C{L0H2}, \C{L7H4}, \C{L11.mlp}, \C{L14H1}, and the full model. Each of these 24 conditions uses two chains under three protocols: the documented SGLD settings; a control with the loss-gradient term removed; and strongly localized SGLD with $\eta=10^{-7}$ and $\gamma=10^6$. The first two protocols share every Gaussian increment. We execute the optimizer implementation of the reference library \citep{devinterp} and record the applied loss-gradient, localization, and noise terms separately.

\begin{figure}[t]
\centering
\includegraphics[width=\linewidth]{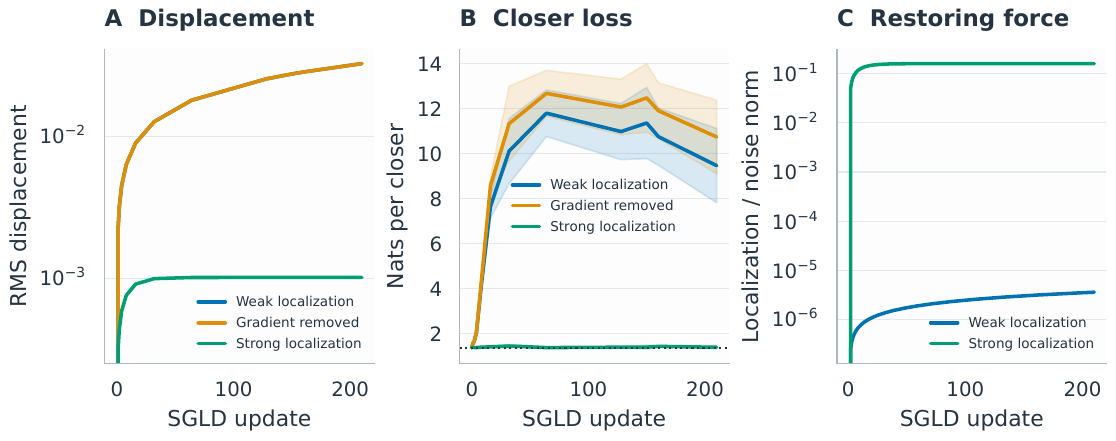}
\caption{\textbf{Parameter diffusion and its predictive effect.} Full-model perturbations at checkpoint 143,000 under the documented SGLD settings, a matched control with the loss-gradient term removed, and strongly localized SGLD. Lines average two chains; shading spans their values. Gaussian increments are identical between the first two protocols.}
\label{fig:sgld-locality}
\end{figure}

The median endpoint displacement relative to weight-matrix RMS is $4.76$ under the documented settings and $0.078$ under strong localization. Holding a chain at a displacement of $0.05$ RMS requires $\gamma\approx7\times10^5$ (the stationary per-coordinate standard deviation is $\gamma^{-1/2}$), seven orders of magnitude above the sweep's value, at which the loss-gradient term is negligible and the chain samples the localizing prior. The median loss-gradient/noise norm ratio per update is $9.8\times10^{-6}$; the corresponding localization/noise ratio is $1.3\times10^{-6}$. Nevertheless, the loss-gradient term measurably changes the predictive response. At the final full-model checkpoint, the clean closer loss is $1.379$ nats, the mean sampled loss is $10.562$ under SGLD, and it is $11.594$ with that term removed. Strong localization gives $1.408$. These finite-path controls establish the perturbation scale; equilibrium posterior measurements use the separate protocol below (Figure~\ref{fig:sgld-locality}).

\subsection{Standard settings}
\label{app:sgld-standard}
The same 24 conditions were rerun with the settings of the reference implementation \citep{devinterp}: localization $\gamma=100$ centered at the checkpoint, inverse temperature $n\beta=n/\log n=636$ with $n=5{,}475$ the scored closer positions of the batch (and $n\beta=1{,}700$), step sizes $\eta\in\{10^{-5},10^{-4}\}$, 160 updates, minibatches of eight strings, two chains sharing their Gaussian increments across settings. Table~\ref{tab:sgld-standard} summarizes the endpoints. No chain stays inside $U$: the displacement is set by the noise term, $\sqrt{\eta s}$ per coordinate, and is the same at both inverse temperatures, while the loss-gradient drift, one hundred times stronger than under the documented settings but still small relative to the noise, fine-tunes the model on the batch, so that the mean loss over the chain falls below the clean loss and the estimator's own statistic $n\beta\,(\overline L-L_{\mathrm{clean}})$ is negative for 21--24 of the 24 conditions ($-172$ and $-544$ in median at the two inverse temperatures for $\eta=10^{-5}$); at $\eta=10^{-4}$ the full model and the MLPs diverge. Localization strong enough to hold a chain inside $U$ ($\gamma\approx7\times10^5$, as above) leaves only the prior, so the region has to be imposed on the draws rather than reached by tuning the chain.

\begin{table}[t]
\centering\footnotesize
\setlength{\tabcolsep}{4pt}
\resizebox{\linewidth}{!}{\begin{tabular}{@{}lcccccc@{}}
\toprule
Setting & displ./RMS & inside $U$ & closer KL & grad/noise & loc/noise & full-model CE \\
\midrule
$\eta=10^{-5}$, $n\beta=636$ & $2.96$ & 0/48 & $1.56$ & $7.8\times10^{-4}$ & $4.4\times10^{-3}$ & $0.81$ \\
$\eta=10^{-4}$, $n\beta=636$ & $6.88$ & 0/48 & $3.20$ & $2.3\times10^{-3}$ & $3.7\times10^{-2}$ & $3.42$ \\
$\eta=10^{-5}$, $n\beta=1{,}700$ & $2.96$ & 0/48 & $2.24$ & $1.4\times10^{-3}$ & $4.4\times10^{-3}$ & $1.12$ \\
$\eta=10^{-4}$, $n\beta=1{,}700$ & $6.88$ & 0/48 & $3.53$ & $4.1\times10^{-3}$ & $3.7\times10^{-2}$ & $7.24$ \\
documented sweep ($n\beta=1.77$) & $4.76$ & 0/48 & $3.28$ & $9.8\times10^{-6}$ & $1.3\times10^{-6}$ & $10.56$ \\
strong localization ($\eta=10^{-7}$, $\gamma=10^6$) & $0.08$ & 48/48 & $1.6\times10^{-4}$ & & & $1.41$ \\
\bottomrule
\end{tabular}}
\caption{\textbf{SGLD endpoints under standard, documented, and strongly localized settings.} Medians over the 24 conditions and two chains: endpoint displacement relative to weight-matrix RMS, chains whose endpoint lies inside $U$, closer KL from the clean checkpoint, per-update norm ratios of the loss-gradient and localization terms to the noise, and the closer cross-entropy of the full-model chain at the final checkpoint (clean $1.379$).}
\label{tab:sgld-standard}
\end{table}

\subsection{Clean-loss and parameter-scale controls}
\input{reproducibility/development/table}
With the covariates of Section~\ref{sec:correspondence} (clean loss, weight scale, and training step), the finite SGLD response's partial association with the opener-distance probe over the 148 confirmation checkpoints is $0.46$ in levels ($0.47$ with $A_c$ added) and $0.02$ in adjacent changes (medians over the 144 components), the values quoted in Section~\ref{sec:correspondence}.

The stored initial loss agrees with independently evaluated closer cross-entropy, after applying the all-position normalization, to within $7.8\times10^{-7}$. The scored fractions are $0.342581$ for $k=2$ and $0.336799$ for $k=4$. We reconstruct the mean perturbed loss as $\overline L=L_{\mathrm{clean}}+R/(n\beta)$ and compare the full-model response with $n\beta(f\log V-L_{\mathrm{clean}})$, where $V=50{,}304$ and $f$ is the closer fraction.

\begin{figure}[t]
\centering
\includegraphics[width=\linewidth]{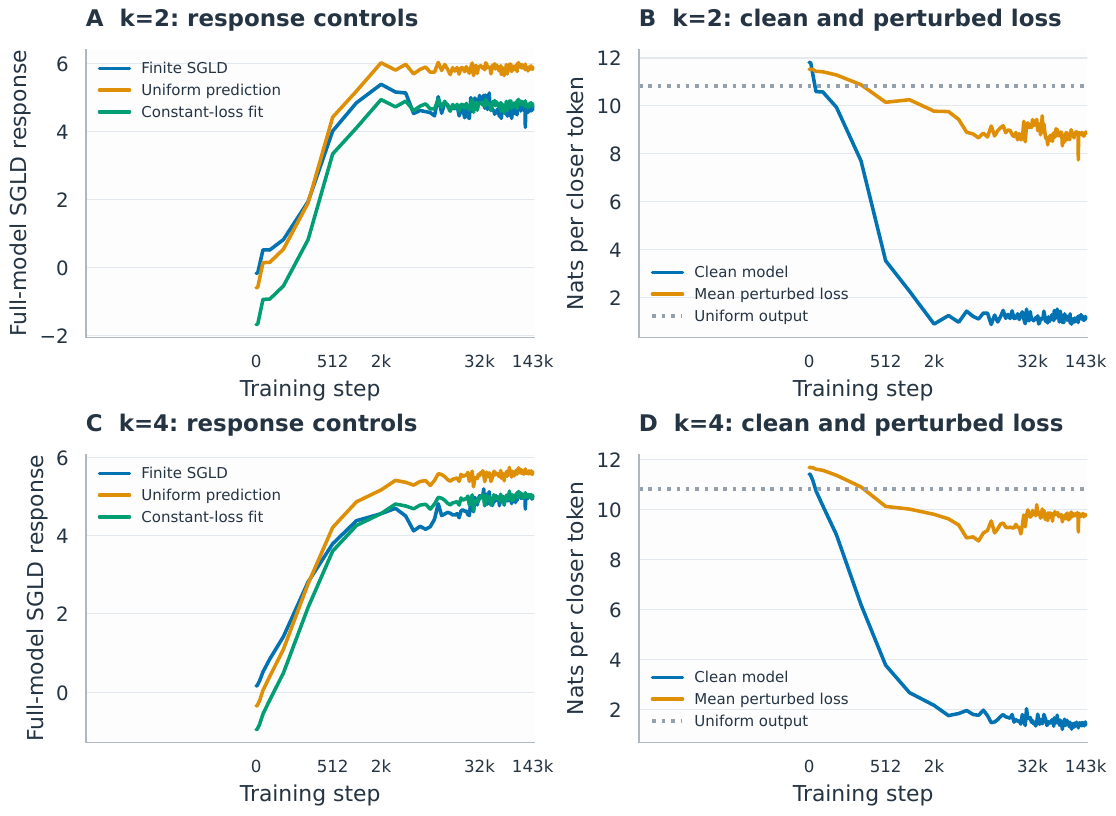}
\caption{\textbf{Full-model response and clean-loss controls.} Left: observed finite SGLD response, the unfitted uniform-output prediction, and a descriptive fit with constant mean perturbed loss. Right: clean loss and mean perturbed loss reconstructed from the stored response. Every panel contains all 154 checkpoints.}
\label{fig:full-model-loss-control}
\end{figure}

Across all 154 checkpoints, Pearson correlations between the response and negative clean loss are $0.989$ and $0.991$ for $k=2$ and $k=4$. The unfitted uniform-output predictions have $R^2=-0.110$ and $0.602$. Fitting a constant mean perturbed loss gives descriptive $R^2=0.855$ and $0.918$, with implied closer losses of $9.04$ and $9.81$ nats. The response is thus closely associated with clean loss without being identical to the uniform-output prediction. Restricting to checkpoints at or after step 2,000 reduces the constant-loss fit to $R^2=0.182$ and $0.486$ (Figure~\ref{fig:full-model-loss-control}).

At component resolution, the negative-clean-loss association with the primary $k=4$ distance target is $-0.111$ after rank-time adjustment. Controlling both checkpoint rank and clean-loss rank leaves a median SGLD association of $0.605$. Inverse weight-matrix RMS has an adjusted association of $0.509$; adding each component's weight RMS to the controls reduces the SGLD association to $0.456$ ($0.237$ for $k=2$). These results separate the full-model clean-loss effect from the component response and its dependence on perturbation scale.

%% file: reproducibility/development/table.tex
\begin{table}[t]
\centering
\small
\begin{tabular}{lrrr}
\toprule
Measurement & Trajectory & Rank-time adjusted & Adjacent change \\
\midrule
Finite SGLD response & $0.75$ & $0.60$ & $0.05$ \\
Empirical Fisher (raw) & $0.39$ & $0.04$ & $0.06$ \\
Model Fisher (raw) & $0.38$ & $-0.08$ & $-0.02$ \\
Gradient activity (raw) & $0.27$ & $-0.01$ & $-0.03$ \\
Zero ablation & $0.15$ & $0.02$ & $0.00$ \\
Clean closer loss & $-0.37$ & $0.11$ & $-0.12$ \\
\bottomrule
\end{tabular}
\caption{\textbf{Tracking feature development across all 154 checkpoints.} Median within-component association with the fixed $k=4$ layer-2 prefix-state distance probe. Rank-time adjustment removes the linear checkpoint-rank trend. Adjacent changes use log Fisher and gradient values. Component measurements cover the same 144 components; clean loss is one shared task-level series. The finite SGLD response uses the nonlocal perturbation regime in Appendix~\ref{app:locality}.}
\label{tab:development-baselines}
\end{table}

%% file: sections/matched_gibbs_findings.tex
\subsection{A matched comparison of circuit geometry and causal effects}
\label{sec:matched-gibbs}
\input{reproducibility/fresh_gibbs/derived/metrics}

At the final checkpoint, geometric and causal measurements use the same full-vocabulary reference KL at identical closer positions. Each frozen batch contains 128 sequences: 5,493 synthetic $k=4$ and 177 natural-context closer targets. We measure
\begin{equation}
 q_C=\tau\E_{\pi_C}[K_C(u)],\qquad
 \pi_C(u)\propto e^{-\tau K_C(u)}\mathcal N(u;0,r^2I),
\end{equation}
where $K_C$ is Equation~\eqref{eq:component-kl} for component $C$ in current-checkpoint RMS coordinates. We use $r=0.1$, $\tau=1{,}000$, and prior-preserving Metropolis sampling of the complete fixed batch \citep{cotter2013mcmc}. All 288 target responses pass the four-chain diagnostic screen \citep{vehtari2021rhat}. Zero or mean output replacement gives the matched MI scores. The estimate rank-correlates with the model Fisher trace in the same coordinates (a Hutchinson estimate with 32 probes) at $0.997$ on the task inputs and $0.999$ on natural contexts, as Equation~\eqref{eq:soft-rank} predicts at this temperature.

\begin{figure}[t]
\centering
\includegraphics[width=\linewidth]{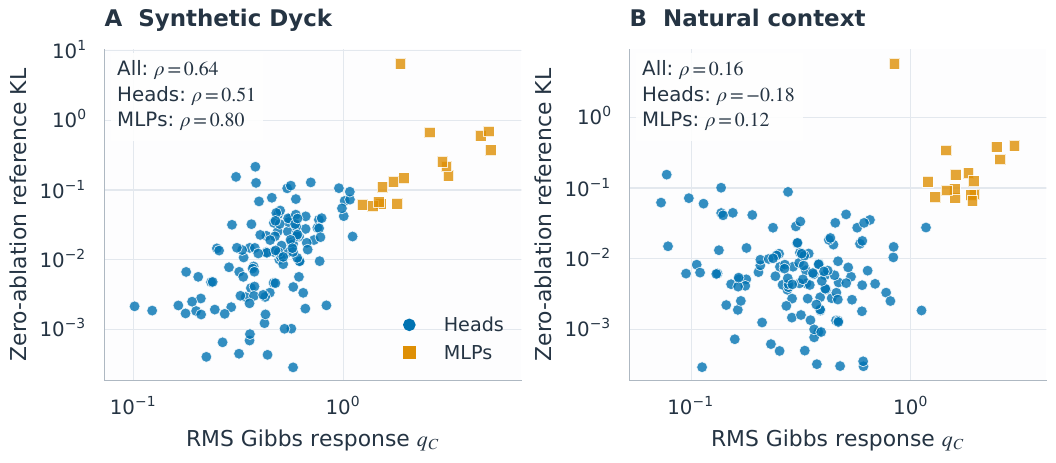}
\caption{\textbf{Geometric sensitivity and causal circuit effects measured on the same behavior.} Each point is one component: attention heads are blue and MLPs orange. The horizontal axis is the LLC estimate under component-restricted parameter perturbations; the vertical axis is reference-model KL under zero replacement of that component's output. Both measurements use identical inputs and closer positions. Insets report Spearman correlations over all components and within each component class.}
\label{fig:matched-gibbs}
\end{figure}

On synthetic bracket completion, the LLC estimate tracks component effects: its Spearman association with zero-ablation KL is $\GibbsSynZero$ over all components, $\GibbsSynHeadZero$ within the 128 heads, and $\GibbsSynMlpZero$ within the 16 MLPs (Figure~\ref{fig:matched-gibbs}). Mean replacement gives an overall association of $\GibbsSynMean$. The correspondence therefore extends within both component classes. Natural-context associations are weaker: $\GibbsNatZero$ for zero replacement and $\GibbsNatMean$ for mean replacement, with head-only values of $\GibbsNatHeadZero$ and $\GibbsNatHeadMean$.

The matched panel also separates dependence on the input distribution from dependence on the measurement. The cross-distribution rank correlation is $\GibbsCross$ for the LLC estimate, $\MIZeroCross$ for zero-ablation KL, and $\MIMeanCross$ for mean-ablation KL. Appendix~\ref{app:matched-gibbs} gives the complete comparisons for conditional KL, target cross-entropy, pairwise closer contrast, and accuracy.

Across intervention strengths $\epsilon=0.01$ to $1$, the synthetic association remains approximately $0.64$, while the natural-context association changes from $0.19$ to $0.16$. KL scales quadratically at the smallest strengths. The disagreement thus persists locally, distinguishing directional sensitivity from the LLC estimate's aggregation over parameter directions (Appendix~\ref{app:intervention-scale}).

\paragraph{A coordinate control.}
\input{reproducibility/fresh_gibbs/raw_comparison/metrics}
Repeating all 288 posteriors with an isotropic prior in the fixed converted TransformerLens tensors gives zero-ablation associations of $\RawGibbsSynZero$ on synthetic inputs and $\RawGibbsNatZero$ on natural contexts. This prior has absolute standard deviation $0.001$, while the primary prior uses relative RMS radius $0.1$, changing both parameter weighting and perturbation scale. Their causal alignment therefore depends on the specified parameter metric and localization (Appendix~\ref{app:raw-coordinate}).

%% file: reproducibility/fresh_gibbs/derived/metrics.tex
\providecommand{\GibbsSynZero}{0.64}
\providecommand{\GibbsNatZero}{0.16}
\providecommand{\GibbsSynMean}{0.64}
\providecommand{\GibbsNatMean}{0.14}
\providecommand{\GibbsCross}{0.61}
\providecommand{\MIZeroCross}{0.67}
\providecommand{\MIMeanCross}{0.59}
\providecommand{\GibbsSynHeadZero}{0.51}
\providecommand{\GibbsSynMlpZero}{0.80}
\providecommand{\GibbsNatHeadZero}{-0.18}
\providecommand{\GibbsNatHeadMean}{-0.19}

%% file: reproducibility/fresh_gibbs/raw_comparison/metrics.tex
\providecommand{\RawGibbsSynZero}{0.861}
\providecommand{\RawGibbsNatZero}{0.561}
\providecommand{\RawGibbsSynFisher}{0.999}
\providecommand{\RawGibbsNatFisher}{0.999}

%% file: sections/sensitivity_findings.tex
\paragraph{Matched joint geometry.}
Across checkpoints 8,000, 32,000, and 143,000, the bank's matched LLC estimate increases $1.07\to1.80\to3.90$ on synthetic inputs and $0.20\to0.35\to0.91$ on natural-context inputs. At the final checkpoint, the joint synthetic response is $5.8\%$ below the sum of the eight head responses ($3.90$ versus $4.14$). The gap exceeds a conservative Monte Carlo standard error (MCSE) bound accounting for shared seeds by a factor of $5.7$. This joint deficit matches the coupling prediction of Equation~\eqref{eq:joint-geometry}; a separable local loss gives exact additivity. Natural-context joint and summed responses differ by less than $0.7\%$ across all three checkpoints (Appendix~\ref{app:jointgeometry}).
Across the fixed 18-component panel, factor-two changes in radius or inverse temperature preserve ranks ($\rho=0.975$--$1.000$); median responses follow $\tau r^2$ scaling within $3.6\%$. Changing the proposal coefficient reproduces all 36 component responses within three conservative MCSE bounds (Appendix~\ref{app:sensitivity}).

%% file: sections/gibbs_methods.tex
\section{Matched localized posterior measurements}
\label{app:gibbs-methods}

The matched experiment evaluates the parameter geometry and causal responses of the same model on the same frozen inputs and output readout. At checkpoint 143,000, we measure every attention head and MLP on synthetic and natural-language-context bracket data. This experiment uses a reference-model KL objective and a precisely specified localized posterior. We denote its response by $q_c$, and the developmental sweep's finite SGLD response by $R_c$.

\subsection{Reference loss and parameter chart}

We use TransformerLens 3.1.0 \citep{nanda2022transformerlens} with its standard LayerNorm folding, value-bias folding, residual-write centering, and unembedding centering. The parameter vector $\theta$ denotes these converted tensors. Raw coordinates apply equal per-coordinate weighting to this fixed parameterization; RMS coordinates rescale its tensors as below. Value-bias folding places the pretrained constant head writes in the shared attention output bias, which remains fixed in both the head interventions and the parameter restrictions. All matched comparisons use this same component decomposition.
Let $\theta_\star$ denote the checkpoint parameters and let $c$ select a component with $d_c$ parameters. We write
\begin{equation}
 \theta(u)=\theta_\star+D_cu,\qquad u\in\mathbb R^{d_c},
 \label{eq:gibbs-chart}
\end{equation}
where $D_c$ embeds the component coordinates into the full parameter space and rescales each selected tensor by its pretrained root-mean-square value. For a selected tensor $j$, its scale is
\begin{equation}
 s_j=\max\!\left\{
  \sqrt{\frac{1}{d_j}\|\theta_{\star,j}\|_2^2},\;
  10^{-3}\operatorname{median}_{k\in\mathcal W_c}
  \sqrt{\frac{1}{d_k}\|\theta_{\star,k}\|_2^2}
 \right\},
 \label{eq:gibbs-rms-chart}
\end{equation}
with $\mathcal W_c$ the selected weight matrices. The floor gives zero or very small tensors a positive perturbation scale. All scales are fixed before sampling and depend exclusively on the checkpoint parameters. A head restriction selects its slices of $W_Q,W_K,W_V,W_O,b_Q,b_K,b_V$; the shared attention output bias remains fixed. An MLP restriction selects $W_{\mathrm{in}},W_{\mathrm{out}},b_{\mathrm{in}},b_{\mathrm{out}}$.

Let $\mathcal A_D$ be the frozen set of scored closer positions in distribution $D$, and let $x_a$ denote the token prefix at position $a$. The primary loss is
\begin{equation}
 K_{c,D}(u)
 =\frac{1}{|\mathcal A_D|}\sum_{a\in\mathcal A_D}
 \KL\!\left(
 p_{\theta_\star}(\cdot\mid x_a)
 \,\middle\|\,
 p_{\theta(u)}(\cdot\mid x_a)
 \right),
 \label{eq:gibbs-reference-kl}
\end{equation}
where both distributions range over the full vocabulary. Thus $K_{c,D}(0)=0$, and the checkpoint is a minimum of the reference loss. Full-vocabulary KL registers changes both in the total probability assigned to brackets and in the relative probabilities of bracket types.

For coordinate scale $r>0$ and inverse temperature $\tau>0$, define
\begin{equation}
 \pi_{c,D}(du)
 =\frac{1}{Z_{c,D}(\tau,r)}
   e^{-\tau K_{c,D}(u)}\,N(du;0,r^2I),
 \qquad
 q_{c,D}(\tau,r)=\tau\mathbb E_{\pi_{c,D}}K_{c,D}(u).
 \label{eq:gibbs-target-response}
\end{equation}
The scalar $r$ is the prior standard deviation of each coordinate: $\mathbb E\|u\|^2/d_c=r^2$. The quantity $q_{c,D}$ is the LLC estimate of Equation~\eqref{eq:estimator}. Equivalently, $q_{c,D}=-\tau\,\partial_\tau\log Z_{c,D}$ at fixed $r$, connecting the measured loss response to the local partition function.

\subsection{Connection to circuit-sensitive directions}
The reference KL has zero gradient at $u=0$. Its quadratic expansion is
\begin{equation}
 K_{c,D}(u)=\tfrac12u^\top H_{c,D}u+o(\|u\|^2),
 \qquad H_{c,D}=D_c^\top F_DD_c\succeq0,
 \label{eq:gibbs-local-fisher}
\end{equation}
where $F_D$ is the categorical Fisher metric pulled back through the model outputs on $\mathcal A_D$. Consequently, $H_{c,D}$ measures the component's task-sensitive directions in the fixed coordinates of Equation~\eqref{eq:gibbs-chart}.

For the quadratic model $K_{c,D}(u)=u^\top H_{c,D}u/2$, the posterior is Gaussian with covariance
\begin{equation}
 \Sigma_{c,D}=(\tau H_{c,D}+r^{-2}I)^{-1}.
\end{equation}
Taking the expectation of the quadratic form gives
\begin{align}
 q_{c,D}(\tau,r)
 &=\tfrac12\operatorname{tr}\!\left[
    \tau H_{c,D}(\tau H_{c,D}+r^{-2}I)^{-1}\right]
 \label{eq:gibbs-quadratic-response}\\
 &=\tfrac12\sum_{i=1}^{d_c}
    \frac{\tau r^2\mu_i}{1+\tau r^2\mu_i},
 \nonumber
\end{align}
where $\mu_i$ are the eigenvalues of $H_{c,D}$. Each positive-curvature direction contributes between zero and one half, and null directions contribute zero. In this quadratic setting, increasing $\tau r^2$ makes the response approach $\operatorname{rank}(H_{c,D})/2$. At finite localization, the response weights the same task-sensitive directions by their curvature. The scale and temperature therefore specify the geometric measurement, while the proposal parameter below determines how efficiently it is sampled.

\subsection{Prior-preserving Metropolis sampling}
We sample Equation~\eqref{eq:gibbs-target-response} using the preconditioned Crank--Nicolson (pCN) proposal \citep{cotter2013mcmc}. Given the current state $u$, propose
\begin{equation}
 u'=\sqrt{1-a^2}\,u+ar\xi,
 \qquad \xi\sim N(0,I),\qquad 0<a\leq1,
 \label{eq:gibbs-pcn-proposal}
\end{equation}
and accept with probability
\begin{equation}
 A(u,u')=\min\!\left\{1,
   \exp\!\left[-\tau\bigl(K_{c,D}(u')-K_{c,D}(u)\bigr)\right]
 \right\}.
 \label{eq:gibbs-pcn-acceptance}
\end{equation}
This transition preserves the specified posterior. To see this, set $b=\sqrt{1-a^2}$. Under the Gaussian prior, the pair $(u,u')$ has covariance
\begin{equation}
 r^2\begin{pmatrix} I&bI\\ bI&I\end{pmatrix},
\end{equation}
which is unchanged by exchanging the two states. The proposal is therefore reversible with respect to the prior, and its Gaussian density terms cancel in the Metropolis ratio. Multiplying by the likelihood factor $e^{-\tau K_{c,D}}$ yields Equation~\eqref{eq:gibbs-pcn-acceptance} and detailed balance for $\pi_{c,D}$.

Each proposal evaluates the complete frozen empirical loss. Each chain begins with an independent draw from $N(0,r^2I)$; rejected proposals retain the preceding state and its metrics. We use $r=0.1$, $\tau=1000$, and $a=0.9$, with four chains, 256 burn-in transitions, and 512 retained transitions per chain. Chain $c\in\{0,1,2,3\}$ uses seed $20260918+100003c$. The settings were chosen from acceptance and loss-trace diagnostics on 32-sequence pilots of \C{L0H0} and \C{L11.mlp} on both distributions, before the component-association analysis.

\subsection{Matched data, observables, and interventions}
The final datasets each contain 128 sequences of length 128. The synthetic and natural-context datasets contain 5,493 and 177 scored closer positions, respectively. Tokens, closer masks, correct and distractor tokens, and same-surface-form closer options are frozen in the dataset archive. Both sampling and MI interventions use these exact arrays and normalize every metric by the number of scored positions.

In addition to full-vocabulary KL, every sampling transition records closer-conditional KL, full-vocabulary cross-entropy, and closer-conditional cross-entropy. The conditional distribution normalizes over the four closing-bracket tokens with the observed target's surface form. These are observables under the primary full-KL posterior. For any recorded loss $f$, the corresponding scaled response is
\begin{equation}
 R_f=\tau\bigl(\mathbb E_{\pi_{c,D}}f(\theta(u))-f(\theta_\star)\bigr).
\end{equation}
A separately designated closer-conditional-KL run substitutes that objective into the posterior itself. This distinguishes changes in the measured observable from changes in the distribution over parameter perturbations.

The matched MI panel measures zero and position-mean replacement for each of the 144 components, together with joint first-layer attention and named-subspace interventions. Position means average the clean activations across all input sequences at each position, including padded rows. A head intervention replaces its pre-output-projection activation; linearity of the fixed $W_O$ makes this equivalent to replacing its projected residual-stream write. The panel records the same four losses together with pairwise closer contrast and full-vocabulary accuracy.

\subsection{Multichain diagnostics and numerical verification}
For each recorded observable, we calculate the maximum of rank-normalized split and folded-split $\widehat R$, together with mean, bulk, and tail effective sample sizes using ArviZ \citep{kumar2019arviz} \citep{vehtari2021rhat}. The prespecified target screen requires four chains, $\widehat R<1.01$, and all three effective sample sizes at least 400. Auxiliary observables and the mean squared coordinate have separately reported diagnostics. Mean Monte Carlo standard errors use the effective sample size of the raw observable; the standard error of a scaled response is multiplied by $\tau$. These errors quantify simulation uncertainty conditional on the checkpoint and frozen evaluation inputs.

When a target fails this screen, its diagnostic extension uses 1,024 retained transitions with unchanged seeds, burn-in, target, and sampler settings. Replaying the same random-number sequence extends the initial trajectory, and both outputs are retained. Extension decisions use sampling diagnostics. The saved trace places the initial prior draw at index zero; the primary retained window occupies indices 257 through 768, inclusive.

Model computation uses float32 with TF32 disabled, while log probabilities and KL calculations use float64. We compare the selected-position readout with complete model outputs. Computation before the restricted layer is cached, and complete scored log probabilities from cached and full forward passes are checked at the clean and first perturbed states with tolerance $10^{-5}$. Every cell records its parameter coordinates, seeds, proposal metrics, retained and rejected states, and diagnostics (Table~\ref{tab:estimands}).

\begin{table}[t]
\centering\small
\begin{tabularx}{\linewidth}{@{}>{\raggedright\arraybackslash}p{.22\linewidth}>{\raggedright\arraybackslash}X>{\raggedright\arraybackslash}X@{}}
\toprule
Measurement & Quantity & View of the circuit \\
\midrule
Causal intervention & Reference KL after zero or mean replacement & Dependence on a component's activation \\
Empirical Fisher & Mean squared per-token loss-gradient norm & Sensitivity to the observed task labels \\
CE Hessian & Signed trace of task-loss curvature & Local curvature at current performance \\
Model Fisher / KL Hessian & Curvature of reference-model KL & Local change in the task output distribution \\
LLC estimate & $q_C=\tau\E_{\pi_C}K_C$ & Task constraints across nearby parameter settings \\
\bottomrule
\end{tabularx}
\caption{\textbf{Matched measurements of the circuit.} All final-checkpoint comparisons use the same inputs, scored positions, and component definitions. Derivative baselines are measured in both raw parameter coordinates and the pretrained-RMS coordinates of the posterior experiment.}
\label{tab:estimands}
\end{table}

%% file: sections/matched_gibbs_appendix.tex
\section{Complete matched geometric and intervention comparisons}
\label{app:matched-gibbs}

The primary comparison uses full-vocabulary reference KL for both the localized posterior and the intervention. We additionally evaluate five output metrics on the same closer positions: KL conditional on the four closer types sharing the target's tokenization, full-vocabulary target cross-entropy, conditional target cross-entropy, the seeded pairwise closer contrast, and full-vocabulary closer accuracy. Higher importance denotes larger KL, increased cross-entropy, or decreased contrast or accuracy. The posterior response $q_C$ remains the response under the full-KL target throughout these comparisons.

\begin{figure}[t]
\centering
\includegraphics[width=\linewidth]{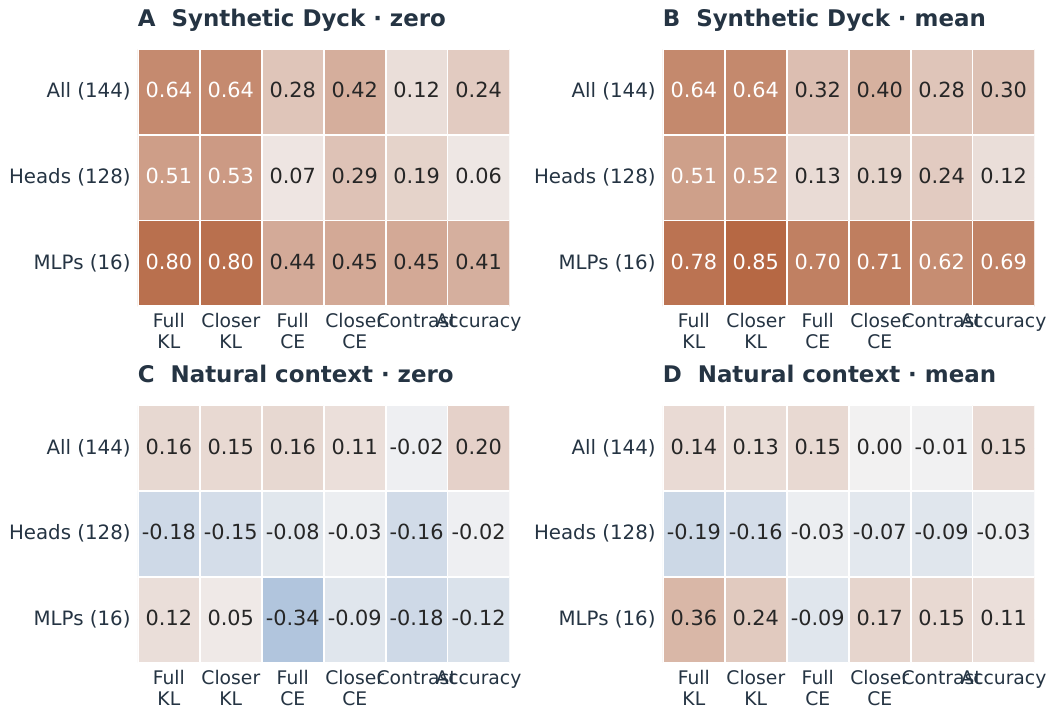}
\caption{\textbf{Complete matched measurement comparison.} Cells show the Spearman correlation between the full-KL LLC estimate and each intervention score. Positive importance uses increased KL or cross-entropy and decreased contrast or accuracy. All 144 components, 128 heads, and 16 MLPs are reported separately. Zero and position-conditioned mean replacements act on identical components and evaluation positions.}
\label{fig:matched-all-metrics}
\end{figure}

\begin{figure}[t]
\centering
\includegraphics[width=\linewidth]{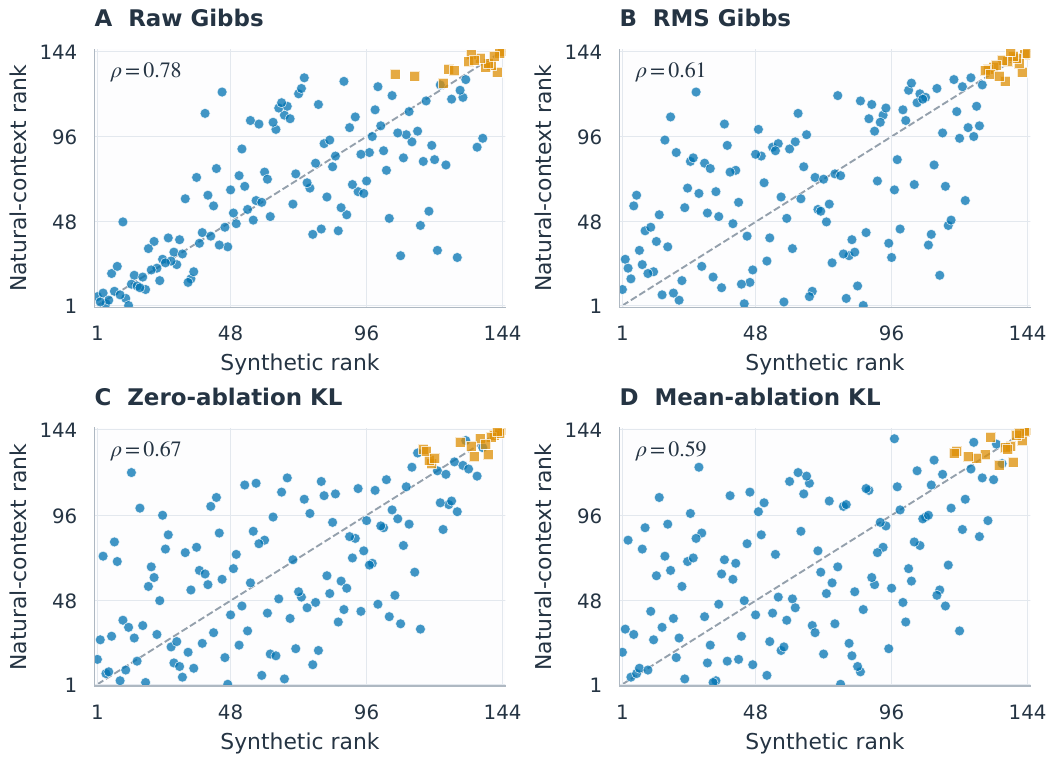}
\caption{\textbf{Ranking stability across distributions with matched objectives.} The synthetic and natural-context ranks of each component are compared for both raw- and RMS-coordinate LLC estimates and full-KL intervention effects. Blue points are heads and orange points are MLPs. All panels use the same 144 components; ties receive average ranks.}
\label{fig:matched-cross}
\end{figure}

Figure~\ref{fig:matched-all-metrics} reports all six metrics under both replacement conventions and both distributions. Each comparison includes the complete component class indicated on its row. These stratified comparisons distinguish within-class correspondence from a difference in scale between attention heads and MLPs. Figure~\ref{fig:matched-cross} compares the ranking of the same components across distributions, with full-KL interventions as the matched causal reference.

For every condition we record the mean response, Monte Carlo standard error, chain means, convergence diagnostics, and whether the estimate uses 512 draws or a diagnostic extension. Extensions preserve the target, prior, parameter coordinates, chain seeds, and initial transition sequence and are selected by sampling diagnostics alone. Component correlations describe the fixed model and evaluation batches; Monte Carlo errors quantify posterior sampling uncertainty.

\subsection{Raw-coordinate posterior control}
\label{app:raw-coordinate}
We evaluate all 144 components on both distributions with an isotropic Gaussian prior of absolute standard deviation $\sigma=0.001$ in the converted TransformerLens tensors. Reference KL, inputs, output masks, inverse temperature $\tau=1{,}000$, proposal coefficient, and the four-chain protocol match the RMS experiment. All 288 conditions satisfy the diagnostic criteria: 287 use 512 retained transitions per chain, and \C{L1H0} on natural contexts uses a 1,024-transition extension that preserves the initial trajectory.

\begin{figure}[t]
\centering
\includegraphics[width=\linewidth]{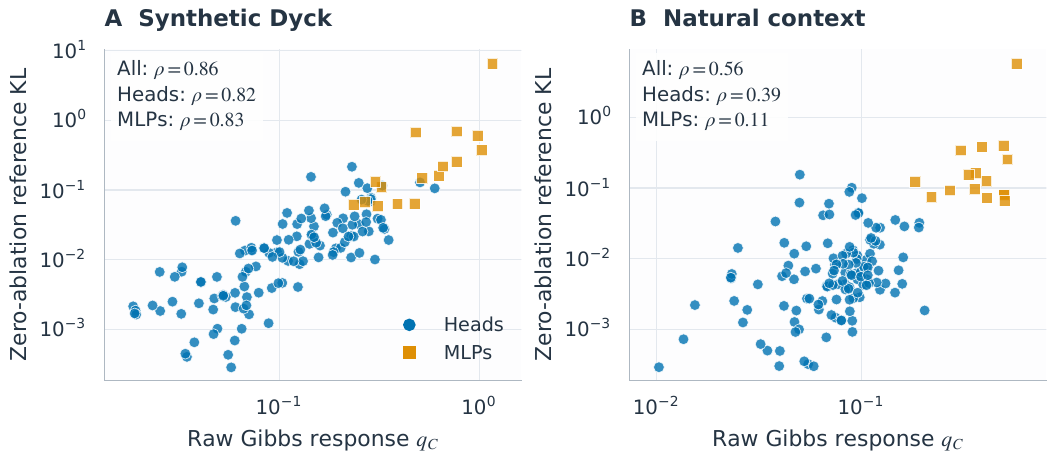}
\caption{\textbf{Raw-coordinate geometry and causal effects.} The isotropic raw prior uses absolute standard deviation $0.001$ in the fixed converted parameterization. Each point is one component; blue denotes attention heads and orange MLPs. Insets give Spearman correlations over all components and within each component class. The causal scores and evaluation inputs are identical to Figure~\ref{fig:matched-gibbs}.}
\label{fig:raw-gibbs}
\end{figure}

Figure~\ref{fig:raw-gibbs} reports the full comparison and both component classes. The raw posterior closely follows the independently measured raw-coordinate model Fisher: Spearman correlations are $0.9992$ on synthetic inputs and $0.9986$ on natural contexts. The componentwise median ratio to the linear-trace prediction $\tau\sigma^2\operatorname{tr}(F_C)/2$ is $1.004$ and $1.007$, respectively. Across components, the corresponding fifth-to-ninety-fifth percentile ranges are $[0.957,1.042]$ and $[0.983,1.069]$. These measurements connect the raw-coordinate response to the small-curvature regime of Equation~\eqref{eq:finite-bridge}.

Raw and RMS posterior responses have component-rank correlations of $0.722$ on synthetic inputs and $0.556$ on natural contexts. The priors differ in directional weighting and absolute scale while sharing the circuit, behavioral discrepancy, and inputs.

\subsection{Named circuit variables on the posterior evaluation inputs}
We evaluate STACK and ANSWER interventions on the posterior-evaluation datasets (Table~\ref{tab:matched-named-groups}). We remove the rank-24 STACK span in layers 1--5 or the rank-3 ANSWER span from the answer-writing components at every sequence position. Both interventions reduce closer accuracy on synthetic and natural-context inputs. Each batch contains 128 strings of length 128; the circuit-identification and held-out subspace experiments use length-48 strings. These measurements assess the same circuit variables across sequence lengths and input contexts.

\begin{table}[t]
\centering\small
\begin{tabular}{lrrrr}
\toprule
Inputs & Closer targets & Clean & Remove STACK & Remove ANSWER \\
\midrule
Synthetic & 5,493 & $60.1\%$ & $20.9\%$ & $34.1\%$ \\
Natural context & 177 & $83.6\%$ & $58.2\%$ & $76.8\%$ \\
\bottomrule
\end{tabular}
\caption{Full-vocabulary closer accuracy under STACK and ANSWER removal on the posterior-evaluation batches. Subspace bases are fitted on synthetic strings and held fixed across evaluation conditions.}
\label{tab:matched-named-groups}
\end{table}

\subsection{Intervention strength and direction}
\label{app:intervention-scale}
For every component, we replace its output $h_C(x)$ by
\begin{equation}
 h_C^{(\epsilon)}(x)=(1-\epsilon)h_C(x)+\epsilon r_C(x),
 \qquad \epsilon\in\{0.01,0.03,0.1,0.3,1\},
\end{equation}
where $r_C$ is zero or the position-conditioned mean. Inputs, masks, reference predictions, and replacement means match the full-replacement comparison. We evaluate 2,304 partial-replacement conditions alongside the 576 full-replacement endpoints, recording all six output metrics and per-string numerators.

In the local regime, writing $v_C=r_C-h_C$ for the stacked activation perturbation and $H_C^{a}$ for the downstream pullback of the categorical Fisher metric gives
\begin{equation}
 \lim_{\epsilon\to0}\frac{K_\epsilon}{\epsilon^2}
 =\frac12\E_D\!\left[v_C(x)^\top H_C^{a}(x)v_C(x)\right].
\end{equation}
This is a directional quadratic form; Equation~\eqref{eq:finite-bridge} instead aggregates curvature over parameter directions. The two measurements are distinct contractions of a shared behavioral metric.

\begin{figure}[t]
\centering
\includegraphics[width=\linewidth]{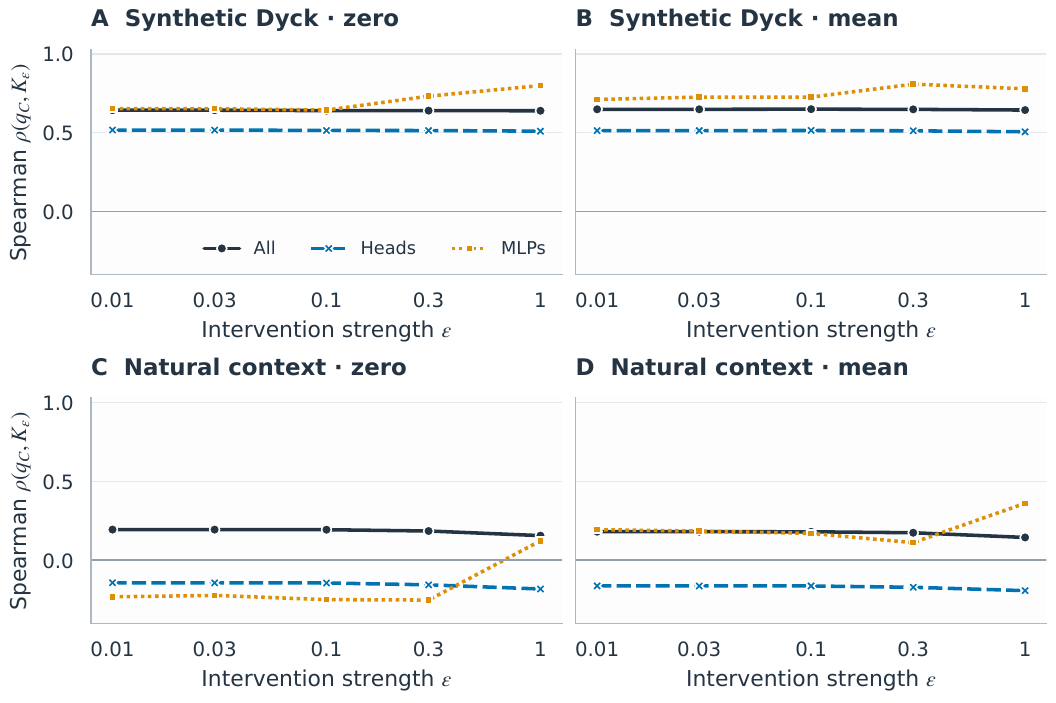}
\caption{\textbf{Local and full-strength interventions probe different directions from the posterior average.} Each curve compares the same fixed LLC estimates with reference-KL effects along zero- or mean-replacement directions. All components and both component classes are included at every strength. Dividing KL by the fixed $\epsilon^2$ leaves the displayed rank correlation unchanged.}
\label{fig:intervention-scale}
\end{figure}

The measured scaling is quadratic: between $0.01$ and $0.03$, the median fitted power is $2.000$--$2.001$ across the four distribution/replacement combinations, and median changes in $K_\epsilon/\epsilon^2$ are $0.09$--$0.21\%$. Figure~\ref{fig:intervention-scale} shows that synthetic rank correspondence persists across strengths, while natural-context head associations remain negative. The latter difference is already present in the local directional response.

%% file: sections/semantic_projection_appendix.tex
\section{Posterior perturbations in semantic circuit subspaces}
\label{app:semantic-projection}
We measure the contribution of circuit subspaces to parameter-induced output changes using a fixed subset of RMS-posterior states. For \C{L11.mlp} and \C{L14H1} on each distribution, we select retained indices 128 and 384 from each of four chains, giving eight states per condition. Inputs are identical to those used for posterior sampling. Each state is reconstructed from its random-number stream and acceptance decisions; the maximum difference between replayed and stored full KL is $5.32\times10^{-9}$.

Let $o_C(x;u)$ be the output of component $C$ under a sampled parameter state $u$, and define
\begin{equation}
 \Delta o_C(x;u)=o_C(x;u)-o_C(x;0).
\end{equation}
After restoring the checkpoint parameters, we patch the clean output with $\Delta o_C$, its projection $P_C\Delta o_C$ onto the exact named span, its complement $(I-P_C)\Delta o_C$, its ANSWER projection, or projections onto two fixed random spans matching the named rank. Interventions act at all positions in each evaluated sequence. The named spans have rank 166 for \C{L11.mlp} and 94 for \C{L14H1}; ANSWER has rank 3. The random seeds are 8201 and 8202.

\begin{figure}[t]
\centering
\includegraphics[width=\linewidth]{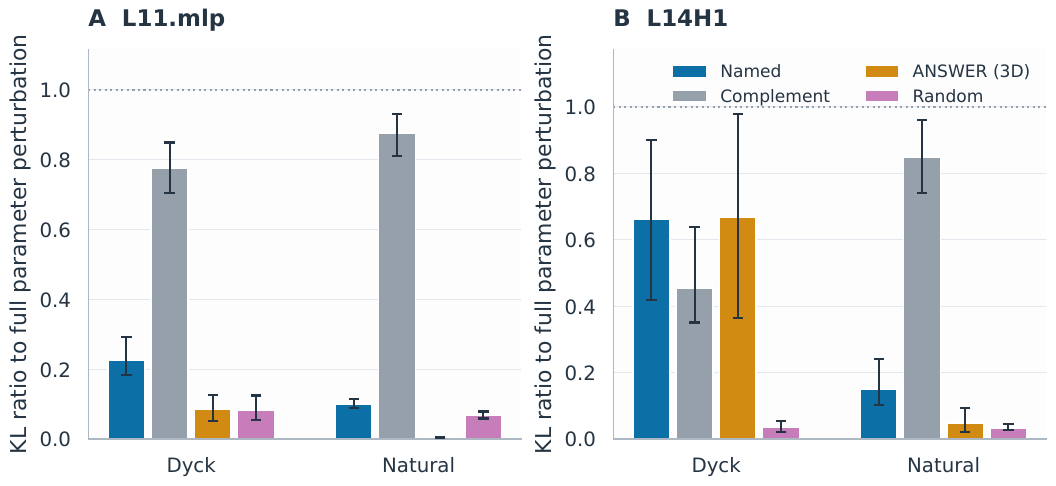}
\caption{\textbf{Projecting posterior-induced changes onto circuit variables.} Bars pool KL over eight fixed states per cell. Whiskers show the range of per-state ratios; random bars average the two fixed subspaces, with whiskers spanning both. Projection and complement losses have signed cross effects, so their ratios are nonadditive and may exceed one. These states sample one fixed evaluation batch per distribution.}
\label{fig:semantic-projection}
\end{figure}

The full activation patch verifies the connection between parameter and activation interventions. It reproduces every scored log probability exactly for the MLP, and within $2.14\times10^{-5}$ for the head. The latter difference comes from the equivalent per-head and summed-output floating-point computations (Figure~\ref{fig:semantic-projection}).

\input{sections/semantic_projection_table}

For each projection, Table~\ref{tab:semantic-projection} reports the sum of patch KL across the eight states divided by the sum of full parameter-perturbation KL. The strongest concentration occurs in the synthetic \C{L14H1} readout: the three-dimensional ANSWER space alone yields ratio $0.668$, compared with $0.032$--$0.037$ for random 94-dimensional spans. The broad named span also exceeds both random controls for the MLP and on natural-context inputs. These comparisons link the localized response to specific represented variables, while retaining the different degrees of concentration across components and distributions.

We retain the signed interaction
\begin{equation}
 K(\Delta o_C)-K(P_C\Delta o_C)-K((I-P_C)\Delta o_C)
\end{equation}
for every state. It captures cross terms and higher-order effects of the downstream computation.

%% file: sections/semantic_projection_table.tex
\begin{table}[t]
\centering\small
\begin{tabular}{llrrrr}
\toprule
Component & Inputs & Named & Complement & ANSWER & Random range \\
\midrule
L11.mlp & synthetic & 0.227 & 0.775 & 0.086 & 0.082--0.084 \\
L11.mlp & natural & 0.101 & 0.877 & 0.004 & 0.066--0.070 \\
L14H1 & synthetic & 0.662 & 0.454 & 0.668 & 0.032--0.037 \\
L14H1 & natural & 0.150 & 0.848 & 0.049 & 0.031--0.033 \\
\bottomrule
\end{tabular}
\caption{Pooled reference-KL ratios over eight fixed posterior states per cell. Named spans have rank 166 for L11.mlp and 94 for L14H1; each random control matches that named rank. ANSWER has rank 3. The numerator sums projected-patch KL across states and the denominator sums full parameter-perturbation KL.}
\label{tab:semantic-projection}
\end{table}

%% file: sections/sensitivity_appendix.tex
\section{Sampler sensitivity and joint circuit geometry}
\label{app:sensitivity}

\subsection{A fixed sensitivity panel}
The sensitivity analysis includes 18 components: all eight first-layer heads, \C{L0.mlp}, \C{L11.mlp}, \C{L14H1}, \C{L7H4}, \C{L9H2}, \C{L15.mlp}, and four randomly selected comparators (\C{L11H0}, \C{L5H7}, \C{L2H4}, and \C{L15H1}). The components and comparator seed are fixed before measurement. The 288 conditions comprise checkpoints 8,000 and 32,000, four factor-two radius or inverse-temperature changes, a proposal-coefficient control, and a closer-conditional-KL target, each evaluated on both distributions.

Radius and inverse temperature define the posterior; the proposal coefficient controls the transition kernel. The closer-conditional target normalizes over the four closer tokens with the observed surface form. All conditions use the four-chain diagnostic criteria of the posterior experiment. Runs that fail at 512 draws are extended with the same seeds, preserving the initial acceptance sequence and metric trace. Of the 288 conditions, 246 meet the criteria at 512 retained draws, 37 at 1,024, and three at 2,048. The two wide-radius \C{L15.mlp} conditions use a proposal coefficient selected by the minimum effective samples per transition across distributions and meet the criteria with 4,096 retained draws.

Radius and inverse-temperature controls preserve component ranks ($\rho=0.975$--$1.000$); their median response ratios differ from the weak-localization prediction $q\propto\tau r^2$ by at most $3.6\%$. Changing proposal coefficient from $0.9$ to $0.5$ yields rank correlations $0.998$ and $0.996$ on synthetic and natural-context inputs, respectively. All 36 paired differences lie within three conservative MCSE bounds, computed by summing the two marginal MCSEs to allow shared-seed correlation. The closer-conditional target yields rank correlations $0.926$ and $0.810$, with median response ratios $0.683$ and $5.91\times10^{-4}$. Restricting the prediction to relative bracket identity therefore changes the measured geometry substantially on natural-context inputs.

\subsection{Joint geometry and single components}
\label{app:jointgeometry}
The bank restriction varies all eight first-layer heads together using the exact product of their atomic RMS priors. Both its parameter values and each member's scales are matched to the atomic runs. Write $u=(u_1,\ldots,u_8)$, $K_h(u_h)=K(0,\ldots,u_h,\ldots,0)$, and
\begin{equation}
 Z_{\rm bank}(\tau)=\int e^{-\tau K(u)}\prod_h\varphi_h(u_h)\,du,
 \qquad q_{\rm bank}=-\tau\partial_\tau\log Z_{\rm bank}.
\end{equation}
If $K(u)=\sum_hK_h(u_h)$, the partition function factorizes, yielding
\begin{equation}
 q_{\rm bank}=\sum_hq_h.
 \label{eq:joint-separability}
\end{equation}
A joint-versus-atomic difference consequently measures a departure from this separable local-loss model.

\begin{figure}[t]
\centering
\includegraphics[width=\linewidth]{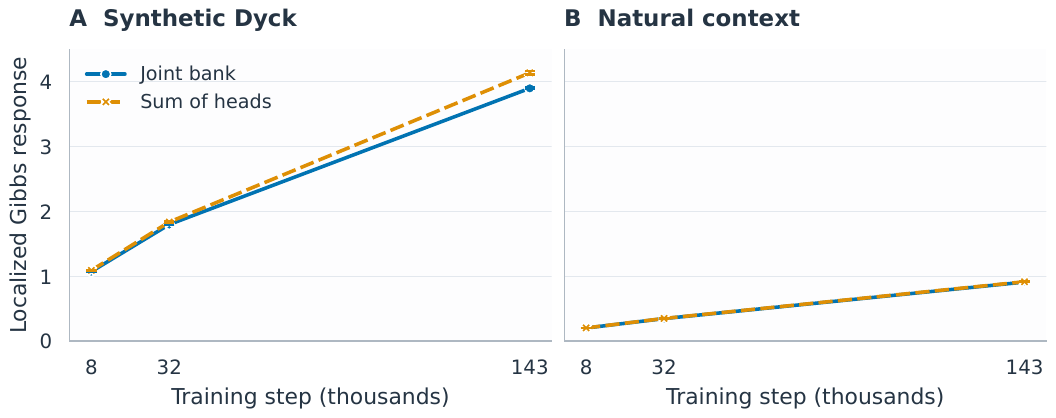}
\caption{Joint and summed single-head LLC estimates on the same frozen inputs and product prior. Error bars show individual bank MCSEs and conservative summed-head MCSE bounds. The matched reference-KL response increases across the three checkpoints on both distributions.}
\label{fig:bank-gibbs-development}
\end{figure}

\begin{figure}[t]
\centering
\includegraphics[width=\linewidth]{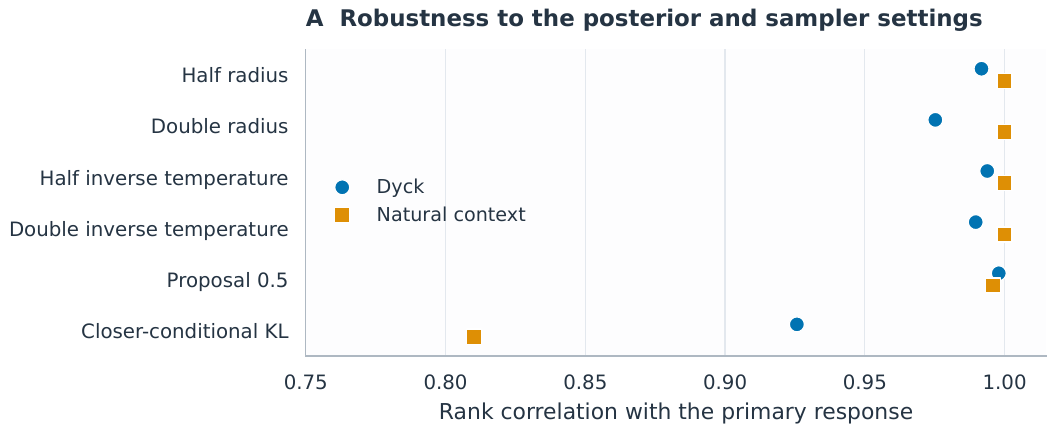}
\caption{Sensitivity-panel rank correlations with the primary LLC estimate. The first four controls alter the posterior; the proposal control preserves it. The closer-conditional objective changes the readout. Every rank comparison includes the complete fixed 18-component panel.}
\label{fig:gibbs-sensitivity}
\end{figure}

For the quadratic reference loss, Equation~\eqref{eq:joint-geometry} strengthens this comparison to $q_{\rm bank}\leq\sum_hq_h$, with strict inequality for cross-head curvature. This supplies a directional prediction for coupled task-sensitive parameter directions under the matched product prior (Figure~\ref{fig:bank-gibbs-development}) (Figure~\ref{fig:gibbs-sensitivity}).

\subsection{Monte Carlo uncertainty with shared seeds}
The head runs and bank run reuse the same chain-seed convention. We therefore propagate uncertainty conservatively: for $S=\sum_hq_h$, we report $s_S=\sum_h s_h$, and for $\Delta=q_{\rm bank}-S$ we use $s_\Delta=s_{\rm bank}+s_S$. These upper bounds on the propagated standard error allow arbitrary cross-run correlation. The ratio uses the first-order bound $s_{q/S}=s_{\rm bank}/S+q_{\rm bank}s_S/S^2$. Ratios and signed differences are both retained.

\input{reproducibility/sensitivity/derived/sensitivity_tables}

%% file: reproducibility/sensitivity/derived/sensitivity_tables.tex
\begin{table}[t]
\centering\small
\begin{tabular}{llrrrr}
\toprule
Distribution & Step & Bank $q$ & Head sum & Ratio & Gap/$s_\Delta$ \\
\midrule
natural & 8,000 & $0.201\pm0.001$ & $0.202\pm0.001$ & 0.995 & -0.81 \\
natural & 32,000 & $0.346\pm0.001$ & $0.349\pm0.001$ & 0.993 & -1.10 \\
natural & 143,000 & $0.910\pm0.003$ & $0.915\pm0.004$ & 0.994 & -0.75 \\
synthetic & 8,000 & $1.072\pm0.006$ & $1.091\pm0.006$ & 0.983 & -1.65 \\
synthetic & 32,000 & $1.798\pm0.009$ & $1.837\pm0.013$ & 0.979 & -1.81 \\
synthetic & 143,000 & $3.899\pm0.016$ & $4.137\pm0.026$ & 0.942 & -5.68 \\
\bottomrule
\end{tabular}
\caption{Matched joint-bank and atomic-head measurements. The head-sum and difference uncertainties use sums of marginal MCSEs, allowing arbitrary correlation from shared seeds. The ratio is joint response divided by the atomic sum.}
\label{tab:bank-gibbs}
\end{table}
\begin{table}[t]
\centering\small
\begin{tabular}{lrrrr}
\toprule
& \multicolumn{2}{c}{Synthetic} & \multicolumn{2}{c}{Natural context} \\
Control & Rank $\rho$ & Median $q/q_0$ & Rank $\rho$ & Median $q/q_0$ \\
\midrule
Half radius & 0.992 & 0.253 & 1.000 & 0.25 \\
Double radius & 0.975 & 3.86 & 1.000 & 4.03 \\
Half inverse temperature & 0.994 & 0.507 & 1.000 & 0.502 \\
Double inverse temperature & 0.990 & 1.95 & 1.000 & 1.99 \\
Proposal 0.5 & 0.998 & 0.996 & 0.996 & 1 \\
Closer-conditional KL & 0.926 & 0.683 & 0.810 & 0.000591 \\
\bottomrule
\end{tabular}
\caption{Qualified sensitivity comparisons on all 18 prespecified components. $q_0$ is the primary $r=0.1$, $\tau=1000$ full-KL response at checkpoint 143,000. Radius and inverse temperature alter the posterior, proposal coefficient preserves it, and closer KL changes the readout.}
\label{tab:gibbs-sensitivity}
\end{table}

%% file: sections/development_detail.tex
\subsection{Detailed developmental and group comparisons}
\label{app:development-detail}

\begin{figure}[t]
\centering
\includegraphics[width=\linewidth]{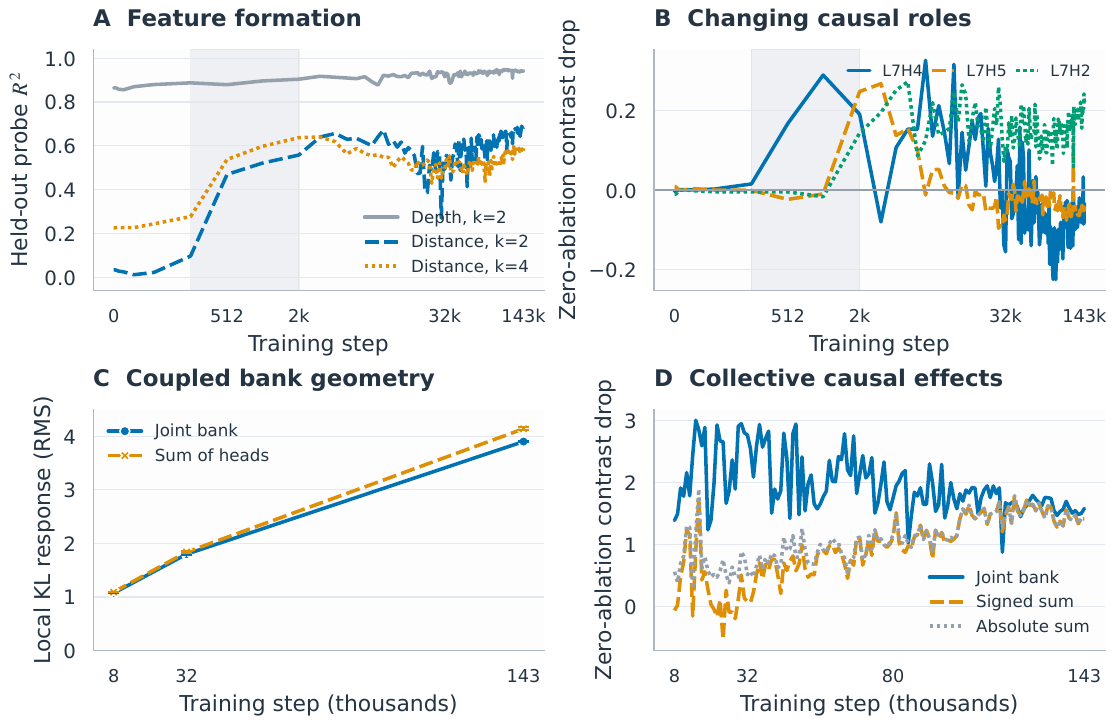}
\caption{\textbf{Circuit development and collective organization.} A: layer-2 state-feature probes. B: causal effects of three layer-7 heads on $k=2$. C: local RMS-posterior responses of the first-layer bank and the sum of its heads at three checkpoints; whiskers show Monte Carlo standard errors, with conservative bounds for the sum. D: joint and marginal bank ablations on $k=4$. A and B include all 154 checkpoints; D shows training from step 8,000 onward. A, B, and D use the developmental batches; C uses the fixed posterior calibration batch.}
\label{fig:development-late}
\end{figure}

\begin{figure}[t]
\centering
\includegraphics[width=\linewidth]{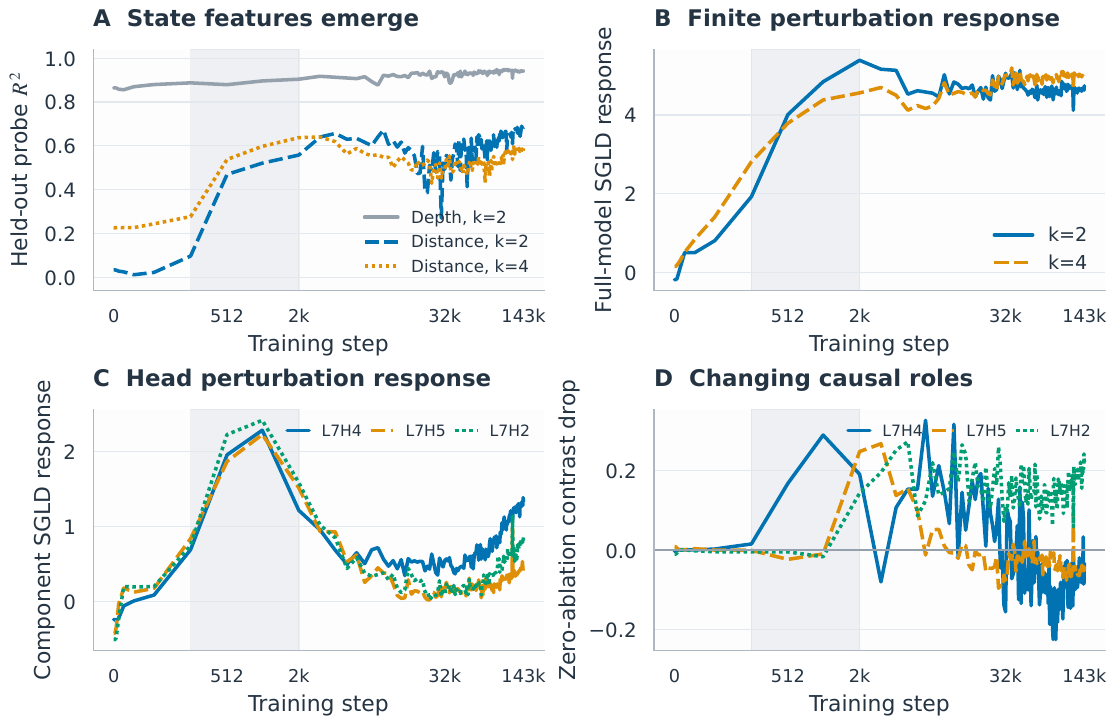}
\caption{\textbf{Feature formation and changing component roles across all 154 checkpoints.} A: held-out layer-2 prefix-state probes. B: full-model finite SGLD response. C--D: the same three layer-7 heads on $k=2$, measured by finite SGLD and signed zero-ablation. The horizontal axis is logarithmic above step 128; shading marks steps 256--2,000. The finite SGLD response uses the perturbation regime characterized in Appendix~\ref{app:locality}.}
\label{fig:development-overview}
\end{figure}
The developmental sweep evaluates 49,588 component-, layer-, and model-restricted SGLD responses across both synthetic task variants and all 154 checkpoints. These measurements perturb the weights at the larger scale characterized in Appendix~\ref{app:locality}. The statistic averages the full finite sampling trajectory; Appendix~\ref{app:methods} gives the protocol, and Appendix~\ref{app:trace-reanalysis} compares averaging windows. Across the 12 checkpoints measured on both input distributions, finite SGLD response component rankings have median Spearman correlation $0.89$ between synthetic and natural-context data, compared with $0.35$ for signed zero-ablation effects and $0.37$ for their magnitudes. The difference is strongest among attention heads: the corresponding finite SGLD response and signed-ablation correlations are $0.88$ and $0.27$, compared with $0.87$ and $0.81$ among MLPs. These paired observations establish substantial stability of the perturbation-response ranking across the two bracket distributions.

We also compare the $k=4$ layer-2 prefix-state distance trajectory with zero-ablation effects and gradient activity (Table~\ref{tab:development-baselines}): the finite SGLD response's association ($0.75$, and $0.60$ after removing the linear checkpoint-rank trend) persists where the ablation and gradient baselines' associations ($0.15$ and $0.27$) fall to zero after the same adjustment.

This correspondence is stable under changes to the averaging window. On matched observations, all-step and retained-only component rankings have median correlation $0.994$; their rank-time-adjusted feature associations are $0.589$ and $0.615$. Excluding the initial sampling updates therefore preserves the developmental association (Appendix~\ref{app:trace-reanalysis}).

A componentwise intervention comparison provides a further connection to the represented variable. We zero-ablate each of 27 early components and refit the distance probe to measure the remaining recoverable information. Across 12 checkpoints, median componentwise correlations with the resulting loss of probe $R^2$ are $0.25$ for finite SGLD response and $0.28$ for contrast-based ablation; at the final checkpoint they are $0.59$ and $0.45$. Both measurements therefore carry information about the components supporting this representation, with particularly clear correspondence at the trained checkpoint.

\paragraph{Probe trajectories.}
On the $k=2$ strings the held-out $R^2$ of the layer-2 prefix-depth probe is $0.864$ at initialization, whereas that of the layer-2 opener-distance probe rises from $0.034$ at initialization to $0.558$ at step 2,000 and $0.684$ at the final checkpoint (on $k=4$ strings, $0.226$, $0.637$, and $0.589$); Figure~\ref{fig:development-overview}A shows the full trajectories.

\subsection{A distributed attention bank relates parameter responses to group effects}
\label{app:bank-detail}
\begin{table}[t]
\centering\footnotesize
\begin{tabular}{@{}lccc@{}}
\toprule
Target of $\sum_hA_h$ (eight layer-0 heads) & Spearman & partial & changes [95\%] \\
\midrule
Sum of the eight single-head contrast drops & $0.67$ & $0.31$ & $0.22$ [$0.06$, $0.40$] \\
Sum of their absolute values & $0.61$ & $0.16$ & $0.25$ [$0.08$, $0.37$] \\
Sum of the eight single-head zero-ablation KLs & $0.54$ & $0.16$ & $0.71$ [$0.63$, $0.79$] \\
Joint zero-ablation contrast drop of the group & $-0.38$ & $-0.36$ & $0.03$ [$-0.15$, $0.20$] \\
\bottomrule
\end{tabular}
\caption{\textbf{Summed single-head geometry against single and joint ablations (P3).} Associations across the 148 confirmation checkpoints between the sum of the eight layer-0 heads' Fisher traces and each target; partial rank correlations given clean loss, weight scale, and training step, adjacent-change partials given the same, and moving-block bootstrap intervals (blocks of 16). The summed trace reaches its minimum at step 6,000 and the joint ablation effect its maximum at step 15,000, which produces the negative level association.}
\label{tab:bank-associations}
\end{table}
All eight first-layer heads rank among the top 23 of 144 components by synthetic finite SGLD response at the final checkpoint. Joint interventions establish the causal contribution of this group. In the cross-distribution intervention batch, zeroing the bank lowers closer contrast by $1.78$ on synthetic data and $2.28$ on natural-context data. Position-conditioned mean replacement gives drops of $0.14$ and $1.69$, respectively. Relative to clean contrasts of $3.56$ and $13.16$, the mean-replacement drops are $3.9\%$ and $12.8\%$. This comparison uses 11,882 scored synthetic positions and 187 natural-context positions.

The contrast between replacement rules reveals how the bank contributes to the task. A position-conditioned mean retains the average write at each position; zero replacement removes the entire write. Their different effects show that much of the bank's contribution on synthetic inputs survives position-wise averaging, whereas natural-context completion depends more strongly on its input-specific output. The perturbation-response ranking identifies the bank as a candidate group, and the intervention battery resolves how that group supports the computation.

\begin{table}[t]
\centering
\begin{tabular}{rrrr}
\toprule
Training step & Joint zero drop & Signed marginal sum & Absolute marginal sum \\
\midrule
1,000 & $-0.153$ & $-0.274$ & 0.387 \\
8,000 & 1.378 & $-0.079$ & 0.572 \\
27,000 & 2.378 & $-0.064$ & 0.614 \\
32,000 & 2.797 & 0.078 & 0.462 \\
64,000 & 2.794 & 0.695 & 0.781 \\
143,000 & 1.560 & 1.332 & 1.332 \\
\bottomrule
\end{tabular}
\caption{First-layer bank effects under the fixed longitudinal protocol. From step 8,000 on, the joint drop exceeds even the absolute marginal sum, establishing nonadditivity beyond signed cancellation.}
\label{tab:bank-detail}
\end{table}

\subsection{Late perturbation-response changes accompany computational reorganization}
\label{app:late-detail}
In the first-layer bank, the late change of the nonlocal SGLD response accompanies a marked shift in the relationship between joint and marginal causal effects. At step 32,000, joint zero-ablation lowers contrast by $2.80$, while the signed sum of the eight individual drops is $0.078$. Their absolute sum is also small, $0.462$, establishing substantial nonadditivity beyond cancellation of positive and negative effects. At the final checkpoint, the joint and marginal effects are much closer: $1.56$ and $1.33$, respectively.

\begin{figure}[t]
\centering
\includegraphics[width=\linewidth]{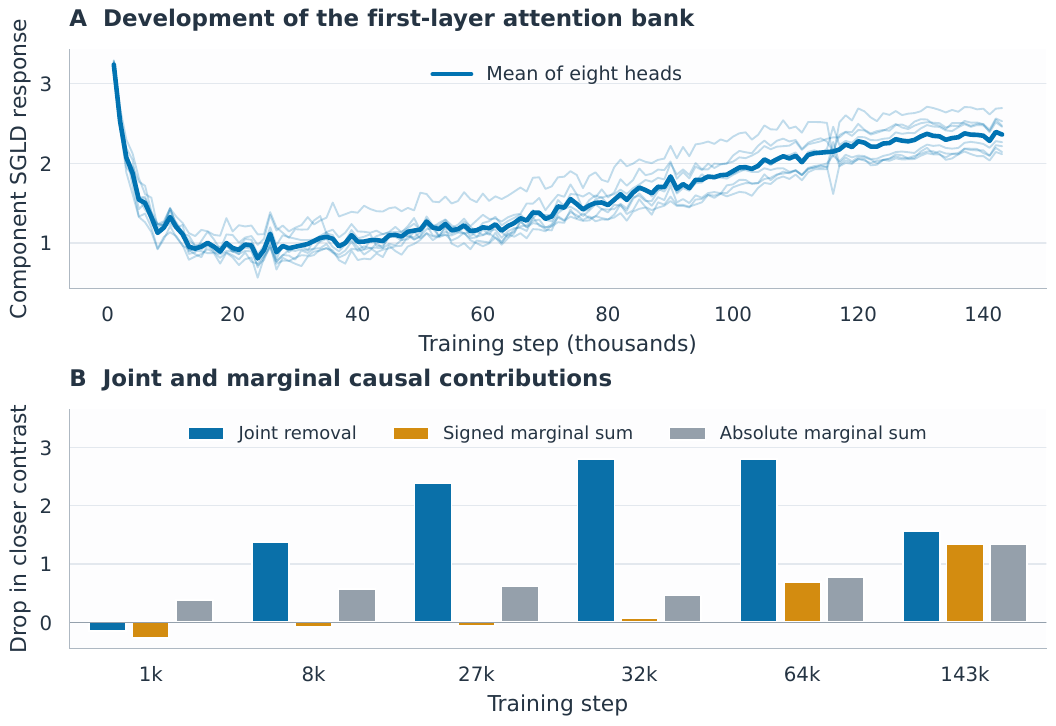}
\caption{\textbf{Late perturbation-response change accompanies changing causal organization.} The first-layer bank's finite SGLD response trajectory is compared with joint and marginal zero-ablation under a fixed synthetic protocol at six checkpoints. Signed and absolute marginal sums are both small near the response trough and approach the joint effect late in training.}
\label{fig:bank}
\end{figure}

\begin{figure}[t]
\centering
\includegraphics[width=\linewidth]{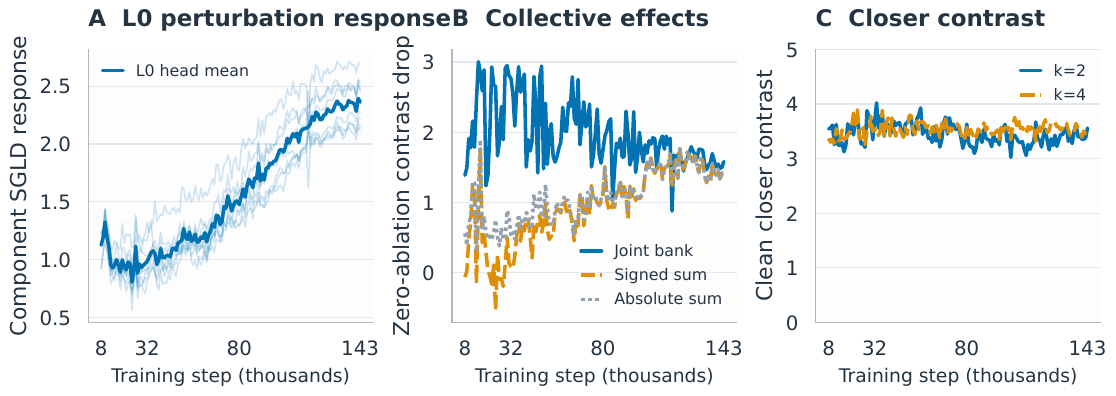}
\caption{\textbf{Finite perturbation responses and late circuit development.} A: the eight first-layer finite SGLD trajectories on $k=4$. B: joint bank ablation and signed and absolute sums of individual effects. C: clean closer contrast. All checkpoints from step 8,000 onward are included; the SGLD regime is characterized in Appendix~\ref{app:locality}.}
\label{fig:finite-development-late}
\end{figure}

The 128-string evaluation across all checkpoints shows the same change in joint and marginal contributions (Figure~\ref{fig:development-late}). Clean contrast is $3.55$ at step 32,000 and $3.54$ at step 143,000. Full-vocabulary accuracy fluctuates across the intervening checkpoints. The joint interventions distinguish changes in the allocation of causal contributions despite the similar endpoint contrasts (Figure~\ref{fig:finite-development-late}).

%% file: sections/algorithm_appendix.tex
\section{Algorithmic structure and causal verification}
\label{app:algorithm-detail}
\input{sections/algorithm}

\subsection{A distributed realization of bracket matching}
Let $\sigma_t$ denote the stack before token $x_t$ is consumed. The reference computation pushes the type of an opener, pops the last unmatched type on a closer, and predicts $\operatorname{close}(\operatorname{top}(\sigma_{t+1}))$ whenever the next token is a closer. This formulation separates the represented prefix state from the output decision. In Pythia, the prefix state is distributed across attention and MLP outputs, and several later components write the corresponding closer evidence into the residual stream.

Writing $h_{\ell,t}$ for the residual stream entering layer $\ell$, a parallel Pythia block has the form
\begin{equation}
 h_{\ell+1,t}=h_{\ell,t}
 +\sum_n a_{\ell n,t}(h_{\ell,\leq t})
 +m_{\ell,t}(h_{\ell,t}).
\end{equation}
At layer zero, the MLP input depends only on the token embedding, so $m_{0,t}=v_{x_t}$ is exactly a token lookup. First-layer attention acts in parallel, supplying prefix information. The early stack-writing group consists of \C{L1--L5.mlp}, \C{L1H1}, \C{L2H1}, and \C{L2H5}. The selected attention readouts are
\begin{equation*}
\begin{split}
\mathcal H_{\rm read}={}&\{\C{L7H0},\C{L7H2},\C{L7H4},\C{L7H6},
\C{L8H1},\C{L8H4},\\
&\C{L9H0},\C{L9H4},\C{L9H5},\C{L13H0},\C{L14H1}\}.
\end{split}
\end{equation*}
Together with \C{L11.mlp}, these heads provide positive closer evidence at several depths. \C{L13--L15.mlp} contribute negative answer-direction effects on average. Their signed contributions explain why removing a component can improve the task score even when that component participates in the same computation.

\subsection{Subspace interventions}
For an orthonormal basis $U_C$, projection replaces a component output $o_C$ by $U_CU_C^\top o_C$; removal replaces it by $(I-U_CU_C^\top)o_C$. The broad named space spans TOKEN, STACK, POSITION, and ANSWER directions. TOKEN is obtained from first-layer token-conditioned MLP outputs; STACK from additive slot/type fits; POSITION from position-conditioned residual means; and ANSWER from closer unembedding directions. The broad STACK span combines the fitted spaces at several neighboring layers, allowing the representation to rotate as computation progresses.

All named spans are constructed by rank-revealing SVD with relative tolerance $10^{-6}$. The six-slot, four-type additive stack code has rank 24; the centered closer-answer space has rank 3. TOKEN has rank 11 and POSITION retains 32 principal directions. Their narrow union has rank 38, while the broad union spans 70--190 directions depending on the number of subsequent layers included. Augmentation adds 64 principal directions of each component's residual output covariance, fitted on the same 256 seed-42 strings.

\begin{figure}[t]
\centering
\includegraphics[width=\linewidth]{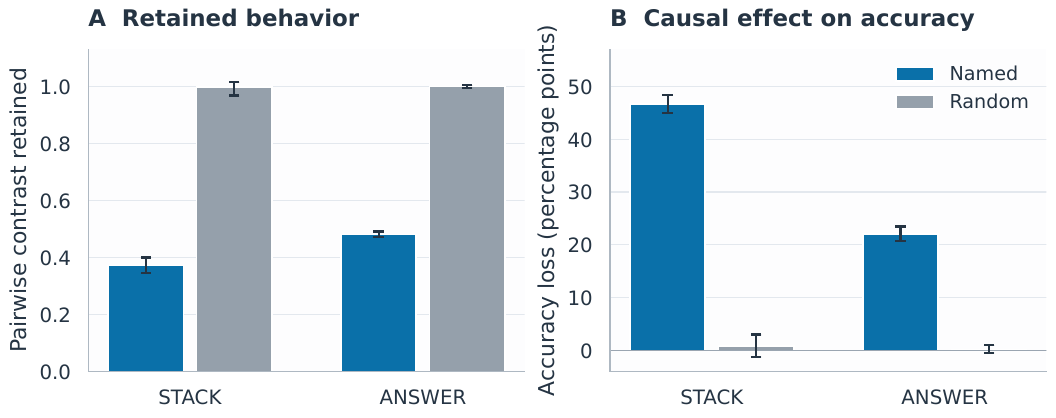}
\caption{\textbf{Effects of stack- and answer-subspace removal.} Interventions are evaluated on 384 held-out strings. Left: retained pairwise closer contrast. Right: loss of full-vocabulary closer accuracy. Error bars for the identified subspaces are $95\%$ paired string-bootstrap intervals; random-control bars summarize five fixed rank-matched bases and their range.}
\label{fig:algorithm-validation}
\end{figure}

The cumulative replacement acts on 10 MLPs and 14 attention heads, with the eight first-layer heads additionally replaced by position-conditioned means. All other components execute normally. The held-out recovery in Table~\ref{tab:coverage} measures the behavior retained by this representational bottleneck in the complete network. The named variables retain $76.3\%$ of the closer contrast; adding 64 orthogonal principal directions increases this to $91.9\%$ (Figure~\ref{fig:algorithm-validation}).

\input{sections/algorithm_holdout_results}

\subsection{Geometry of the represented stack variables}
\label{app:stack-geometry}
The stack directions expose structure below the component level. For each layer, an additive fit represents the stack-dependent part of the residual as
\begin{equation}
 \widehat h_{\ell,t}=b_\ell+
 \sum_{i=1}^{d_t}\bigl(o_{\ell i}+e_{\ell i,\sigma_t[i]}\bigr),
 \qquad \sum_{a=1}^{4}e_{\ell i,a}=0.
\end{equation}
Here $o_{\ell i}$ denotes slot occupancy and $e_{\ell i,a}$ the type offset at that slot. This separates how much of the stack is occupied from which bracket type occupies it.

\begin{figure}[t]
\centering
\includegraphics[width=\linewidth]{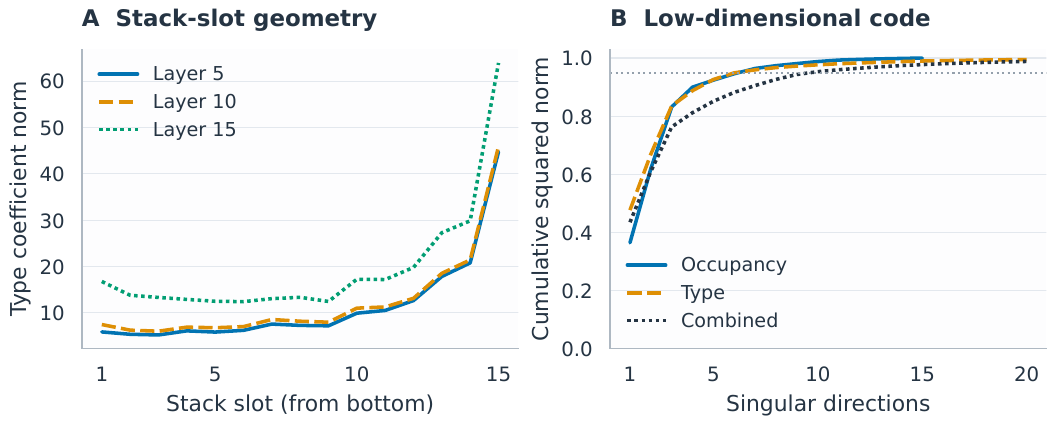}
\caption{\textbf{Structure within the fitted stack code.} Left: the norm of centered type coefficients by slot in a separate fully nested dataset. Right: cumulative squared singular values of the layer-10 type and occupancy coefficient matrices. These spectra describe the stack-dependent additive fit; the fits account for $8.1\%$, $8.8\%$, and $13.1\%$ of total residual variance at layers 5, 10, and 15, respectively. Slots are counted from the bottom.}
\label{fig:stack-geometry}
\end{figure}

A separate fully nested dataset covers depths 1--15. At layer 10, the top four singular directions capture $90.0\%$ of the squared norm of the fitted occupancy coefficients. The four type offsets at each slot form a centered tetrahedral configuration. After orthogonal Procrustes alignment, every slot has similarity at least $0.948$ to the first slot; the type-offset Frobenius norm at slot 15 is $6.15$ times that at slot 1. Thus the fitted type configurations retain a similar shape across slots while their scale changes substantially. The combined occupancy/type coefficient matrix requires 10 singular directions to capture $95\%$ of its squared norm (Figure~\ref{fig:stack-geometry}).

The geometry identifies reusable variables for interventions. The held-out projection and removal experiments then establish how those variables support behavior: compact stack-dependent directions can matter substantially for bracket prediction even when most residual variance carries other information.

\subsection{Coverage of the mechanistic reference}
\input{sections/algorithm_coverage_table}
Adding position and multilayer stack directions increases held-out recovery from $0.449$ to $0.763$; the removal experiments above test the causal contribution of these variables.

\input{sections/probe_appendix}

%% file: sections/algorithm.tex
\begin{figure}[t]
\centering
\scriptsize
\setlength{\tabcolsep}{4pt}
\renewcommand{\arraystretch}{0.86}
\begin{tabularx}{\linewidth}{@{}>{\raggedright\arraybackslash}p{.31\linewidth}>{\raggedright\arraybackslash}X@{}}
\toprule
\textbf{Symbolic bracket matching} & \textbf{Transformer components and the operations they implement} \\
\midrule
\textbf{1. Encode the current token.}\newline
Read its bracket type $\kappa(x_t)$ and whether it opens or closes.
&
\textbf{\C{L0.mlp} and \C{L0H0--H7}.}\newline
$m_{0,t}=v_{x_t}$ is a token lookup, and the eight parallel heads add $\sum_n a_{0n,t}$, mixing token features across the prefix; replacing the MLP by its lookup preserves the contrast. \\
\addlinespace[2pt]
\textbf{2. Update the stack.}\newline
$\sigma_{t+1}=\operatorname{push}(\sigma_t,\kappa(x_t))$ for an opener;\newline
$\sigma_{t+1}=\operatorname{pop}(\sigma_t)$ for a closer.\newline
Depth is $d_t=|\sigma_t|$.
&
\textbf{\C{L1--L5.mlp}, \C{L1H1}, \C{L2H1}, \C{L2H5}.}\newline
These components write into a fitted STACK subspace with additive slot and type terms,
$\widehat h^{\rm stack}_{\ell,t}=b_\ell+\sum_{i\leq d_t}(o_{\ell i}+e_{\ell i,\sigma_t[i]})$; removing these writes reduces held-out closer accuracy from $61.4\%$ to $14.8\%$. \\
\addlinespace[2pt]
\textbf{3. Read the required type.}\newline
Let $j^*$ be the most recent unmatched opener. Then\newline
$\kappa^*=\kappa(x_{j^*})=\operatorname{top}(\sigma_{t+1})$.
&
\textbf{Readout heads, including \C{L7H2}, \C{L7H4}, \C{L7H5}, \C{L9H4}, and \C{L14H1}.}\newline
$a_{\ell n,t}=\sum_{j\leq t}\alpha_{\ell n,tj}\,\mathrm{OV}_{\ell n}(h_{\ell,j})$, with $h_{\ell,j}$ the normalized residual stream and $\mathrm{OV}$ the value--output map; attention to opener positions moves type information into the ANSWER subspace. \\
\addlinespace[2pt]
\textbf{4. Map the type to a closer.}\newline
$y^*=\operatorname{close}(\kappa^*)$.
&
\textbf{\C{L11.mlp} and distributed answer-writing heads.}\newline
The MLP writes positive closer evidence already at layer~11; later heads add more. Removing its ANSWER projection costs $8.4$ accuracy points. \\
\addlinespace[2pt]
\textbf{5. Emit the next-token prediction.}\newline
The stack rule specifies the closer that is legal after the observed prefix.
&
\textbf{Final readout and \C{L13--L15.mlp}.}\newline
$p(\cdot\mid x_{\le t})=\operatorname{softmax}(W_U\operatorname{LN}_{f}(h_{16,t})+b_U)$; the last three MLPs write opposing evidence, and removing any one's ANSWER projection raises accuracy by $2.5$--$2.8$ points. \\
\bottomrule
\end{tabularx}
\caption{\textbf{A symbolic algorithm beside its transformer realization.} Each row pairs a step of the stack algorithm ($\sigma_t$ is the stack before token $x_t$) with the components that implement it and the causal test behind the assignment (Appendix~\ref{app:algorithm-detail}).}
\label{fig:algorithm-mapping}
\end{figure}

%% file: sections/algorithm_holdout_results.tex
\paragraph{Held-out causal validation.}
Rank-revealing SVD constructs the 24-dimensional STACK and three-dimensional ANSWER spaces from the frozen seed-42 fits, using a relative singular-value tolerance of $10^{-6}$. We evaluate all interventions on the same 384 strings (128 each from seeds 43--45). The model, strings, distractors, and scored positions are identical across conditions. The early group removes STACK from five MLPs and three heads; the answer-writing group removes ANSWER from 11 heads and four MLPs. For each joint intervention, five independently sampled random subspaces match the saved basis rank at every intervened layer. All measurements use fp32 at step 143,000.

\begin{table}[t]
\centering\small
\begin{tabular}{lrr}
\toprule
Removed directions & Contrast recovery [95\% CI] & Accuracy change (pp) \\
\midrule
Early group: STACK & $0.372\;[0.345,0.400]$ & $-46.6$ \\
Answer writers: ANSWER & $0.481\;[0.471,0.491]$ & $-22.1$ \\
L1.mlp: STACK & $0.849\;[0.844,0.855]$ & $-2.5$ \\
L2.mlp: STACK & $0.817\;[0.812,0.822]$ & $+0.4$ \\
L3.mlp: STACK & $0.773\;[0.759,0.786]$ & $-18.7$ \\
L11.mlp: ANSWER & $0.810\;[0.802,0.818]$ & $-8.4$ \\
L14H1: ANSWER & $0.843\;[0.840,0.847]$ & $-8.2$ \\
L13.mlp: ANSWER & $1.086\;[1.081,1.092]$ & $+2.5$ \\
L14.mlp: ANSWER & $1.056\;[1.047,1.065]$ & $+2.6$ \\
L15.mlp: ANSWER & $1.062\;[1.053,1.071]$ & $+2.8$ \\
Random: early group & $0.968$--$1.015$ & $-3.0$ to $+1.2$ \\
Random: answer group & $0.995$--$1.004$ & $-1.0$ to $+0.4$ \\
\bottomrule
\end{tabular}
\caption{Held-out removal of algorithmic subspaces. Recovery is the intervened pairwise contrast divided by the clean contrast. Confidence intervals use 2,000 paired string bootstraps stratified by evaluation seed. Accuracy changes are for full-vocabulary argmax at closer targets, in percentage points; negative values indicate a decrease. Random-control rows give the range over five frozen random bases, separately from sampling uncertainty.}
\label{tab:algorithm-holdout}
\end{table}

The matched controls establish that the named directions carry task-specific information (Table~\ref{tab:algorithm-holdout}). Joint STACK removal reduces closer accuracy from $61.4\%$ to $14.8\%$, whereas random removal retains $58.4$--$62.6\%$. ANSWER removal reduces accuracy to $39.3\%$, compared with $60.4$--$61.8\%$ for its random controls. The agreement between contrast and full-vocabulary accuracy connects the recovered stack representation to the model's output decisions.

%% file: sections/algorithm_coverage_table.tex
\begin{table}[h]
\centering
\begin{tabular}{lrr}
\toprule
Replacement & Seed 42 & Held-out seeds 43--45 \\
\midrule
Narrow named subspaces & 0.439 & $0.449\pm0.013$ \\
Broad named subspaces & 0.762 & $0.763\pm0.011$ \\
Broad named + 64 orthogonal PCs & 0.920 & $0.919\pm0.001$ \\
PCA (128/64) & 0.901 & $0.903\pm0.002$ \\
Linear surrogate & 0.113 & $0.110\pm0.007$ \\
\bottomrule
\end{tabular}
\caption{Clean pairwise contrast retained under cumulative replacement. Values are means and sample standard deviations over three held-out batches of 128 strings. Bases are fitted on 256 seed-42 strings; the seed-42 column evaluates 128 strings. The PCA control uses 128 directions per MLP and 64 per head.}
\label{tab:coverage}
\end{table}

%% file: sections/probe_appendix.tex
\subsection{A matched held-out stack-decoding control}
\label{app:probe-repair}
We additionally evaluate a frozen linear decoder of stack contents at
layer~5, using independently generated strings for probe fitting and
assessment. We fit four logistic-regression probes on $128$ strings from
seed~42 and evaluate each on three disjoint batches of $128$ strings
(seeds~43--45). Rank-revealing SVD constructs the rank-$24$ stack spaces
at layers~1--5 from the seed-42 coefficient fits.
Every clean and perturbed score uses exactly the same
held-out strings and positions. The probes and subspaces remain fixed throughout the
intervention battery.

The model is Pythia-1B at checkpoint~$143{,}000$, evaluated in fp32 on
Dyck-$(4,16)$ with opening probability $0.45$ and maximum bracket length~48.
For slot $i$, the target is the bracket type at the $i$th stack position
counting from the bottom, immediately before the current token is consumed;
we score all real positions where that slot exists. The current token remains
visible to the model. These scores quantify stack-state decodability, separately
from the next-closer task metric. We compare the clean
residual, removal of each layer's stack subspace from the outputs of
L1--L5.mlp or L1--L3.mlp, and removal of an independently drawn random
orthonormal subspace of the same rank at each of L1--L5.mlp (one fixed
random seed). All four probes converge before the $1{,}000$-iteration cap (Table~\ref{tab:probe-repair}).

\begin{table}[t]
\centering
\small
\setlength{\tabcolsep}{3pt}
\begin{tabular}{ccccc}
\toprule
Slot & Clean & STACK L1--5 & STACK L1--3 & Random L1--5 \\
\midrule
1 & $0.794\pm0.012$ & $0.452\pm0.046$ & $0.500\pm0.052$ & $0.792\pm0.013$ \\
2 & $0.606\pm0.017$ & $0.336\pm0.040$ & $0.461\pm0.032$ & $0.586\pm0.014$ \\
3 & $0.569\pm0.028$ & $0.371\pm0.007$ & $0.329\pm0.021$ & $0.560\pm0.031$ \\
4 & $0.500\pm0.037$ & $0.205\pm0.075$ & $0.234\pm0.064$ & $0.490\pm0.043$ \\
\bottomrule
\end{tabular}
\caption{Matched held-out stack-decoding accuracy. Entries are means
$\pm$ sample standard deviations across three independently generated
batches. The pooled position counts for
slots~1--4 are $7{,}605$, $4{,}769$, $2{,}869$, and $1{,}690$.
Every entry in a row uses the same examples.}
\label{tab:probe-repair}
\end{table}

Removing the selected stack directions at L1--L5 lowers held-out decoding
accuracy by $19.8$--$34.1$ percentage points, whereas the rank-matched
random control lowers it by $0.1$--$2.1$ percentage points. This supports the information-bearing role of the
selected directions under the stated intervention and decoder. The behavioral contribution of these directions is tested separately
through the held-out output interventions in Appendix~\ref{app:algorithm-detail}.
The three evaluation batches quantify variation across input strings with
one fixed basis fit and one random-control draw.

%% file: sections/readout_development.tex
\subsection{Matching-opener retrieval: content and position}
\label{app:readout-detail}
The late readout head \C{L14H1} combines content and positional information. Its 256-dimensional query/key space contains 64 rotary coordinates and 192 unrotated coordinates. Recomputing attention from the two score contributions separately shows that the unrotated coordinates retain most of the observed matching-opener localization. Averaged equally over matching-distance bins 2--24, the attention argmax coincides with the matching opener in $17.57\%$ of clean cases, $15.27\%$ with unrotated coordinates alone, and $0.041\%$ with rotated coordinates alone. This diagnostic uses 256 $k=4$ strings generated with seed 30, maximum bracket length 64, and token windows of length 80.

A complementary key intervention holds the positional phase fixed. At the matching-opener position, we substitute a key from another sequence with a different opener type at that same position. Across 6,212 query--donor pairs, mean attention at the matching opener falls from $0.1833$ to $0.0632$, a $65.5\%$ reduction. The matched position fixes the rotary phase while the donor changes key content. Together, the score decomposition and key intervention identify content-dependent retrieval as an important part of the head's computation. The head's answer-direction intervention separately establishes how its output contributes to bracket prediction.

\subsection{Redistribution of a computational role during training}
\label{app:head-succession}
The mechanistic trace resolves changes in which components support a behavior. Table~\ref{tab:head-succession} follows three layer-7 heads on the $k=2$ task. \C{L7H4} contributes early, \C{L7H5} is prominent at step 3,000, and \C{L7H2} retains a positive contribution at the final checkpoint. By step 32,000, removing \C{L7H4} or \C{L7H5} improves the contrast. These signed trajectories distinguish a component's represented information from the way its output is used by the surrounding network.

\begin{table}[t]
\centering
\begin{tabular}{rrrr}
\toprule
Training step & \C{L7H4} & \C{L7H5} & \C{L7H2} \\
\midrule
1,000 & $0.280$ & $-0.014$ & $-0.014$ \\
3,000 & $-0.139$ & $0.244$ & $0.205$ \\
5,000 & $0.111$ & $0.137$ & $0.279$ \\
32,000 & $-0.242$ & $-0.080$ & $0.123$ \\
143,000 & $-0.053$ & $-0.027$ & $0.219$ \\
\bottomrule
\end{tabular}
\caption{\textbf{A changing allocation of the bracket computation.} Entries are the drop in pairwise closer contrast under individual zero-ablation in the saved $k=2$ trajectory. Positive values indicate a contribution to the score, and negative values indicate an opposing effect.}
\label{tab:head-succession}
\end{table}

This developmental account supplies a circuit-level reference for the geometric trajectories: both the represented variables and the components carrying their causal effects evolve during pretraining.

\paragraph{Consolidation in the greedy circuit trace.}
The $k=2$ greedy search first reaches $80\%$ clean-contrast recovery with 24, 22, 16, 17, and 13 components at steps 1,000, 4,000, 16,000, 32,000, and 143,000, respectively. The threshold size varies across training and is smaller at convergence. \C{L14H1} moves from eighteenth in the addition order at step 1,000 to second at the three intermediate checkpoints and fifth at convergence.